\documentclass{fairmeta}
\usepackage[T1]{fontenc}
\usepackage{wrapfig}
\usepackage{placeins}
\usepackage{needspace}

\usepackage{amsmath,amsfonts,bm}

\def\eqref#1{equation~\ref{#1}}

\def\1{\bm{1}}

\DeclareMathAlphabet{\mathsfit}{\encodingdefault}{\sfdefault}{m}{sl}
\SetMathAlphabet{\mathsfit}{bold}{\encodingdefault}{\sfdefault}{bx}{n}

\usepackage{graphicx}
\usepackage{wrapfig}
\usepackage{booktabs}
\usepackage{makecell}
\usepackage{subcaption}
\usepackage{hyperref}
\usepackage{url}
\usepackage{array}
\usepackage[dvipsnames,table]{xcolor}
\usepackage{tikz}
\usetikzlibrary{calc}
\usepackage{amssymb}
\usepackage{comment}
\usepackage{lipsum}
\usepackage{enumitem}
\usepackage{multirow}
\usepackage{caption}
\usepackage{pifont}
\usepackage{longtable}
\usepackage{textcomp}
\usepackage[var0]{inconsolata}
\usepackage{tabularx}
\usepackage{ragged2e}
\usepackage{listings}
\usepackage{multicol}
\usepackage{adjustbox}
\usepackage{mdframed}
\definecolor{faintblue}{RGB}{240,245,255}
\usepackage{float}
\usepackage{tcolorbox}
\newcolumntype{P}[1]{>{\RaggedRight\ttfamily\arraybackslash}p{#1}}
\newcommand{\code}[1]{\texttt{\nolinkurl{#1}}}

\definecolor{monokai-bg}{RGB}{255, 254, 250}
\definecolor{monokai-fg}{RGB}{2, 2, 2}
\definecolor{monokai-comment}{RGB}{117,113,94}
\definecolor{monokai-keyword}{RGB}{249,38,114}
\definecolor{monokai-string}{RGB}{230, 169, 116}
\definecolor{monokai-function}{RGB}{73, 179, 114}
\definecolor{monokai-number}{RGB}{174,129,255}
\definecolor{monokai-class}{RGB}{0, 118, 207}
\definecolor{monokai-decorator}{RGB}{253,151,31}

\definecolor{md-bg}{RGB}{255, 254, 250}
\definecolor{md-text}{RGB}{36,41,47}
\definecolor{md-heading}{RGB}{74, 72, 72}
\definecolor{md-code-bg}{RGB}{246,248,250}
\definecolor{md-code}{RGB}{39, 122, 217}

\newcommand{\airsbench}{\textsc{AIRS-Bench}}
\newcommand{\airadojo}{\textsc{AIRA-dojo}}

\title{Agentic Discovery of Neural Architectures: AIRA-Compose and AIRA-Design}

\author[1,*]{Alberto Pepe}
\author[1,*]{Chien-Yu Lin}
\author[1]{Despoina Magka}
\author[1]{Bilge Acun}
\author[1,2]{Thomas Simon Foster}
\author[1]{Yannan Nellie Wu}
\author[1]{Anton Protopopov}
\author[1,\ddag]{Carole-Jean Wu}
\author[1,\ddag]{Yoram Bachrach}

\affiliation[1]{FAIR at Meta}
\affiliation[2]{University of Oxford}

\contribution[*]{Joint first author}
\contribution[\ddag]{Joint last author}

\abstract{
As a step toward recursive self-improvement, we investigate the ability of LLM agents to autonomously design foundation models beyond the standard Transformer paradigm. We introduce a dual-framework approach: \textbf{AIRA-Compose} for high-level architecture search, and \textbf{AIRA-Design} for low-level mechanistic implementation. \newline \textbf{AIRA-Compose} deploys an ensemble of 11 agents to navigate a combinatorial design space of fundamental computational primitives (Attention, MLP, Mamba) under a fixed 24-hour compute budget. Operating in two stages, agents iteratively design and evaluate candidates at the million-parameter scale, after which top-performing designs are extrapolated to 350M, 1B, and 3B parameter scales. This search yields 14 novel architectures spanning two families: \textit{AIRAformers} (Transformer-based) and \textit{AIRAhybrids} (Transformer-Mamba-based). When pre-trained at the 1B scale under a fixed token budget, agent-discovered top-performing architectures consistently outperform both Llama 3.2 and Composer-found alternatives. On downstream tasks, AIRAformer-D and AIRAhybrid-D improve accuracy by 2.4\% and 3.8\% over Llama 3.2, respectively. AIRA-Compose also finds novel model architectures that achieve steeper, more efficient compute-optimal scaling frontiers. AIRAformer-C scales 54\% and 71\% faster than Llama~3.2 and the best Composer-found Transformer, while AIRAhybrid-C scales 23\% and 37\% faster than the modified Nemotron-2 and the best Composer-found hybrid, respectively. \newline \textbf{AIRA-Design} tasks up to 20 agents with directly writing novel attention mechanisms to handle long-range dependencies and implementing high-performing training scripts. Evaluated on the Long Range Arena (LRA) benchmark, the best agent-designed architectures achieve accuracy within 2.3$\%$ of human state-of-the-art on document matching and 2.6$\%$ on text classification. On the Autoresearch benchmark, Greedy Opus 4.5 optimizes training under a fixed time budget to achieve 0.968 validation bits-per-byte, surpassing the published minimum reference. \newline Together, AIRA-Compose and AIRA-Design demonstrate that AI research agents can autonomously discover hybrid architectures and algorithmic optimizations that rival or surpass hand-designed baselines. This establishes a flexible, powerful paradigm for discovering the next generation of foundation models and a step towards recursive self improvement.
}

\date{\today}
\correspondence{Despoina Magka at \email{despoinam@meta.com}, Yoram Bachrach at \email{yorambac@meta.com}}

\begin{document}

\maketitle


\section{Introduction}


Agents powered by Large Language Models (LLMs) are radically transforming scientific research and have evolved into autonomous systems capable of executing end-to-end research loops \citep{wang2024survey,andrews2025arescalingagentenvironments,schmidgall2025agent,yamada2025aiscientistv2workshoplevelautomated}. Today, agents can independently formulate hypotheses, write and execute code, evaluate results, and iteratively refine their approaches to solve complex tasks. These include discovering new mathematical constructions \citep{romera2024mathematical,novikov2025alphaevolve}, achieving human-level performance in competitive programming \citep{leblond2023alphacode2,li2024autokagglemultiagentframeworkautonomous,jimenez2024swebenchlanguagemodelsresolve}, replicating existing AI research literature \citep{xiang2025scireplicate,starace2025paperbench}, designing and generating correct, faster, and more efficient machine learning (ML) kernels for custom AI hardware~\citep{liao2026kernelevolvescalingagentickernel,hammond2025agenticoperatorgenerationml}, and driving open-ended ML discoveries \citep{zhao2025automatedllmspeedrunningbenchmark,nathani2025mlgym,lupidi2026airs}.

\begin{figure}[H]
    \centering
    \includegraphics[width=\linewidth]{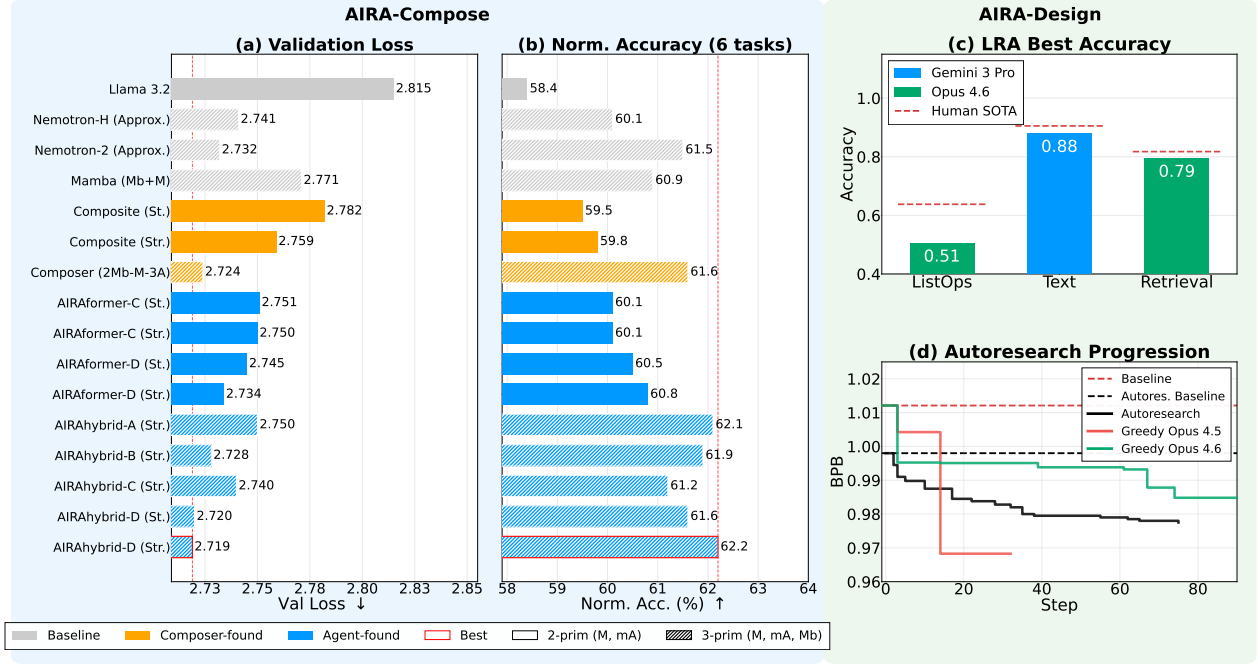}
    \caption{\textbf{AIRA-Compose and AIRA-Design: agentic frameworks for neural architecture search and model design.} \textbf{(a--c)}~Downstream evaluations of selected agent-found architectures scaled-up to 1B scale with fixed token budget, alongside baselines and traditional NAS-found models: (a)~validation loss, and (b)~zero-shot average normalized accuracy across 6 tasks. \textbf{(c)}~Best test accuracy after 24 GPU hours on the three Long Range Arena tasks. Greedy Opus~4.6 achieves the highest scores on ListOps (0.51) and Retrieval (0.79); Greedy Gemini~3 Pro leads on Text (0.88). \textbf{(d)}~Autoresearch training-script optimization: cumulative best bits-per-byte (BPB) over agent steps. Greedy Opus~4.6 achieves the lowest BPB across 100 runs.}
    \label{fig:papersummary}
\end{figure}

A compelling frontier in agentic research is \textbf{Recursive Self-Improvement (RSI)} \citep{good1966speculations,schmidhuber1987evolutionary,schmidhuber2003godel}. In this manuscript, we will use RSI to refer to agents that autonomously discover and optimize the very neural architectures that power them, ultimately advancing their capabilities. Most LLMs today rely on the Transformer architecture \citep{vaswani2017attention,radford2018improving,devlin2019bert}, which is characterized by a 1:1 interleaving of multi-head attention and multi-layer perceptron (MLP) layers. Although Transformers remain the gold standard, the research community is increasingly shifting toward post-Transformer paradigms \citep{peng2023rwkv,sun2023retentive,gu2024mamba,rahmani2025implicit} to address their inherent limitations, such as the quadratic complexity of attention and the memory cost of key-value (KV) caching during inference \citep{tay2022efficient,pope2023efficiently}. Moreover, arranging distinct computational primitives into sophisticated hybrid patterns has been shown to yield more efficient and performant foundation models \citep{adler2024nemotron,blakeman2025nemotron,lieber2024jamba,yang2025qwen3,singhal2025llama,nemotron3super2026,bae2025hybrid}. We refer to such models as \textit{hybrid} LLMs. Model design has historically been driven by domain expertise and human intuition. However, as we move toward hybrid models, the design space becomes vast and combinatorial \citep{liu2021survey,ren2021comprehensive}. Relying solely on manual exploration means that highly performant, non-obvious configurations or entirely new computational primitives may be overlooked.
We hence argue that \textbf{Neural Architecture Search (NAS)} represents an ideal candidate for AI research agents: agents pair educated, context-aware hypotheses with robust automated search and iterative refinement, allowing to explore the design space systematically and propose fundamentally new architectures.

We propose two approaches for the agentic discovery of neural architectures for future hybrid LLMs:

\begin{itemize}
    \item \textbf{AIRA-Compose, or high-level architecture search}: Agents are tasked to find new model architectures based on \textit{predefined} computational primitives, by searching and evaluating model architecture candidates at small scale. We do so by recasting the small scale model architecture exploration of the \textit{Composer} framework \citep{acun2025composer} into an agentic task.  Only top-performing model architectures are scaled up.
    \item \textbf{AIRA-Design, or low-level mechanistic design}: Agents implement \textit{novel} computational primitives and train them efficiently. These models are crafted to tackle the Long Range Arena (LRA) \citep{tay2020long} and Autoresearch benchmarks \citep{karpathy2026autoresearch}.
\end{itemize}

The two approaches represent complementary perspectives on a single goal. The former relies on predefined computational primitives, restricting the agents' freedom exclusively to the optimization of their arrangement. The latter is an open-ended code-generation task in which agents must implement fundamentally new models from scratch. Both methodologies are implemented as \airsbench{} \textit{tasks} \citep{lupidi2026airs}, which evaluate an agent's ability to conduct independent scientific research. \airsbench{} introduces a flexible and scalable structure that enables virtually any ML problem to be cast into a format that agents can understand and operate on. We list contributions below:

\begin{itemize}
  \item We introduce AIRA-Compose and AIRA-Design --- two flexible, agent-based discovery frameworks built on the \airsbench{}~task standard to enable Recursive Self-Improvement. These frameworks introduce 12 agentic tasks with a scalable structure for autonomous research loops. They can be readily extended to operate with different agent harnesses and/or use different reasoning models.

  \item We demonstrate that our agents can autonomously discover novel, highly performant hybrid architectures by searching and optimizing the arrangement of predefined computational primitives. We categorize them into two families: \textit{AIRAformers}, which include multi-head attention and MLP blocks, and \textit{AIRAhybrids}, which also include Mamba2 State Space Model (SSM) blocks. Both families are characterized by original interleavings of such primitives. When extrapolated to 350M, 1B, and 3B parameter scales, these agent-discovered models exhibit superior isoFLOP scaling properties and consistently outperform established baselines, such as, Llama~\citep{llama2024} and Nemotron~\citep{blakeman2025nemotron}, as well as those found via optimization-based NAS.

  \item We show that our agents are capable of autonomously engineering high-performing attention mechanisms. On the Long Range Arena (LRA) benchmark, agent-designed models achieved a peak accuracy within 2.3$pp$ of human SOTA on document matching (82\%) and 2.6$pp$ on text classification (91\%). Four of our agents achieve an average normalized score above 0.3 across the 3 tasks, with 1 corresponding to human SOTA.

\item We show that our agents are capable of iteratively improving the efficiency of small language models training loops. On the open-ended Autoresearch task, our best agent surpassed the published Autoresearch reference minimum by achieving a validation BPB of 0.968. Moreover, we could reproduce a delta with respect to baseline larger than that reported in \cite{karpathy2026autoresearch} in 20$\%$ of the experiments.

\end{itemize}

We summarize them in Figure~\ref{fig:papersummary}. Panels~(a--b) present downstream evaluations of selected agent-found architectures scaled-up to 1B scale with a fixed token budget, alongside baselines and traditional NAS-found models: (a)~validation loss and (b)~zero-shot average normalized accuracy across 6 tasks. Agent-discovered AIRAformer and AIRAhybrid models consistently outperform both established baselines (Llama~3.2, approximated Nemotron-2) and Composer-found alternatives across all three metrics.

Panels~(c) and~(d) present results from AIRA-Design, where agents must write functional code from scratch rather than arrange predefined blocks. On the Long Range Arena benchmark~(c), agents implement novel sub-quadratic attention mechanisms and train them within a fixed GPU budget; the best agent-designed models reach accuracy within 2--3 percentage points of human SOTA across all three tasks. On the Autoresearch benchmark~(d), agents iteratively optimize a GPT training script to minimize validation bits-per-byte within a 5-minute wall-clock budget; augmenting the strongest agents with curated literature and code repositories shifts their optimization strategies and yields the lowest BPB across all 100~runs.

The rest of the paper is structured as follows: fundamentals on the \airadojo{}~harness and the \airsbench{}~task structure are described in Section~\ref{sec:method}; details on AIRA-Compose and AIRA-Design tasks, their objective and datasets employed are presented in Sections~\ref{sec:compose} and \ref{sec:design}, respectively; the experimental setup and relevant metrics are introduced in Section~\ref{sec:experiments}; results are presented in Section~\ref{sec:results} and grouped into 4 subsections: NAS on two (Section~\ref{sec:results1}) and three (Section~\ref{sec:results2}) computational primitives within AIRA-Compose, Long Range Arena (Section~\ref{sec:results3}) and Autoresearch (Section~\ref{sec:results4}) within AIRA-Design; conclusions are drawn in Section~\ref{sec:conclusions}. A review of related work is provided in Appendix~\ref{app:appendix_related}.

\section{Methodology}
\label{sec:method}

We build on the definitions and evaluation framework established by \airsbench{} \citep{lupidi2026airs}. Because this infrastructure has been extensively detailed in prior work, we provide only a brief overview. We adopt the definition of an \textit{agent} as the combination of an LLM and a \textit{scaffold}. The scaffold represents the algorithmic search policy and the specific set of operators that determine how the agent explores the solution space. This scaffold is instantiated within a \textit{harness}, the system that manages the agent's environment, execution, and tool access.

We utilize \airadojo~\citep{toledo2025airesearchagentsmachine}, a harness designed to evolve code solutions through tree-based exploration. \airadojo~guides the agent using structured search policies (e.g., greedy search or Monte Carlo Tree Search) and interacts with candidate Python solutions using four operators:
\begin{itemize}
    \item \textit{Draft:} Generates the initial set of candidate solutions.
    \item \textit{Debug:} Identifies and corrects execution or logical errors.
    \item \textit{Improve:} Refines working solutions to maximize (or minimize) the target evaluation metric.
    \item \textit{Analyze:} Reads and analyzes the agent solution at each step.
\end{itemize}

We employ one-shot and greedy scaffolds. The one-shot scaffold corresponds to calling the \textit{draft} operator once, and exactly one solution will be produced per run. The greedy scaffold, on the other hand, explores several solutions through a tree-based search policy, and it starts the search by drafting 5 initial solutions through the \textit{draft} operator. The \textit{improve} operator is applied onto the agent solution that achieves the highest validation fitness. A layer is populated until a new best is found. The \textit{debug} operator is applied whenever the \textit{analyze} operator deems a solution buggy (e.g., the agent produces an architecture with an incorrect number of primitives or an OOM error is raised). Each greedy run will explore several solutions, or ``steps'', ranging from tens to hundreds per run, based on the compute required by the evaluation script of each task. Each step of the search has a validation fitness, which comes from the agent-generated code to support its submission (i.e., an arrangement of primitives for AIRA-Compose tasks, or a \texttt{model.py/train.py} for AIRA-Design tasks), and a test fitness, which is obtained by independently evaluating the agent submission through an independent evaluation script (see below).

Both our architecture search and mechanistic design approaches are formulated as \airsbench{} tasks (see Table~\ref{tab:tasks}). An \airsbench{} task is fully specified by a \{problem, dataset, metric\} triplet: The \textbf{problem} defines the challenge to be solved (e.g., Neural Architecture Search); the \textbf{dataset} specifies which data to solve the challenge over (e.g., DCLM); the \textbf{metric} is used to quantify fitness performance (e.g., loss).

\definecolor{pastelblue}{RGB}{230, 242, 255}
\definecolor{pastelgreen}{RGB}{232, 248, 236}

\begin{table}[htbp]
\centering
\caption{\textbf{The 12 RSI tasks introduced in this manuscript}. AIRA-Compose includes 4 tasks encompassing 3 datasets with two and three primitive pools, requiring agents to submit a \texttt{submission.csv} file containing a string of 16 primitives to describe an architecture. AIRA-Design tasks are split into LRA (Long Range Arena) for architectural implementation and Autoresearch for high-performing training scripts, requiring agents to submit a \texttt{model.py} or a \texttt{train.py} artifact, respectively.}
\small
\label{tab:tasks}
\begin{tabularx}{\textwidth}{
  >{\RaggedRight\arraybackslash}p{3cm}
  >{\RaggedRight\arraybackslash}X
  >{\RaggedRight\arraybackslash}p{4.5cm}
}
\toprule
\textbf{Framework} & \textbf{Tasks (Problem, Dataset, Metric)} & \textbf{Submission Artifact} \\
\midrule
\textbf{AIRA-Compose}: High-level architecture search &
\textbf{2 Primitives (M, mA) Search:} \par
\quad {\footnotesize \texttt{NeuralArchitectureSearchMAD2PrimitivesAccuracy}} \par
\quad {\footnotesize \texttt{NeuralArchitectureSearchBabiStories2PrimitivesLoss}} \par
\quad {\footnotesize \texttt{NeuralArchitectureSearchDCLM2PrimitivesLoss}} \vspace{0.5em}

\textbf{3 Primitives (M, mA, Mb) Search:} \par
\quad {\footnotesize \texttt{NeuralArchitectureSearchMAD3PrimitivesMbAccuracy}}

\vspace{0.5em}

&
\texttt{submission.csv} \vspace{0.5em}

\begin{tcolorbox}[colback=blue!5, boxrule=0pt, frame hidden, arc=4pt, left=4pt, right=4pt, top=4pt, bottom=4pt]
\texttt{\footnotesize mh-attention mlp mlp} \par
\texttt{\footnotesize mlp mh-attention mlp} \par
\texttt{\footnotesize ...} \par
\texttt{\footnotesize mh-attention mlp} \par
{\scriptsize (16 primitives total)}
\end{tcolorbox} \\
\midrule
\textbf{AIRA-Design}: Low-level mechanistic design &
\textbf{LRA (Long Range Arena):} \par
\quad {\footnotesize \texttt{LongContextReasoningLRATextAccuracy}} \par
\quad {\footnotesize \texttt{LongContextReasoningLRAListOpsAccuracy}} \par
\quad {\footnotesize \texttt{LongContextReasoningLRARetrievalAccuracy}} \par
\quad {\footnotesize \texttt{LongContextReasoningLRATextConfigurableAccuracy}} \par
\quad {\footnotesize \texttt{LongContextReasoningLRAListOpsConfigurableAccuracy}} \par
\quad {\footnotesize \texttt{LongContextReasoningLRARetrievalConfigurableAccuracy}} \vspace{0.5em}

\textbf{Autoresearch:} \par
\quad {\footnotesize \texttt{AutoregressiveLanguageModellingAutoresearchBPB}} \par
\quad {\footnotesize \texttt{AutoregressiveLanguageModellingWithLiteratureAutoresearchBPB}}
&
\texttt{model.py / train.py} \vspace{0.5em}

\begin{tcolorbox}[colback=green!5, boxrule=0pt, frame hidden, arc=4pt, left=4pt, right=4pt, top=4pt, bottom=4pt]
\texttt{\footnotesize import flax.linen as nn} \par
\texttt{\footnotesize class CustomEncoder(} \par
\texttt{\footnotesize \ \ nn.Module):} \par
\texttt{\footnotesize \ \ @nn.compact} \par
\texttt{\footnotesize \ \ def \_\_call\_\_(self, x):} \par
\texttt{\footnotesize \ \ \ \ ...}
\end{tcolorbox} \\
\bottomrule
\end{tabularx}
\end{table}

\airsbench{} tasks have a standardized, modular file structure that translates open-ended ML research problems into a format parsable by agentic harnesses. A standard task directory consists of the following components:
\begin{itemize}
    \item \texttt{project\_description.md}: The prompt provided to the agent. It details the research objective, the dataset schema, and the specific evaluation setup (e.g., instructing the agent what artifact to submit).
    \item \texttt{prepare.py} \& \texttt{evaluate\_prepare.py}: Scripts responsible for one-time data downloading and environment setup. They handle data sanitization, ensuring test labels are hidden while the agent is building its solution.
    \item \texttt{evaluate.py}: The isolated scoring script containing the ground-truth metric implementation used to automatically evaluate the agent's submission against the test set. For tasks in this manuscript, \texttt{evaluate.py} trains the agent-generated architecture or launches the training script produced.
    \item \texttt{metadata.yaml}: A configuration file defining the task constraints, evaluation metrics, required dataset splits, and any specific library dependencies needed for the evaluation environment.
\end{itemize}

\textbf{Enhancing AIRS-Bench}. The roles of \texttt{submission.csv} (i.e., the artifact that the agent is required to produce) and \texttt{evaluate.py} (i.e., how the artifact is evaluated) differ between our RSI tasks and standard \airsbench{} tasks. In standard \airsbench{}, \texttt{submission.csv} contains model predictions on test data (e.g., a regression quantity or predicted class). In our RSI tasks, \texttt{submission.csv} specifies a complete architecture or a training loop. Similarly, \texttt{evaluate.py} does not simply include metric calculations, like F1 scores or Spearman correlations as in \airsbench{}, but it encapsulates full training pipelines ported directly from the Composer, LRA, and Autoresearch codebases. This structural choice guarantees consistent evaluation across all agent-generated neural architectures and baselines, enabling agents to focus exclusively on architectural innovations rather than the engineering overhead of implementing training loops.

\section{The AIRA-Compose Pipeline}
\label{sec:compose}

\begin{figure}[!htbp]
    \centering
    \includegraphics[width=\linewidth]{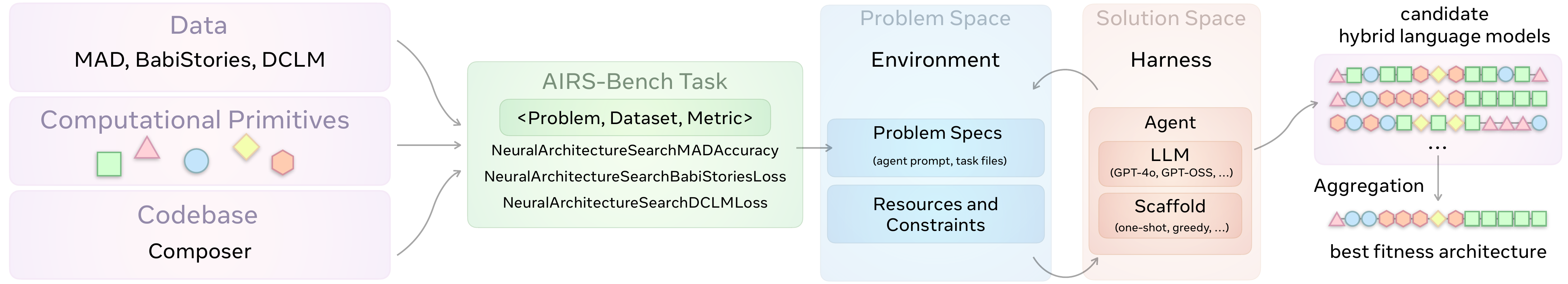}
    \caption{\textbf{The AIRA-Compose pipeline}. We recast \textit{Composer} into equivalent \airsbench{} tasks. These tasks instruct agents to combine predefined computational primitives to assemble a 16-layer small scale architecture. Tasks are run employing the \airadojo{} harness with 6 LLMs as core reasoning models. Search results are then ranked, aggregated and scaled to 350M, 1B and 3B parameters size.}
    \label{fig:composertask}
\end{figure}

AIRA-Compose (Figure~\ref{fig:composertask}) relies on the \textit{Composer} framework \citep{acun2025composer}. Composer discovers hybrid foundation models with a four-step process: (1) The \textit{Search Engine} uses Bayesian Optimization combined with incremental layer search and width-scaling to discover primitive arrangements;  (2) The \textit{Evaluator} is a fast-proxy training and evaluation loop. Candidate architectures from the search are trained on small proxy datasets; (3) The \textit{Aggregator} post-processes the top candidate architectures discovered during the search. It employs layer-wise clustering techniques to select the most frequent computational primitive to obtain a robust small scale architecture while smoothing out the noise and overfitting coming from proxy training; (4) The \textit{Extrapolator} scales the aggregated small scale architecture to a desired target parameter count (e.g., 350M, 1B, or 3B). This is achieved through \textit{stretching} (proportionally expanding contiguous blocks) or \textit{stacking} (repeating the entire discovered architecture sequentially).

\begin{figure}[!htbp]
    \centering
    \includegraphics[width=\linewidth]{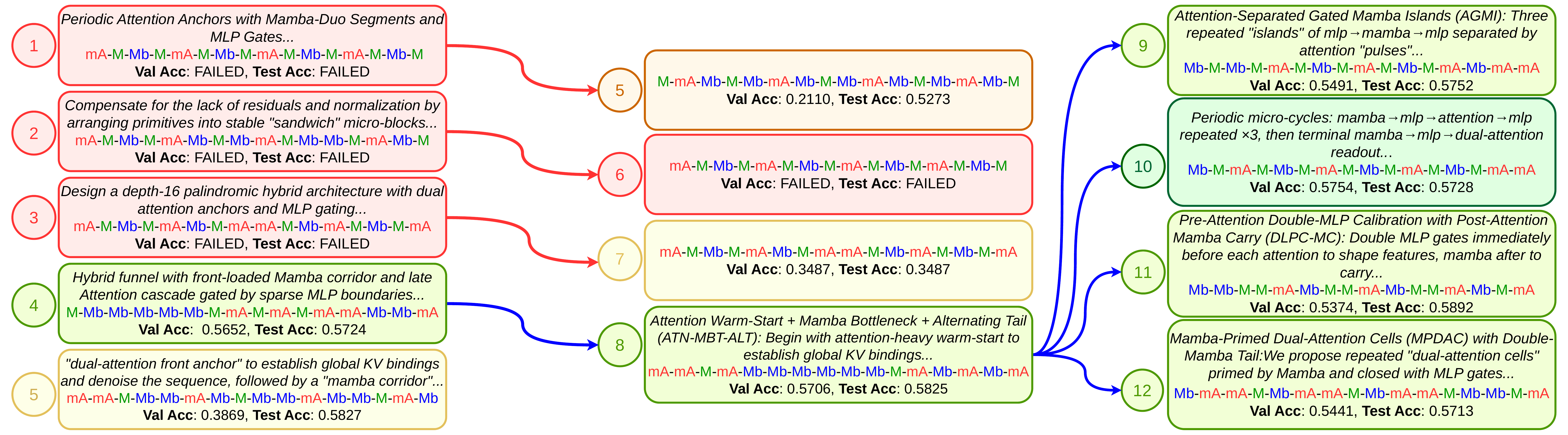}
   \caption{\textbf{Greedy GPT-5 search (partial) on the 3-primitive NAS task within AIRA-Compose}. The agent drafts five initial solutions and iteratively explores the architecture space via a greedy tree search. At each node, the agent reasons on different design choices, writes a candidate architecture to \texttt{submission.csv} and evaluates it. The highest-scoring node is then selected for further exploration. Red arrows indicate \textit{debug} operations, while blue arrows indicate \textit{improve} operations. At each improvement, the agent proposes a new architecture informed by the parent's reasoning and score. We report a snippet of the agent's design rationale, the submitted architecture string, the agent's internal validation accuracy, and the true test accuracy obtained from the isolated scoring script.}
    \label{fig:agentsearch}
\end{figure}

AIRA-Compose recasts steps 1 and 2 as \airsbench{} tasks. Rather than relying on rigid Bayesian Optimization and deterministic incremental search, agents can freely formulate structural hypotheses and propose novel primitive arrangements  (see Figure~\ref{fig:agentsearch}), evaluate them, and iteratively refine their designs based on their prior knowledge. We focus on 16 layer search, since they are small scale proxies that correlate well with equivalent large scale model. Once the agentic exploration is concluded, we leverage steps 3 and 4 to scale the agents' discoveries for final, large-scale evaluation. For the aggregation step, we collect the submitted architectures along with their test scores across all steps from all agents (see Appendix~\ref{app:ranking}).

An example of agent-driven NAS is given in Figure~\ref{fig:agentsearch}, showing the first few nodes from a Greedy GPT-5 run on the 3-primitive task. Compared to the Composer framework, which relies on fixed search methodologies, architectures designed at each step come from the agent's own understanding of the problem as presented in \texttt{project\_description.md}. At each node, the agent articulates its design choices, produces a candidate architecture as a \texttt{submission.csv} file, and writes an evaluation script to validate its reasoning. The submitted architecture is then trained and evaluated independently in \texttt{evaluate.py}. The node with the highest validation score is selected for further exploration via \emph{improve} operations, which propose new architectures informed by the parent's reasoning and score. This process allows the agent to leverage domain knowledge to navigate the combinatorial space in a meaningful way, rather than relying on predefined structural templates or hand-crafted mutation operators. Scaled across several hundreds of nodes and multiple independent agents, this process yields a semantically diverse exploration, which we believe to be the ultimate advantage of AIRA-Compose.

\textbf{Primitives}. A \textit{primitive} is a computational building block from which model architectures are assembled. We consider \textit{two-primitive} and \textit{three-primitive} search spaces. We employ MLPs (M), multi-head Attention (mA) and Mamba SSM (Mb). The two- and three- primitive search spaces span $2^{16} = 65{,}536$ and $3^{16} \approx 43$M possible 16-layer arrangements, respectively. Their configurations at small scale and 350M, 1B, and 3B scale are provided in Appendix~\ref{app:scaling}.

\textbf{Datasets} We evaluate architectures on three proxy datasets as in \cite{acun2025composer}, using metrics chosen to reliably predict at-scale performance: (i) \textbf{MAD} \citep{poli2024mechanistic}, a suite of six synthetic token-manipulation tasks (e.g., selective copying, compression, and in-context recall). Models are trained on 800 samples and tested on 1,280, using average accuracy as the metric; (ii) \textbf{BabiStories} \citep{zhang2025memory}, a synthetic corpus of children's stories. Models train on 927,158 samples and are evaluated on 9,275 via cross-entropy loss; (iii) a fixed subset of \textbf{DCLM} \citep{li2024datacomp}. Models train on 10,000 samples and test on 9,275 via cross-entropy loss. A larger portion of the DCLM corpus is reserved for the large-scale pretraining phase. We allow agents to validate their hypotheses by creating a 70:30 train/validation split on the original train set \citep{hambardzumyan2026aira_2}. Submitted architectures are trained from scratch on the full training set and evaluated on the test set withheld from the agent.

\section{AIRA-Design: low-level mechanistic design}
\label{sec:design}

\begin{figure}[!htbp]
    \centering
    \includegraphics[width=\linewidth]{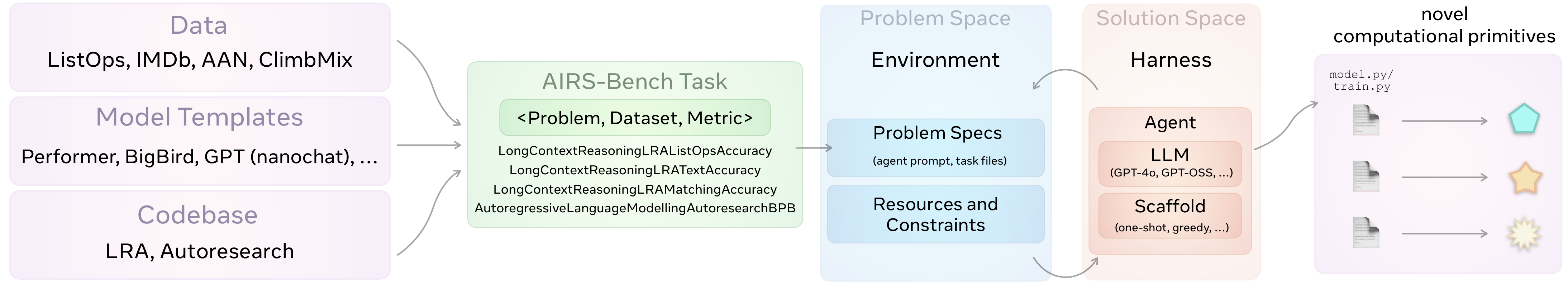}
    \caption{\textbf{AIRA-Design: agent-driven, low-level mechanistic design}. We recast the LRA and Autoresearch codebases into equivalent \airsbench{} tasks. These tasks instruct agents to design novel computational primitives / full training loops and write them into a \texttt{model.py/train.py} file. Tasks are run employing the one-shot and greedy scaffolds of the \airadojo{} harness with 12 different LLMs as core reasoning models, for a total of up to 20 agents.}
    \label{fig:lratask}
\end{figure}

For the set of the AIRA-Design tasks, we leverage (1) the LRA benchmark \citep{tay2020long} --- a benchmark suite designed to evaluate sequence models on their ability to capture long-range dependencies, and (2) the Autoresearch benchmark \citep{karpathy2026autoresearch}, an open-ended challenge where agents must autonomously explore, implement, and validate novel research ideas to improve a model's training efficiency.  Unlike the high-level architecture search, where agents select and arrange predefined computational primitives, the AIRA-Design tasks require agents to engage in \textit{low-level mechanistic design}: implementing novel neural network architectures from scratch (LRA), as well as optimizing their training loop (Autoresearch).

\textbf{Long Range Arena (LRA)} tasks challenge agents to design memory-efficient, sub-quadratic attention mechanisms that avoid materializing full $O(n^2)$ attention matrices for sequences exceeding 2000--4000 tokens, while preserving the capacity to model long-range interactions. To accomplish this, agents must implement their chosen architecture by producing a complete, executable \texttt{model.py} file that defines a \texttt{CustomEncoder} (or \texttt{CustomDualEncoder} for the matching task) compatible with the provided JAX/Flax training infrastructure.

We evaluate performance on two task versions: \textit{Configurable} and \textit{Non-Configurable}. The \textit{Configurable} version of the tasks allows agents to explicitly specify key global variables directly within \texttt{model.py}. This allows them to adjust hyperparameters to better suit their architectures (see Appendix~\ref{app:configurable}, Table~\ref{tab:model_config_defaults}). Conversely, the \textit{Non-Configurable} setup enforces the default LRA parameters, restricting agents freedom to exclusively evaluate agents' design capabilities, independently of their hyperparameter tuning skills.  We include the \textit{Configurable} version of the task to allow agents to move toward configurations closer to current SOTA models, which support up to 2 million tokens \citep{ma2024megalodon}.

\textbf{Autoresearch} is an open-ended optimization benchmark in which agents must autonomously improve a GPT language model training script. The goal is to achieve the lowest possible validation bits per byte within a fixed 5-minute wall-clock training budget on a single GPU. Agents are given a baseline \texttt{train.py} containing a nanochat model implementation with multi-head causal attention, rotary embeddings, flash attention, and a hybrid Muon--AdamW optimizer. Agents are allowed to modify any aspect of this script, including model architecture (depth, width, attention patterns, positional encodings, activation functions), optimizer and learning rate schedule hyperparameters, and training configuration (batch size, gradient accumulation, warmup/cooldown ratios). On the other hand, the data pipeline, tokenizer, and evaluation harness are read-only. Compared to the LRA tasks, we assess agents capability to design an advanced architecture \textit{and} an optimized training loop. The task also rewards simplification, and removing unnecessary complexity for additional training steps is explicitly valued when performance is preserved or improved.

We evaluate performance on two task versions: default and \textit{With Literature}. The \textit{With Literature} version of the task is designed to provide agents with relevant context before attempting a solution, and it includes in the task folder a curated selection of 41 research papers provided to the agent as structured paper analyses (\texttt{task\_info.jsonl}) and 14 reference code repositories for some of the papers. Resources are organized into papers on Architecture Improvements (20), Training Strategies (17), and Optimizers (5). The task \texttt{project\_description.md} directs the agent to consult these resources before attempting a solution. A detailed summary of the resources employed in the \textit{With Literature} version of the task is provided in Appendix~\ref{app:autoresearch}, Tables~\ref{tab:literature_architecture}-\ref{tab:literature_optimizers}.

\subsection{Datasets}

The LRA tasks focus on the three text-based datasets of the benchmark. All LRA tasks employ \textbf{accuracy} as the evaluation metric: \textbf{IMDB Sentiment Classification (Text)}: The Text task uses the IMDB movie reviews dataset for binary sentiment classification. Models must capture semantic relationships across the full document context to accurately classify sentiment; \textbf{ListOps}: ListOps presents hierarchical mathematical expressions in prefix notation with nested operations (MAX, MIN, MEDIAN, SUM\_MOD). Example sequences such as \texttt{[MAX 4 3 [MIN 2 3] 1 0]} require models to parse bracket structure, respect operator precedence, and compute the correct numerical result (0--9); \textbf{ACL Anthology Network (AAN) (Retrieval)}: The Retrieval task uses the ACL Anthology Network (AAN) dataset, tokenized at character level, presenting pairs of academic paper abstracts for binary similarity classification. It requires a dual-encoder setup to test whether architectures can efficiently process and compare long documents. For each task, agents have access to training and validation splits during the search phase. The validation split is used to iteratively evaluate hypotheses when running the greedy scaffold. Once agents finalize their \texttt{model.py} submission, the architecture is trained from scratch on the full training set, and final performance is evaluated exclusively on a held-out test split that remains strictly inaccessible during the search phase.

The Autoresearch task ~\citep{karpathy2026autoresearch} consists of pre-tokenized web text from the \textbf{ClimbMix} corpus \citep{diao2025climb}. A pre-trained BPE tokenizer with a vocabulary size of ${\sim}8192$ is provided. The dataset is split into training shards and a pinned validation shard. The evaluation metric is validation \textbf{bits per byte (BPB)}, where lower is better, which is vocabulary-size-independent and enables fair comparison across architectural changes.

\section{Experiments}
\label{sec:experiments}

Our experimental setup and metrics follow the paradigm defined by \airsbench{} \citep{lupidi2026airs}. We refer the reader to the original manuscript for a detailed description. We briefly summarize them here. We test tasks on one-shot and greedy scaffolds within \airadojo{}. We run experiments on 20 seeds for one-shot agents and on 10 seeds for greedy agents, unless otherwise specified. Each run lasts for at most 24 hours or at most 500 steps. Each agent has access to one H200 GPU to draft and validate its solutions. For the search on BabiStories and DCLM, which are larger, we allocate 60 hours per run. Within this time budget, each seed covers a search space of 100--200 small scale architectures in AIRA-Compose, and several hundreds architectures and training loops in AIRA-Design. For AIRA-Compose large scale experiments, 350M models are trained on 8 H200 GPUs, while 1B and 3B models on 16 H200 GPUs. Optimizers and hyperparameters at small and large scale are kept consistent with prior work \citep{acun2025composer}.

\textbf{AIRA-Compose Metrics}. We rank small scale models based on their raw scores (accuracy or cross-entropy loss). We evaluate the scaled-up models across three metrics: \textbf{(i)} cross-entropy validation loss measured on 1000 samples of the DCLM validation set; \textbf{(ii)} 0-shot downstream accuracy and normalized accuracy on 6 tasks: HellaSwag, WinoGrande, ARC-Easy, ARC-Challenge, PIQA, and SciQ; and \textbf{(iii)} the \textbf{DCLM Core score}~\citep{li2024datacomp}, an unweighted average across 14 tasks evaluated under 0-shot (COPA, CoQA, LAMBADA, OpenBookQA, WinoGrande, XWinograd-EN), 3-shot (AGIEval LSAT-AR), and 10-shot (ARC-Easy, ARC-Challenge, BoolQ, CommonsenseQA, HellaSwag, PIQA, SQuADv2). The primary metric is accuracy for most tasks, normalized accuracy for OpenBookQA, AGIEval LSAT-AR, ARC-Easy, ARC-Challenge, HellaSwag, and PIQA, and F1 score for CoQA and SQuADv2.

\textbf{AIRA-Design Metrics}. We present AIRA-Design results using three metrics: \textbf{(i)} raw score, to allow for an immediate comparison with respect to SOTA and established baselines; \textbf{(ii)} valid submission rate (VSR): To assess an agent's capability to consistently formulate a working solution and submit it confidently. The VSR on task $t$ for an agent $a$ is defined as:
    \begin{equation}
    \text{VSR}_{a,t} =  \frac{V_{a,t}}{T_{a,t}}
    \label{eq:meanvalidrate}
    \end{equation}
where $V_{a,t}$ is the number of successfully submitted runs for agent $a$ on task $t$, $T_{a,t}$ is the total number of runs for agent $a$  on task $t$; \textbf{(iii)} normalized score (NS), to provide an intuitive measure of the progress on the benchmark, we define and use NS of an agent $a$ on a task $t$ as:
    \begin{equation}
    \text{NS}_{t}^a = \frac{\phi_t(s_{t}^a) - \phi_t(s^\mathrm{min}_t)}{\phi_t(s^\mathrm{sota}_t) - \phi_t(s^\mathrm{min}_t)}
    \label{eq:normscore}
    \end{equation}
    where $s_t^\mathrm{min}$ corresponds to the worst score observed across all seeds and agents on task $t$, $s_t^\mathrm{sota}$ is the SOTA score sourced from literature, and $s_{t}^a$ is the score achieved by agent $a$ on the same task. This normalizes scores to a value range between $0$ and $1$ for minimum and SOTA performances, respectively. NS $>1$ if it exceeds SOTA. The transformation $\phi_t$ applies the \textbf{march of 9s} \citep{karpathy_dwarkesh_2025}:
    \begin{equation}
    \phi_t(s) = -\log_{10}(|s - s^\mathrm{opt}_t|)
    \label{eq:marchof9s}
    \end{equation}
    where $s^\mathrm{opt}_t$ is the overall possible optimal score for the task (in our case 1.0). Failed and invalid submissions are treated as submissions with a $0$ normalized score.

We regard the LRA tasks as proper, open-ended AI research challenges in pure \airsbench{} style. Because these tasks require writing novel, functional code from scratch, all three metrics are required for comprehensive evaluation. For AIRA-Compose tasks, agents only output a formatted string mapping to predefined blocks, making VSR virtually 100\%; we therefore report only raw scores and the scaled-up evaluation metrics. Similarly, for Autoresearch tasks, we focus on raw scores expressed as validation BPB.

\section{Results}
\label{sec:results}

\subsection{AIRA-Compose}

\subsubsection{2 primitives (M, mA)}
\label{sec:results1}

We evaluated 10 agents across three datasets by pairing five LLMs (Code World Model \cite{carbonneaux2025cwm}, o3-mini, gpt-oss-20b, gpt-oss-120b, and GPT-4o) with one-shot and greedy scaffolds. This yielded 150 runs per dataset (50 greedy and 100 one-shot). We further augmented the search with 20 additional greedy GPT-5 runs exclusively on the MAD dataset. Since agents only need to output a formatted string that maps to predefined PyTorch blocks, we deliberately employed non-SOTA models, which are fully capable of handling this structural simplicity. The search yielded 2,307 unique architectures explored, probing $3.17\%$ of the search space. See Appendix \ref{app:additional}, Figures~\ref{fig:valtrajectory}--\ref{fig:trajectories_2prim_avg} for the search dynamic per agent. We rank all designed models across agents and seeds and rank them on their test accuracy (see Appendix~\ref{app:ranking}). We then aggregate them into robust base patterns and scale them. This resulted in 6 agent-discovered architectures (base 16-layer pattern in parentheses):

\begin{itemize}
    \item \textbf{AIRAformer-A} (\texttt{(A + M) + (2A + 2M) + 4 $\times$ (A + M) + 2M}): The aggregated 16-layer architecture is \textbf{stretched} to large scale.

    \item \textbf{AIRAformer-B} (\texttt{2A + 5 $\times$ (A + M) + 4M}): evaluated in its \textbf{stacked} variant.

    \item \textbf{AIRAformer-C} (\texttt{(2A + M) + 3 $\times$ (A + M) + (2A + M) + 4A)}: evaluated in its \textbf{stacked} and \textbf{stretched} variants.

    \item \textbf{AIRAformer-D} (\texttt{5 $\times$ (2A + M) + A}): evaluated in its \textbf{stacked} and \textbf{stretched} variants.
\end{itemize}

Aggregation techniques to obtain them and their composition at 350M, 1B, and 3B scale are shown in Appendix~\ref{app:scaling}. Notably, despite different aggregation techniques, these models converge on shared attention-to-MLP ratios: 7:9 for AIRAformers A and B, and 11:5 for C and D.

\textbf{IsoToken Analysis}. We pretrain 9 models: 6 AIRAformers, Llama 3.2 baseline and 2 Composer-found models (Composite Stacked and Stretched, which are the best Composer-found models in 2-primitive space) at the 1B scale for a fixed, predefined token budget corresponding to 71,565 steps, derived from \cite{acun2025composer}. With a batch size of 4, sequence length of 8,192, and DP of 16, this corresponds to  $\sim$37.5 billion tokens, or $\sim38$ TPP. Each architecture is trained with 3 random seeds, and we report the mean performance across them. We report the 3 metrics for the 9 pretrained models in Table~\ref{tab:pretraining_results}. A comprehensive breakdown is given in Appendix~\ref{app:additional}, Table~\ref{tab:dclm_pertask}, Figure~\ref{fig:downstream_eval}.

\begin{table*}[!htbp]
\centering
\caption{Pretraining results of 2-primitive architectures at 1B scale with fixed token budget (37.5B tokens). We report validation loss, 0-shot accuracy on 6 downstream tasks (in $\%$) and few-shots DCLM score on 14 downstream tasks (in $\%$), averaged over 3 seeds ($\pm$std). Values in parentheses show normalized accuracy where applicable. Best results in \textbf{bold}, second best \underline{underlined}.}
\label{tab:pretraining_results}
\resizebox{\linewidth}{!}{%
\begin{tabular}{lccccccccc}
\toprule
\textbf{Architecture} & \textbf{Val Loss $\downarrow$} & \textbf{ARC-C} & \textbf{ARC-E} & \textbf{HellaS.} & \textbf{PIQA} & \textbf{SciQ} & \textbf{WinoG.} & \textbf{Avg $\uparrow$} & \textbf{DCLM Core Score $\uparrow$} \\
\midrule

Llama 3.2 & 2.815{\tiny$\pm$.003} & 26.1 (30.0) & 62.3 (56.0) & 41.4 (53.1) & 72.2 (72.5) & 87.2 (80.2) & 56.0 & 57.5 (58.4) & 46.9{\tiny$\pm$0.3}\\

Composite (St.) & 2.782{\tiny$\pm$.021} & 27.7 (30.2) & 62.9 (57.1) & 42.4 (55.1) & 72.2 (72.3) & 88.0 (82.7) & 57.3 & 58.4 (59.5) & 46.6{\tiny$\pm$1.4} \\
Composite (Str.) & 2.759{\tiny$\pm$.002} & 28.1 (31.0) & 64.0 (58.5) & 42.6 (54.8) & 72.2 (72.6) & 88.1 (81.8) & 55.3 & 58.4 (59.8) & 47.3{\tiny$\pm$0.1} \\
\midrule

AIRAf.-A (Str.) & 2.752{\tiny$\pm$.002} & \textbf{29.8} (\textbf{32.2}) & \textbf{65.4} (\textbf{59.0}) & 43.2 (56.0) & \underline{72.6} (\textbf{72.9}) & 88.0 (82.7) & \underline{58.4} & \underline{59.6} (\underline{60.6}) & 48.5{\tiny$\pm$0.2} \\
AIRAf.-B (St.) & 2.752{\tiny$\pm$.001} & \underline{29.4} (31.3) & \underline{65.3} (\textbf{59.0}) & 43.2 (\underline{56.3}) & \textbf{72.9} (\textbf{72.9}) & 88.6 (\underline{83.7}) & 57.9 & 59.5 (\underline{60.6}) & 48.1{\tiny$\pm$0.1} \\
AIRAf.-C (St.) & 2.751{\tiny$\pm$.002} & 28.4 (31.4) & 63.5 (57.5) & 42.9 (55.3) & 72.5 (\underline{72.7}) & 89.3 (83.6) & 58.3 & 59.1 (60.1) & \underline{48.8}{\tiny$\pm$0.2} \\
AIRAf.-C (Str.) & 2.750{\tiny$\pm$.001} & 29.1 (31.1) & 64.0 (57.9) & 42.9 (55.4) & 71.8 (72.3) & \textbf{89.5} (83.5) & 56.7 & 59.0 (60.1) & 48.1{\tiny$\pm$0.2} \\
AIRAf.-D (St.) & \underline{2.745}{\tiny$\pm$.002} & 28.4 (\underline{31.6}) & 63.1 (\underline{58.9}) & \underline{43.3} (56.2) & 72.4 (72.5) & 88.7 (83.1) & 58.1 & 59.0 (60.5) & 48.4{\tiny$\pm$0.2} \\
AIRAf.-D (Str.) & \textbf{2.734}{\tiny$\pm$.001} & \underline{29.4} (31.5) & 63.7 (\textbf{59.0}) & \textbf{43.7} (\textbf{56.4}) & \textbf{72.9} (72.4) & \underline{89.4} (\textbf{84.5}) & \textbf{58.9} & \textbf{59.7} (\textbf{60.8}) & \textbf{48.9}{\tiny$\pm$0.4} \\
\bottomrule
\end{tabular}%
}
\end{table*}

AIRAformers outperform the Composite baselines across both validation loss and downstream tasks performance. AIRAformer-D Stretched achieves the lowest validation loss of 2.734, highest average 0-shot accuracy of 59.7\% (60.8\%) and highest DCLM Core score (48.9\%), outperforming both Llama 3.2 and Composer-found models. The low variance across seeds indicates that the observed improvements are statistically robust.

\textbf{IsoFLOP Analysis}. We pretrain the 8 models (excluding Llama 3.2) across three parameter scales (350M, 1B, and 3B) under five FLOPs budgets: $2\times10^{19}$, $4\times10^{19}$, $8\times10^{19}$, $2\times10^{20}$, and $4\times10^{20}$ FLOPs, for a total of 120 experiments (see Appendix~\ref{app:scaling}, Table~\ref{tab:iso_flop_steps} for FLOPs count). Attention-heavy AIRAformer-C and AIRAformer-D architectures achieve superior validation loss in the fixed-token-budget setting, as expected (Table~\ref{tab:pretraining_results}), while balanced architectures are more compute-efficient in the isoFLOP regime. The additional attention layers in AIRAformer-C/D provide representational benefits when trained for longer, but they are less efficient when compute is the binding constraint.

\begin{figure}[!htbp]
    \centering
    \includegraphics[width=\linewidth]{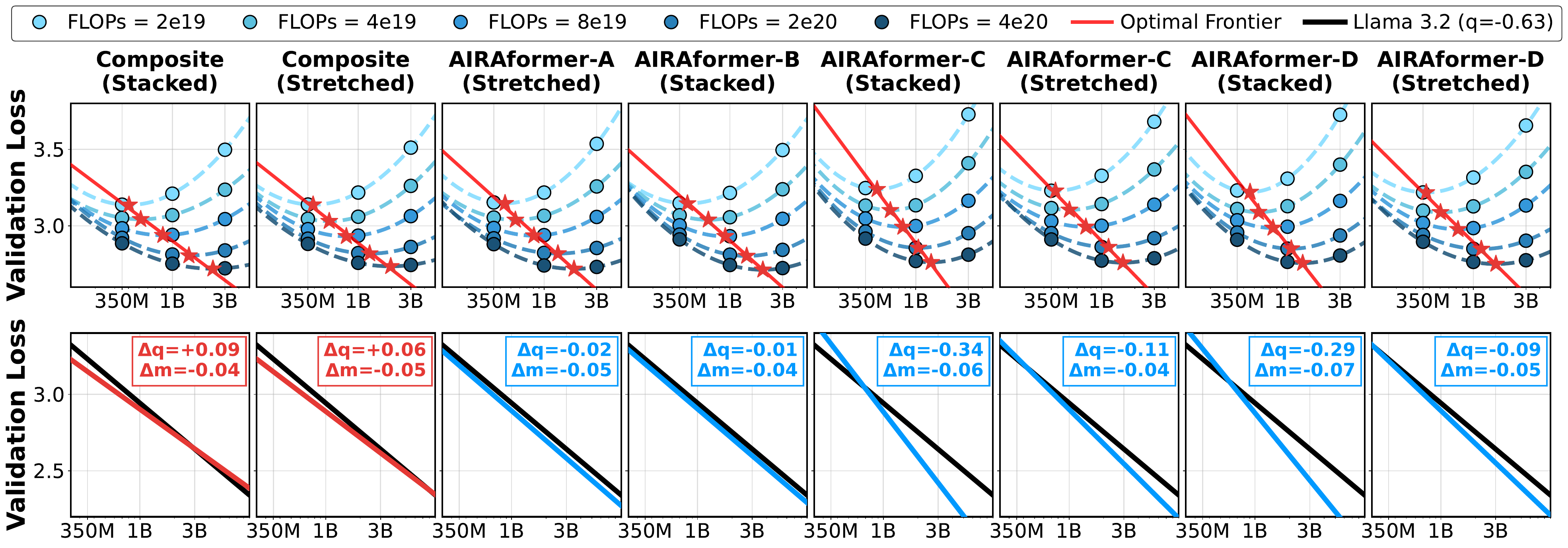}
    \caption{\textbf{IsoFLOP scaling curves and optimal frontier comparison (M, mA).} \textbf{Top:} Validation loss versus model size for each architecture across 5 FLOPs budgets. \textbf{Bottom:} Comparison of each architecture's optimal frontier against the Llama 3.2 scaling law derived from \citep{acun2025composer}.}
    \label{fig:vallossisoflopcurves}
\end{figure}

We fit isoFLOP parabolas to the validation loss as a function of model size for each FLOP budget (Figure~\ref{fig:vallossisoflopcurves}, top) and infer the optimal frontier from their minima. Figure~\ref{fig:vallossisoflopcurves}, bottom, compares each architecture's optimal frontier against Llama 3.2's reported scaling law, and reports $\Delta q$ (slope difference, lower is better) and $\Delta m$ (intercept offset, lower is better) for each architecture with respect to it. AIRAformer-C Stacked and AIRAformer-D Stacked achieve the most favorable scaling coefficients, indicating steeper scaling than Llama 3.2. The Composite baselines exhibit slightly flatter slopes than Llama 3.2, suggesting diminishing advantage as model size increases.

\subsubsection{3 primitives (M, mA, Mb)}
\label{sec:results2}

We use MAD for the 3 primitives search due to better agreement between the small scale ranking and large scale performance.  A total of 170 agentic runs yielded 2,248 unique architectures, probing $0.0052\%$ of the search space. Agent-powered AIRA-Compose discovers robust patterns by probing just a fraction of the design space, mitigating the combinatorial explosion of 3-primitive spaces in traditional NAS. We obtain 8 hybrid architectures:

\begin{itemize}
    \item \textbf{AIRAhybrid-A} (\texttt{2Mb + M + 11Mb + 2M}): evaluated in its \textbf{stretched} variant.

    \item \textbf{AIRAhybrid-B} (\texttt{3 $\times$ (2Mb + M + A) + 2Mb + 2M}): evaluated in \textbf{stacked} and \textbf{stretched} variants.

    \item \textbf{AIRAhybrid-C} (\texttt{2Mb + A + 2 $\times$ (Mb + A) + M + 2 $\times$ (A + Mb + A + M)}): evaluated in its \textbf{stretched} variant.

    \item \textbf{AIRAhybrid-D} (\texttt{2Mb + M + 2 $\times$ (Mb + M) + A + M + Mb + M + A + M + Mb + M + A}): evaluated in \textbf{stacked} and \textbf{stretched} variants.

    \item \textbf{AIRAhybrid-E} (\texttt{5 $\times$ (Mb + M + A) + M}): evaluated in \textbf{stacked} and \textbf{stretched} variants.
\end{itemize}

\begin{table*}[!htbp]
\centering
\caption{Pretraining results of 3-primitive hybrid architectures at 1B scale with fixed token budget (37.5B tokens). We report validation loss, 0-shot accuracy on 6 downstream tasks (in $\%$), and few-shots DCLM score on 14 downstream tasks (in $\%$). Values in parentheses show normalized accuracy where applicable. Best results in \textbf{bold}, second best \underline{underlined}. Results are from a single seed.}
\label{tab:pretraining_results_hybrid_bolded}
\resizebox{\linewidth}{!}{%
\begin{tabular}{lccccccccc}
\toprule
\textbf{Architecture} & \textbf{Val Loss $\downarrow$} & \textbf{ARC-C} & \textbf{ARC-E} & \textbf{HellaS.} & \textbf{PIQA} & \textbf{SciQ} & \textbf{WinoG.} & \textbf{Avg $\uparrow$} & \textbf{DCLM Core Score $\uparrow$} \\
\midrule

Nemotron-H Approx. & 2.741 & 29.1 (31.3) & 65.3 (59.9) & 43.4 (56.3) & 73.3 (73.7) & 87.8 (82.7) & 56.4 & 59.2 (60.1) & 48.5 \\
Nemotron-2 Approx. & 2.732 & \underline{30.7} (33.2) & 65.9 (\textbf{61.9}) & \underline{44.2} (57.2) & \textbf{74.0} (73.9) & 88.7 (84.3) & 58.5 & \underline{60.3} (61.5) & 48.4 \\
Mamba (Mb + M) & 2.771 & 30.1 (32.5) & 64.9 (61.2) & 43.2 (55.8) & 72.9 (73.1) & 88.8 (85.1) & 57.7 & 59.6 (60.9) & 46.1 \\

Composer (2Mb-M-3A) & 2.724 & 30.0 (33.1) & 65.0 (\underline{61.6}) & 43.9 (56.9) & 71.9 (73.9) & 89.6 (85.3) & 58.8 & 59.9 (61.6) & \textbf{49.3} \\
\midrule

AIRAhybrid-A (Str.) & 2.750 & 30.5 (\textbf{33.5}) & \textbf{66.6} (61.5) & 44.0 (56.6) & \underline{73.9} (73.4) & 89.7 (85.2) & 57.2 & \underline{60.3} (\underline{62.1}) & 47.5 \\
AIRAhybrid-B (St.) & 2.732 & 29.9 (\underline{33.4}) & 64.8 (61.2) & 44.0 (57.1) & 73.1 (73.4) & \underline{89.9} (84.1) & 58.9 & 60.1 (61.8) & 49.0 \\
AIRAhybrid-B (Str.) & 2.728 & 30.0 (31.1) & 65.1 (61.2) & \underline{44.2} (\underline{57.6}) & 73.4 (\underline{74.0}) & 89.5 (\underline{85.7}) & \underline{59.2} & 60.2 (61.9) & \underline{49.1} \\
AIRAhybrid-C (Str.) & 2.740 & 30.2 (32.2) & 65.0 (58.5) & 43.3 (56.1) & 72.8 (73.6) & \textbf{90.2} (85.4) & 58.0 & 59.9 (61.2) & 48.9 \\
AIRAhybrid-D (St.) & \underline{2.720} & 29.5 (30.9) & 65.1 (60.0) & \textbf{44.3} (\underline{57.6}) & 72.8 (73.7) & \underline{89.9} (\textbf{85.8}) & \textbf{59.8} & 60.2 (61.6) & 48.4 \\
AIRAhybrid-D (Str.) & \textbf{2.719} & \textbf{32.0} (32.9) & \underline{66.3} (61.4) & 44.1 (\textbf{57.8}) & 73.6 (\textbf{74.7}) & 88.6 (84.2) & 58.2 & \textbf{60.5} (\textbf{62.2}) & 48.5 \\
AIRAhybrid-E (St.) & 2.736 & 27.7 (30.3) & 64.3 (59.5) & 43.5 (56.5) & 72.1 (73.3) & 88.9 (85.3) & 58.6 & 59.2 (61.0) & 48.8 \\
AIRAhybrid-E (Str.) & 2.732 & 29.8 (32.5) & 64.9 (60.4) & 44.1 (57.1) & 72.3 (72.9) & 89.8 (\textbf{85.8}) & 57.7 & 59.8 (61.7) & 48.6 \\

\bottomrule
\end{tabular}%
}
\end{table*}

\textbf{IsoToken Analysis}. We pretrain 12 models: 8 AIRAhybrids, 3 hybrid baselines (Mamba, Nemotron-H and Nemotron-2) and one Composer-found architecture at the 1B scale at $\sim 38$ TPP.  Note that Nemotron-H and Nemotron-2 are only \textit{approximated} versions to smaller scale, with MoEs replaced by MLP layers. The Composer architecture is obtained by performing the search over 6-layer small scale models, which is then stacked to obtain an equivalent model at large scale. Table~\ref{tab:pretraining_results_hybrid_bolded} summarizes the validation loss and downstream task performance. AIRAhybrids often outperform established baselines, including Mamba and approximated Nemotron-2, across both linguistic and reasoning benchmarks. AIRAhybrid-D (Stretched) achieves the lowest validation loss of 2.719 and the highest 0-shot accuracy of 60.5$\%$. While specific architectures show localized strengths (e.g., AIRAhybrid-C Stretched on SciQ and Nemotron-2 on PIQA), AIRAhybrids generally maintains superior average 0-shot accuracy.

\begin{figure}[H]
    \centering
    \includegraphics[width=\linewidth]{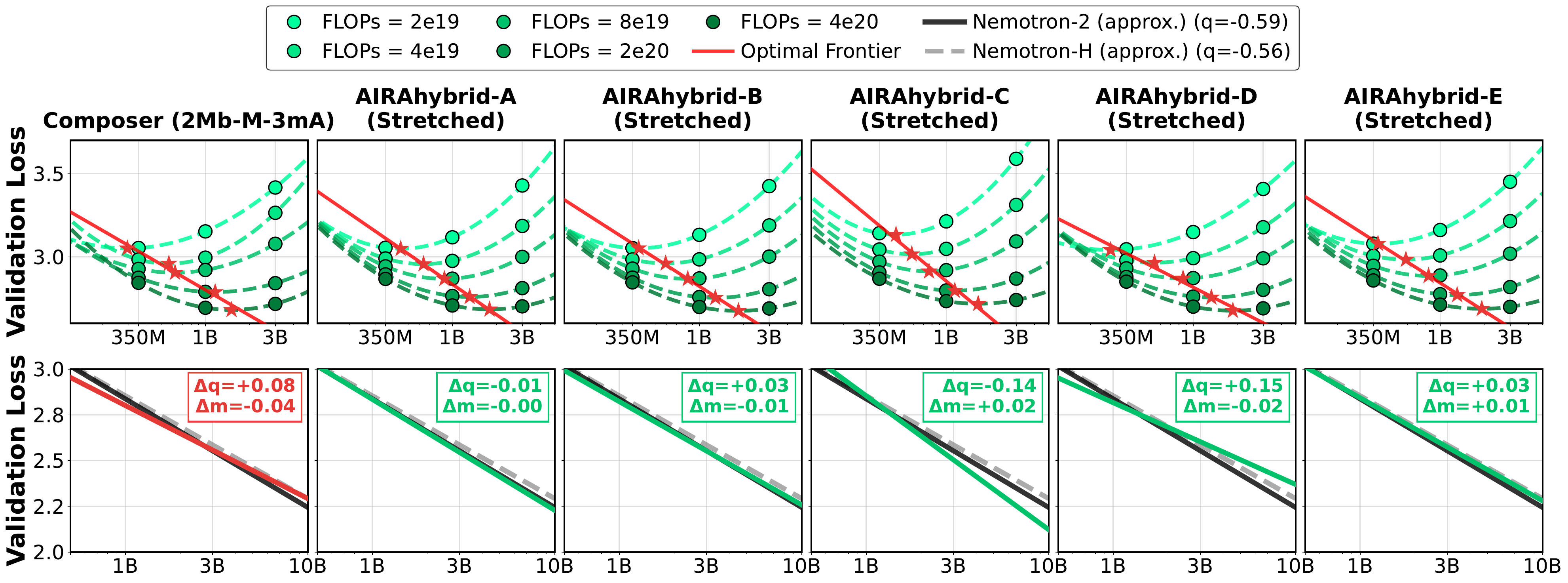}
    \caption{\textbf{IsoFLOP scaling curves and optimal frontier comparison (M, mA, Mb).} \textbf{Top:} Validation loss versus model size for each architecture across 5 FLOPs budgets. \textbf{Bottom:} Comparison of each architecture's optimal frontier against the approximated Nemotron-2 and Nemotron-H.}
    \label{fig:mb_isoflop}
\end{figure}

\textbf{IsoFLOP Analysis}. We pretrain the 11 models (excluding Mamba due to poor isotoken performance) at 3 scales, 350M, 1B and 3B parameters under 5 FLOPs budgets, for a total of 165 experiments. Results for selected models are shown in Figure~\ref{fig:mb_isoflop} (see Appendix~\ref{app:additional} for full results).

We compare each architecture's optimal frontier against the approximated Nemotron-H and Nemotron-2's scaling law in Figure~\ref{fig:mb_isoflop}, reporting $\Delta q$ (slope difference) and $\Delta m$ (intercept offset) relative to Nemotron-2. Architectures with balanced Mamba-to-attention ratios (Nemotron-2, AIRAhybrid-B Stretched and AIRAhybrid-D Stretched) consistently achieve the lowest validation loss across model sizes and FLOP budgets, while the attention-heavy AIRAhybrid-C Stretched ranks last.
However, AIRAhybrid-C Stretched also achieves the steepest scaling slope ($\Delta q = -0.14$), indicating faster improvement with scale than Nemotron-2, despite its higher intercept. AIRAhybrid-A~Stretched, the only pure-SSM design, also scales slightly better ($\Delta q = -0.01$). The Composer baseline, on the other hand, exhibits a flatter slope. These results mirror the two-primitive behavior: attention-heavy architectures are less compute-efficient at fixed FLOP budgets but exhibit steeper scaling slopes that may yield superior performance at larger scale.

\begin{figure}[H]
    \centering
    \includegraphics[width=\linewidth]{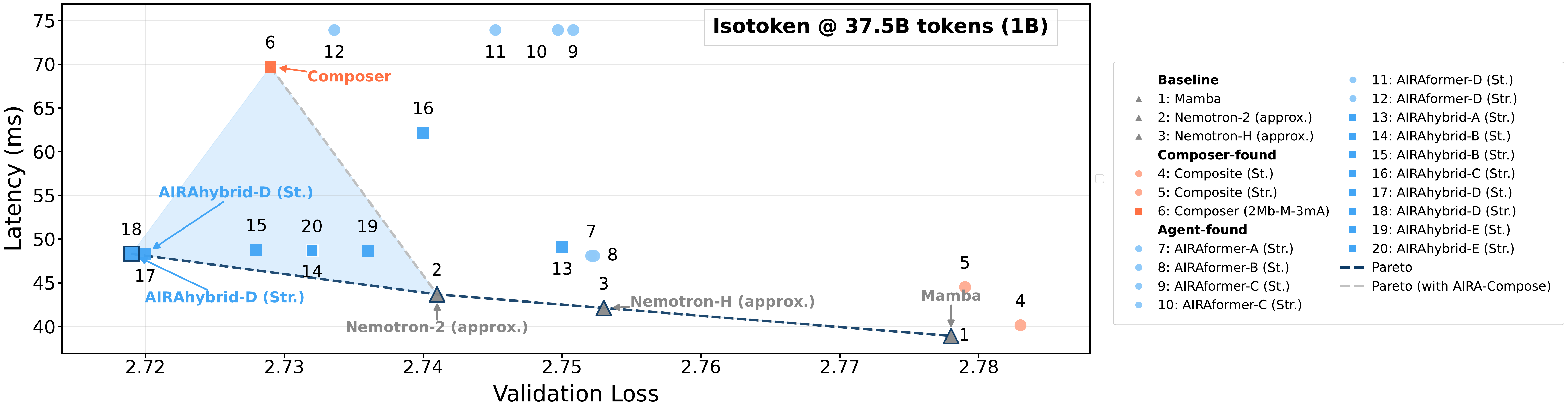}
    \caption{\textbf{Latency--validation loss Pareto frontiers at the 1B-parameter scale}. Isotoken comparison (37.5B). Total latency is the sum of per-block latencies across all layers. Dashed lines indicate Pareto frontiers with and without agent-discovered architectures; the shaded region highlights the gap between the two.}
    \label{fig:latency_pareto}
\end{figure}

Finally, Figure~\ref{fig:latency_pareto} presents the 1B-scale latency -- validation loss Pareto frontiers under isoToken and isoFLOP protocols. Model latency is computed by scaling the number of layers per block type by its corresponding GPU(H100)-profiled block latency. Under the isoToken analysis,
AIRAhybrids-D (Stacked):17 and AIRAhybrids-D (Stretched):18 advance the previously-achieved frontier of Composer~\cite{acun2025composer}, leading to lower validation loss at strictly lower latencies than their Composer counterpart, which compensates strong downstream task performance with high latency.

\subsection{AIRA-Design}

\subsubsection{Long Range Arena}
\label{sec:results3}

We evaluated 20 agents across 12 unique models: all 12 in greedy mode (GPT-5, o3, o3-mini, GPT-4o, CWM, gpt-oss-120b, gpt-oss-20b, Devstral-Small 24B, Opus 4.6, Opus 4.5, Gemini 3.0 Pro, and Gemini 3.1 Pro) and 8 only in one-shot mode (CWM, o3, o3-mini, gpt-oss-20b, gpt-oss-120b, GPT-4o, Devstral-Small 24B, and GPT-5). The 12 greedy agents were run for 10 seeds across three datasets and two setups (Configurable and Non-Configurable), totaling 720 runs. The 8 one-shot agents were run for 20 seeds each, totaling 960 runs. In total, we aggregated results from 1,680 runs, representing a few thousands of architectural primitives autonomously designed, implemented, and validated.

\begin{figure}[!htbp]
    \centering
    \includegraphics[width=\linewidth]{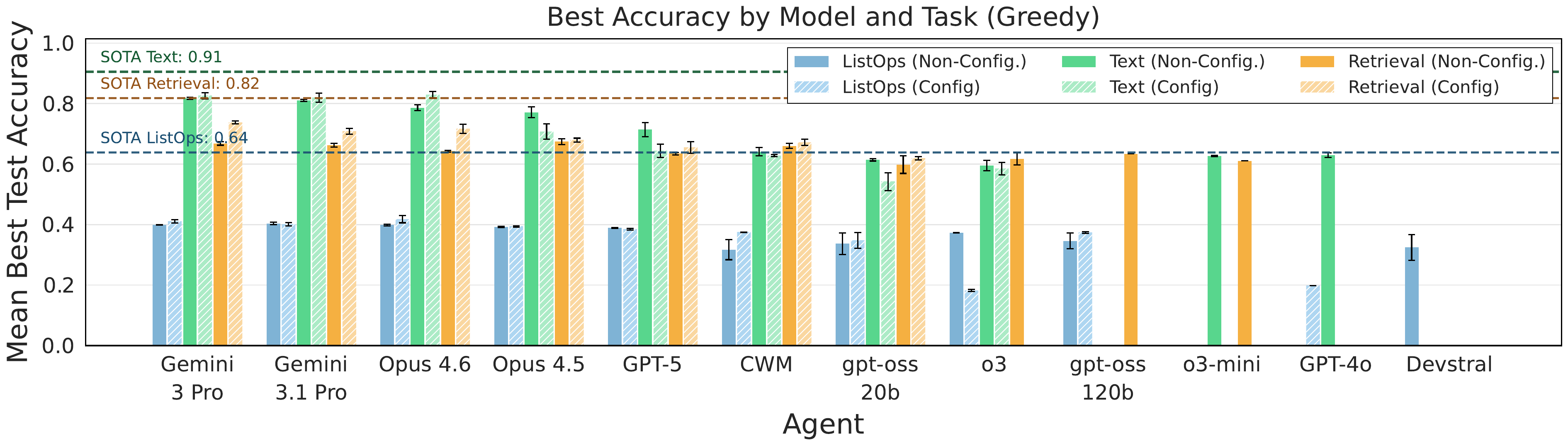}
    \vspace{-1.7em}
    \caption{\textbf{Mean best test accuracy} achieved by the 8 greedy agents across the 6 low-level mechanistic design tasks. Solid bars represent performance using the original LRA benchmark configuration, while striped bars indicate results from the Configurable LRA setup that permits agent-specified hyperparameters. Error bars represent standard error of the mean across 10 random seeds per task. Horizontal dashed lines indicate SOTA performance over the 3 datasets: ListOps (0.64), Text (0.91), and Retrieval (0.82). Agents are ordered by decreasing average performance. }
    \label{fig:lrafullbest}
\end{figure}

Figure \ref{fig:lrafullbest} shows the average best test accuracy per model and agent. It captures the peak accuracy achieved during exploration, even if the agent ultimately failed to submit that model as a valid final solution. Greedy Gemini 3 Pro achieved the highest mean best accuracy across nearly all task-configuration combinations, followed by Greedy Gemini 3.1 Pro and Opus 4.6.

\begin{figure}[!htbp]
    \centering

    \begin{subfigure}{\linewidth}
        \centering
        \includegraphics[width=\linewidth]{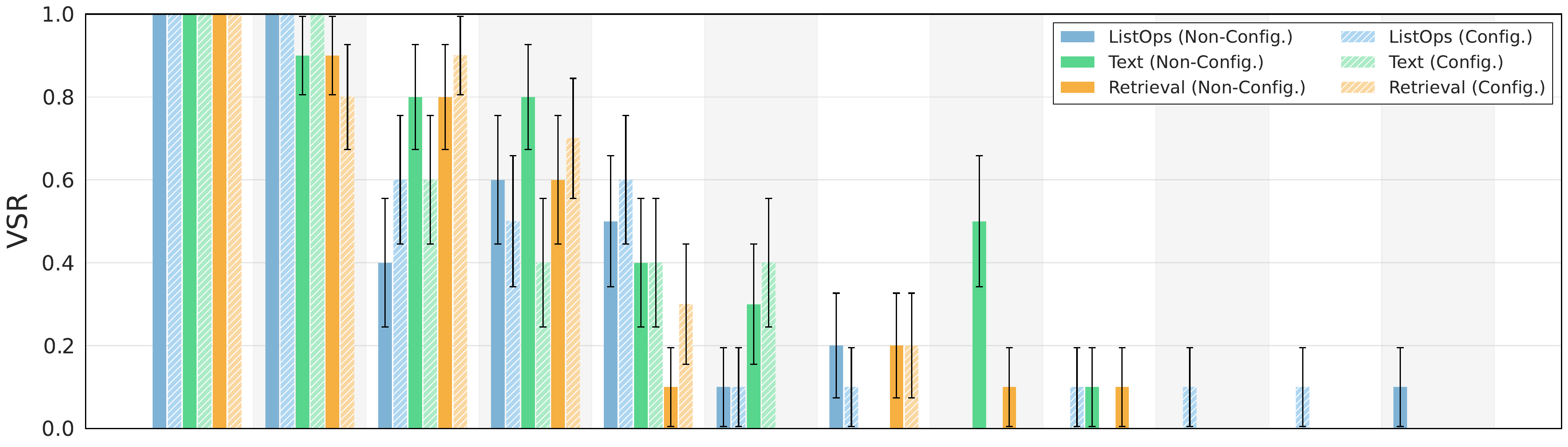}
    \end{subfigure}

    \vspace{-0.40em}

    \begin{subfigure}{\linewidth}
        \centering
        \includegraphics[width=\linewidth]{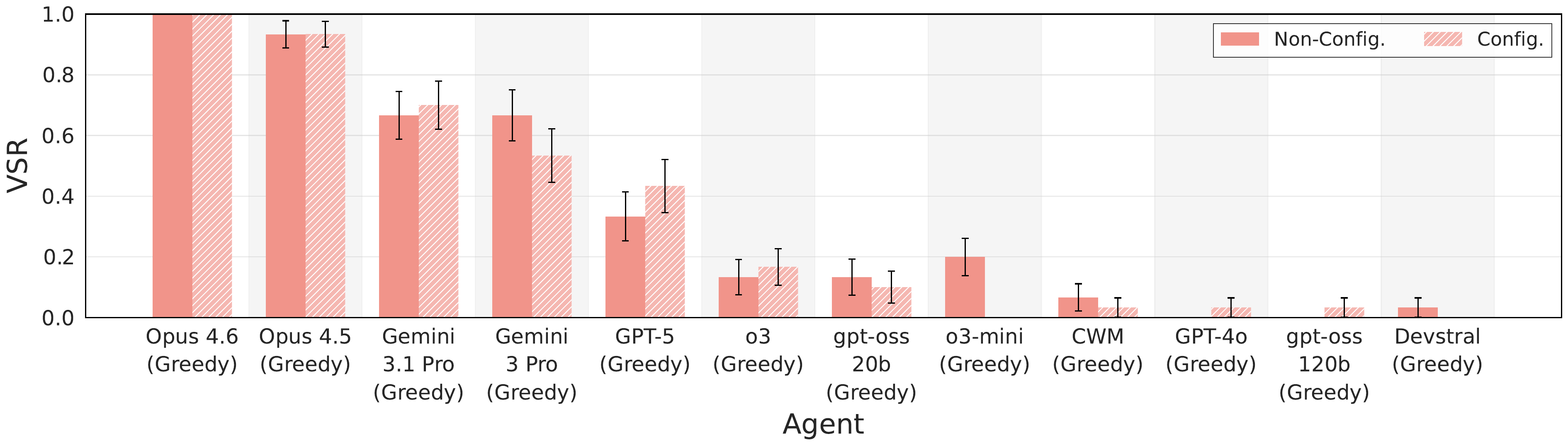}
    \end{subfigure}
    \vspace{-2.3em} 
    \caption{\textbf{Valid Submission Rate (VSR)} across agents and setups. Top: per-task VSR showing the proportion of runs that produced valid submissions (i.e., final accuracy > 0) for each of the 6 task configurations. Bottom: average VSR across 3 tasks, where missing tasks contribute 0 to the average. Solid bars represent the original LRA benchmark configuration; striped bars indicate the Configurable LRA setup. Error bars represent standard error of the mean. Agents are ordered by decreasing average performance. No valid submission was achieved by one-shot agents.}
    \label{fig:lra_combined_vsr}
\end{figure}

Greedy Gemini 3 Pro and Greedy Opus 4.6 generated the best solutions for 2 of the 6 task configurations each, while Greedy Gemini 3.1 Pro and Greedy Opus 4.5 achieved the best accuracy in 1 task each. A summary of each best solution for the 6 tasks provided in Appendix~\ref{app:configurable}, Table \ref{tab:lra_summaries}. Agents produced functional solutions incorporating established efficient attention mechanisms such as linear attention with ELU+1 kernels, hierarchical pooling strategies, and blockwise local attention with recurrent memory. Several solutions integrate known architectural components including depthwise convolutions and gated linear units. While these designs demonstrate competent engineering (combining existing techniques in task-appropriate ways) they do not represent fundamentally novel contributions to the efficient attention literature. The discovered architectures largely recombine and adapt ideas from prior work (e.g., Performer, Longformer, Conformer) rather than introducing new theoretical insights. This suggests that current agents excel at engineering-level synthesis and adaptation, but fall short of genuine scientific innovation in mechanistic design.

Figure \ref{fig:lra_combined_vsr} shows the Valid Submission Rate (VSR), i.e., the proportion of runs yielding working, validated solutions. VSR varies significantly, with Greedy Opus 4.6 always successfully submitting a solution, to below 10\% for weaker models. VSR was generally higher in the Non-Configurable setup, indicating that the expanded hyperparameter search space in the Configurable setup increases the difficulty of producing valid code. Notably, one-shot agents failed to produce any valid submissions. This confirms that single-turn code generation is insufficient for mechanistic design, and that iterative refinement, debugging, and validation are essential.

\begin{figure}[!htbp]
    \centering

    \begin{subfigure}{\linewidth}
        \centering
        \includegraphics[width=\linewidth]{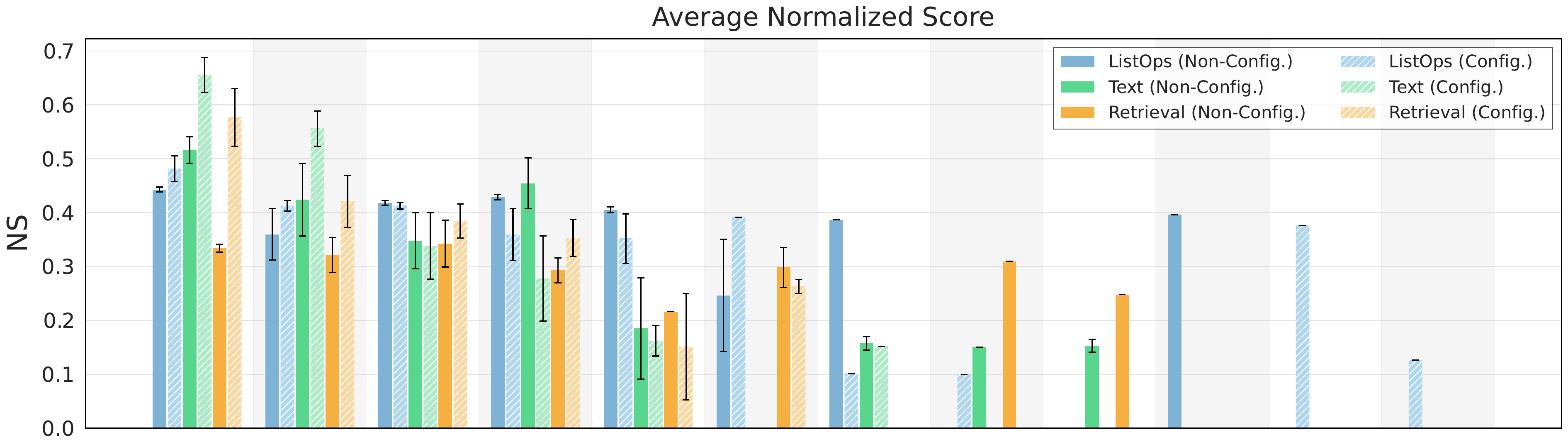}
    \end{subfigure}

    \vspace{-0.40em} 

    \begin{subfigure}{\linewidth}
        \centering
        \includegraphics[width=\linewidth]{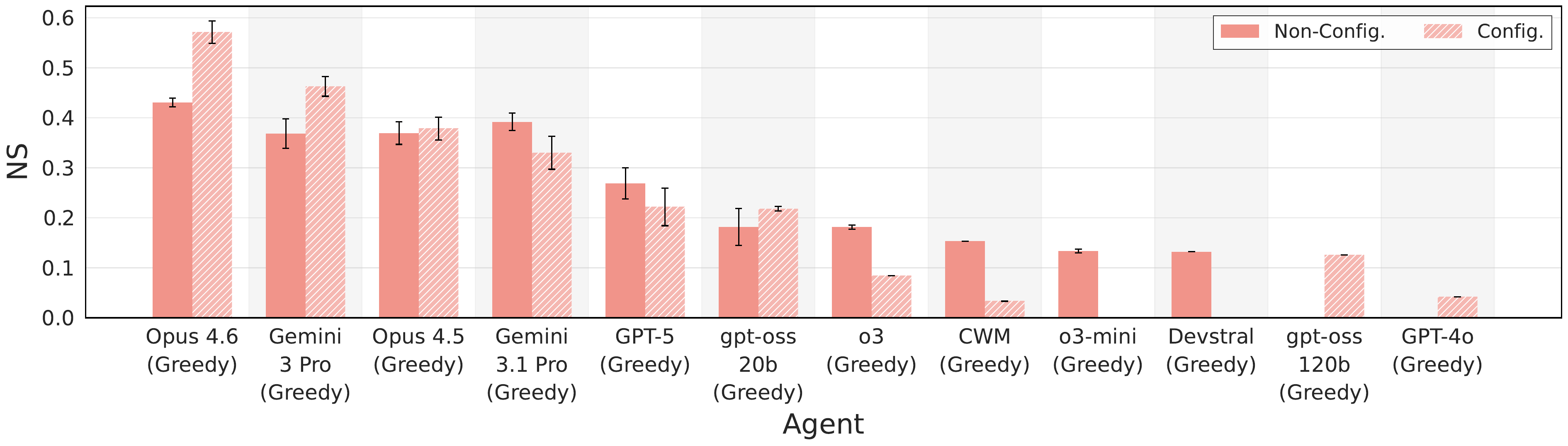}
    \end{subfigure}
    \vspace{-2.3em} 
    \caption{\textbf{Normalized Score (NS)} using the march of 9s across agents and setups. Top: per-task NS computed only over valid submissions. Bottom: average NS across 3 tasks, where missing tasks contribute 0 to the average. Solid bars represent the original LRA benchmark configuration; striped bars indicate the Configurable LRA setup. Error bars represent standard error of the mean. Agents are ordered by decreasing average performance. No valid submission was achieved by one-shot agents.}
    \label{fig:lra_combined_scores}
\end{figure}

Figure~\ref{fig:lra_combined_scores} and Table~\ref{tab:avg_ns_per_task} summarize the average Normalized Score (NS) per agent and per task, respectively. Among greedy agents in the non-configurable setup, ListOps emerges as the most tractable task with an average NS of 0.257 $\pm$ 0.054, while Text classification in the configurable setup remains the most challenging (NS = 0.178 $\pm$ 0.064). The configurable setup generally yields lower NS values across all tasks despite offering greater architectural flexibility. This performance gap suggests that current agents struggle to effectively leverage hyperparameter optimization and often fail to navigate the larger search space. Even top-performing agents achieve NS values well below 1.0, showing a substantial gap between agentic design and human SOTA performance.

 \begin{table}[H]
    \centering
    \caption{Average Normalized Score per task. Agents with no valid submissions contribute 0 to the average. Left: greedy agents only ($n=12$). Right: all agents including one-shot ($n=20$).}
    \label{tab:avg_ns_per_task}
    \begin{tabular}{lcccc}
        \toprule
        & \multicolumn{2}{c}{\textbf{Greedy Agents} ($n=12$)} & \multicolumn{2}{c}{\textbf{One-Shot + Greedy Agents} ($n=20$)} \\
        \cmidrule(lr){2-3} \cmidrule(lr){4-5}
        \textbf{Task} & \textbf{Non-Config.} & \textbf{Config.} & \textbf{Non-Config.} & \textbf{Config.} \\
        \midrule
        ListOps   & 0.257 $\pm$ 0.054 & 0.259 $\pm$ 0.049 & 0.154 $\pm$ 0.043 & 0.156 $\pm$ 0.041 \\
        Text      & 0.199 $\pm$ 0.053 & 0.178 $\pm$ 0.064 & 0.119 $\pm$ 0.039 & 0.107 $\pm$ 0.043 \\
        Retrieval & 0.197 $\pm$ 0.041 & 0.179 $\pm$ 0.058 & 0.118 $\pm$ 0.033 & 0.107 $\pm$ 0.040 \\
        \bottomrule
    \end{tabular}
\end{table}

Since the six mechanistic design tasks follow the \airsbench{}~task structure, we can benchmark agent performance against the broader suite. The 20 tasks in \airsbench{}~are classified into four difficulty bands: \textit{easy}, \textit{medium}, \textit{hard}, and \textit{expert}, based on the normalized scores achieved across agents. When averaged across all 20 agents, all LRA tasks fall into the \textit{hard} category ($0.10 \le \text{NS} < 0.20$). This is in line with other code generation tasks in the \airsbench{}~suite. The challenging nature of LRA tasks stems from four key factors: (i) a training bias toward PyTorch, which limits LLM proficiency in JAX/Flax; (ii) the inherent complexity of writing low-level, functionally complete attention mechanisms from scratch; (iii) fixed compute budgets that penalize inefficient implementations; and (iv) frequent out-of-memory (OOM) errors and numerical instabilities that invalidate theoretically sound models during the compilation or training phase.

\begin{figure}[!htbp]
    \centering
    \includegraphics[width=\linewidth]{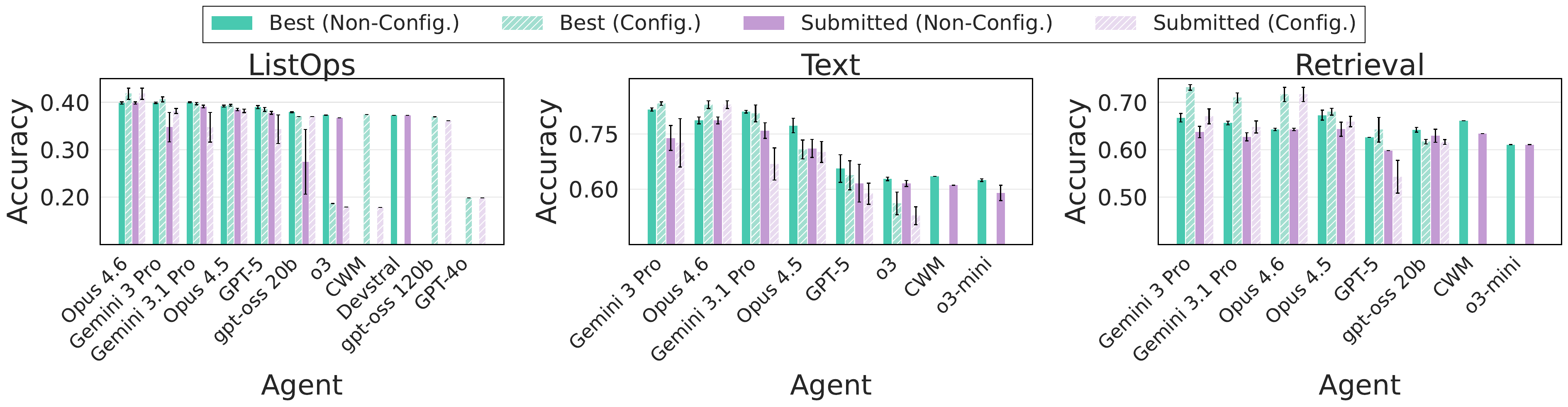}
    \vspace{-1.6em}
    \caption{\textbf{The generalization gap}. Comparison of best observed accuracy versus submitted (final) accuracy for greedy agents across the 3 LRA tasks. Each subplot shows results for a single task, with bars grouped by metric type (best vs submitted) and configuration (solid for original LRA, striped for Configurable). Only runs with valid submissions are included. Error bars represent standard error of the mean. Agents are ordered by decreasing best accuracy per task.  The gap between best and submitted accuracy reflects the generalization gap, measuring the agent's ability to detect the best-performing model for final submission.}
    \label{fig:lrafullbestvssub}
\end{figure}

Figure~\ref{fig:lrafullbestvssub} compares the best accuracy achieved during exploration versus the accuracy of the final submitted model. The gap between these metrics, known as generalization gap, measures an agent's ability to identify and select its best-performing solution. Across all tasks, we observe generally small gaps, with submitted accuracy closely tracking best accuracy for most agents. This suggests that agents are competent at recognizing promising solutions during exploration, though occasional larger gaps indicate room for improved validation strategies.

Configurability produces mixed results across the 12 greedy agents: 7 regressed on average while $4$ improved, with Opus 4.6 and gpt-oss-20b gaining the most ($+4.5pp$ each), followed by Gemini $3$ Pro ($+3.0pp$) and Gemini $3.1$ Pro ($+1.8pp$); o3-mini produced no valid configurable runs. The strongest regressions occurred in weaker agents (CWM $-11.5pp$, Devstral $-10.8pp$, o3 $-8.3pp$), suggesting they struggle more with the expanded hyperparameter search space. The impact is task-dependent: on Retrieval, configurability helped in 6 agents, led by Opus 4.6 and gpt-oss-20b ($+7.4pp$ each) and Gemini 3 Pro ($+6.9pp$)—while on Text, results were mixed, with top agents such as Opus 4.6 improving ($+4.3pp$) but weaker agents suffering large drops (CWM $-32.6pp$, o3 $-12.6pp$). ListOps remained largely stable for top agents. Relative rankings were mostly preserved: Gemini $3$ Pro maintained its top position, though Opus 4.6 rose from fourth to second.

\subsubsection{Autoresearch}
\label{sec:results4}

\begin{figure}[H]
    \centering
    \includegraphics[width=\linewidth]{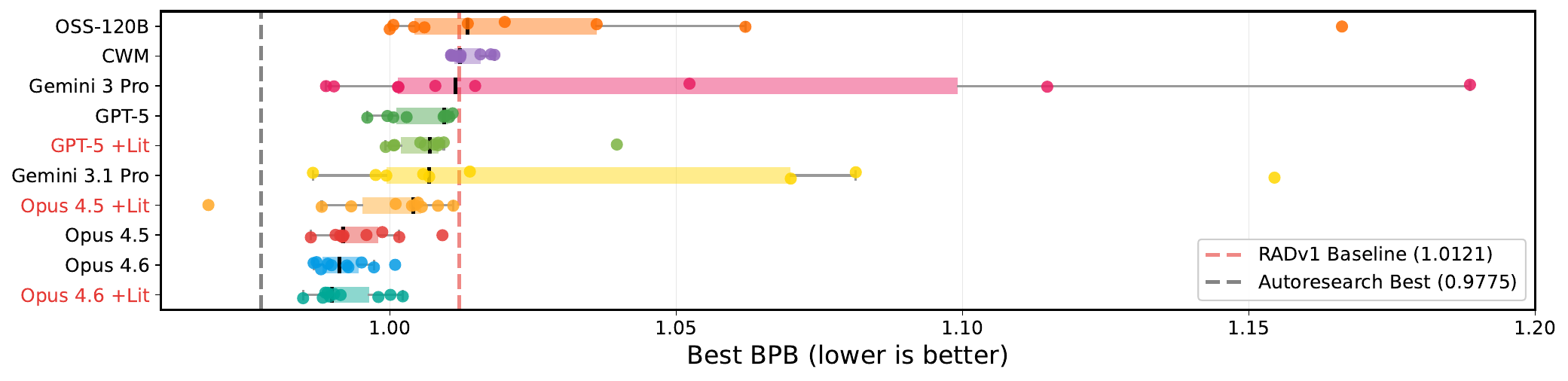}
    \vspace{-0.8cm}
        \caption{\textbf{Autoresearch: distribution of best validation BPB per seed}. Each dot represents one seed's best BPB achieved during its run. Boxes span the interquartile range (Q25--Q75) with bars extending to the most extreme point within $1\times$ IQR. Dashed lines mark the RADv1 (i.e., \airadojo{}) baseline ($\text{BPB}=1.0121$, red) and the Autoresearch best ($\text{BPB}=0.9775$, black). Results are ordered by decreasing median BPB (in black).}
    \label{fig:autoresearch_boxplot}
\end{figure}

We evaluated 7 agents, greedy $\times$ \{CWM, gpt-oss-120b, GPT-5, Opus 4.5, Opus 4.6, Gemini 3.0 Pro, Gemini 3.1 Pro\}. The 7 greedy agents were run for 10 seeds on the single Autoresearch task, for a total of  70 runs. Across all runs, 3{,}936 out of 11{,}530 total steps (34.1\%) yielded a valid numerical fitness value. Additionally, we ran our top-3 agents, greedy GPT-5, Opus 4.5 and Opus 4.6, on the Autoresearch task version enhanced with code and literature, for a total of 30 additional experiments and $\sim$6000 agent-designed training loops. The distribution of the best BPB per seed is shown in Figure~\ref{fig:autoresearch_boxplot}. Both Greedy Gemini~3~Pro and Greedy Gemini~3.1~Pro show the largest variance in their best BPB across seeds, indicating inconsistent optimization: some seeds find strong solutions while others stagnate or crash. By contrast, the Opus~4.5 and Opus~4.6 families cluster tightly below $1.00$, reflecting more reliable convergence. In terms of median performance, Opus~4.6~+Lit (0.990), Opus~4.6 (0.991), and Opus~4.5 (0.992) all perform substantially below the RADv1 baseline. CWM (1.012) and OSS-120B (1.014) have medians near or above the baseline, indicating these agents rarely improve upon the starting script. Notably, adding literature access does not uniformly help: GPT-5~+Lit and Opus~4.5~+Lit underperform compared to their base variant, while Opus~4.6~+Lit slightly improves over Opus~4.6.

The baseline \texttt{train.py} script provided to all agents achieves a validation BPB of 1.0121, shown as a red dashed line. While the model architecture and training pipeline are identical to the original Autoresearch setup, the evaluation environments differ in several ways (see Appendix~\ref{app:autoresearch}, Table~\ref{tab:autoresearchreq}): our agents run on H200 GPUs rather than the H100s used in the original benchmark, and they operate in a sandboxed environment with no internet access and slightly different requirements. These differences could account for the gap in starting BPB ($1.012$ for our agents vs.\ $0.998$ for the Autoresearch work for the same \texttt{train.py} script).

The two setups also differ in how the agent interacts with the code. In the original Autoresearch framework, the agent operates as an interactive coding assistant with full shell access: it reads the current \texttt{train.py}, makes a targeted edit, runs the script, inspects the output, and either commits or reverts the change within a persistent session. This produces smaller but consistent improvements at each step. In \airadojo{} with the greedy scaffold, the agent regenerates the entire \texttt{train.py} from scratch at each step, branching from the current best solution with minimal context on past explorations. Agents therefore introduce multiple changes simultaneously, making it harder to isolate which modifications help. Additionally, task instructions are not always respected---agents may omit required imports or preambles, causing evaluation crashes. This translates into trajectories with fewer improvements per run but characterized by larger deltas.

\begin{figure}[H]
    \centering
    \includegraphics[width=\linewidth]{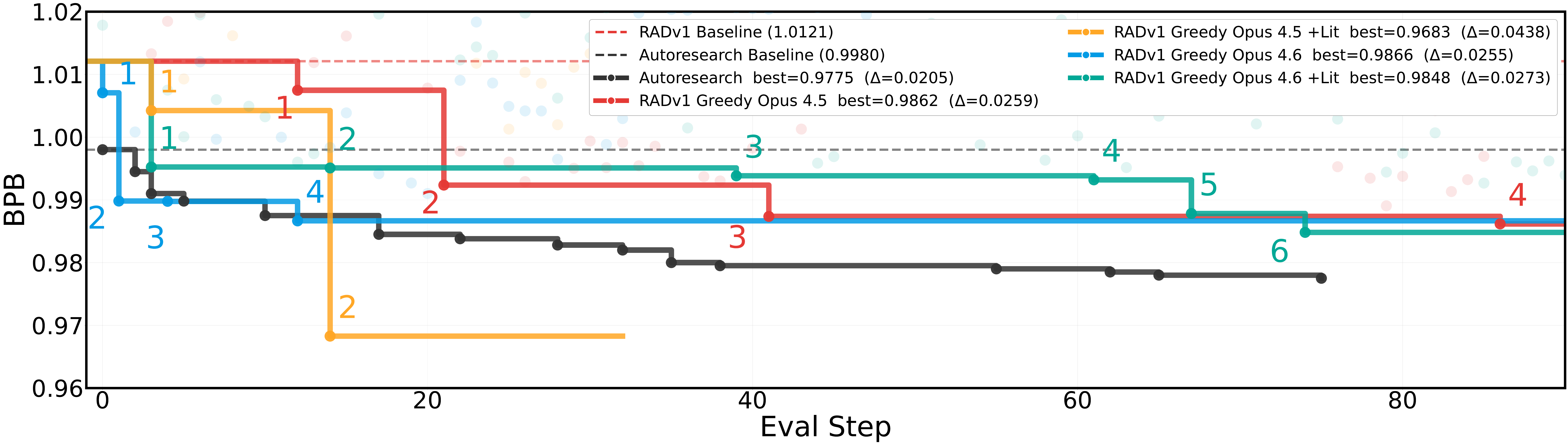}
    \vspace{-0.7cm}
    \caption{\textbf{Autoresearch: best-seed BPB progression for Greedy Opus~4.5 and Greedy Opus~4.6, with and without literature}. Each step line tracks the cumulative best BPB over evaluation steps, with numbered annotations marking improvements. Faded dots indicate non-improving evaluations. The red dashed line marks the RADv1 baseline ($\text{BPB}=1.0121$), while the black dashed line marks the original Autoresearch baseline ($\text{BPB}=0.998$).}
    \label{fig:best_seed_progression}
\end{figure}

Figure~\ref{fig:best_seed_progression} compares the BPB improvement trajectory of the best seed for Opus~4.5 and Opus~4.6, with and without literature access, against the original Autoresearch baseline. We selected the top-performing agents and the seed achieving the lowest final BPB.

Table~\ref{tab:autoresearch_steps} details the improvement steps for each agent run shown in Figure~\ref{fig:best_seed_progression}. Each agent exhibits a distinct optimization strategy. Greedy Opus~4.5 begins by widening the model and removing value embeddings (step~1), then achieves its largest gain ($\Delta=0.015$) by widening further while restoring them (step~2), followed by targeted refinements: sparsifying value embeddings (step~3) and adding learnable attention output scaling (step~4). Greedy Opus~4.6 starts with multiple simultaneous changes---deeper model, smaller head dimension, different window pattern (step~1)---then partially reverts several of these while increasing weight decay (step~2, $\Delta=0.017$), and achieves its final improvement by halving the batch size to fit more optimizer steps (step~4). The variants with Literature show a notably different behavior. Opus~4.5~+Literature achieves the single largest improvement in the table ($\Delta=0.036$) by introducing focal loss to replace cross-entropy around step 15. Opus~4.6~+Literature scores the most number of improvements with steady, moderate gains, progressively tuning depth, batch size, learning rates, and MLP width. These trajectories show how literature access can nudge agents toward qualitatively different optimization strategies.

We refer the reader to Appendix~\ref{app:appendix_autoresearch} for a detailed breakdown of the features modified by the agents across all runs and their impact on the final validation BPB. A detailed progression of the 100 runs is given in Figure~\ref{fig:rawtrajectories}. The full implementation of the solution achieving the lowest validation BPB is given in Appendix~\ref{app:code_implementation}.
\begin{table}[H]
\centering
\caption{Improvement steps for the best seed of Greedy Opus~4.5, Opus~4.6, and the two +Literature variants on the Autoresearch task. Each row describes the changes introduced at that step relative to the previous one.}
\label{tab:autoresearch_steps}
\scriptsize
\newcolumntype{C}{>{\centering\arraybackslash}X}
\begin{tabularx}{\textwidth}{cXXXX}
\toprule
\textbf{Step} & \textcolor{Red}{\textbf{Opus 4.5 (seed 8)}} &
\textcolor{RoyalBlue}{\textbf{Opus 4.6 (seed 2)}} &
\textcolor{BurntOrange}{\textbf{Opus 4.5 +Lit (seed 10)}} &
\textcolor{Emerald}{\textbf{Opus 4.6 +Lit (seed 4)}} \\
\midrule
\multirow{2}{*}{\textbf{Baseline}} & \multicolumn{4}{p{\dimexpr\textwidth-2\tabcolsep-2em\relax}}{BPB = 1.012. ReLU\textsuperscript{2} MLP, \texttt{depth} = 8, \texttt{aspect\_ratio} = 64, \texttt{head\_dim} = 128, MHA (6 heads), \texttt{window\_pattern} = SSSL, \texttt{matrix\_lr} = 0.04, \texttt{embedding\_lr} = 0.6, \texttt{weight\_decay} = 0.2, \texttt{total\_batch} = $2^{19}$, Muon \texttt{ns\_steps} = 5, value embeddings on alternating layers.} \\
\midrule
1 &
\textbf{1.008} ($\Delta$ = 0.005) \newline
Wider (\texttt{aspect} 64$\to$80); removed value embeddings; \texttt{matrix\_lr} $\uparrow$0.045; \texttt{weight\_decay} $\downarrow$0.10; batch $2^{19}\to 2^{18}$; \texttt{ns\_steps} 5$\to$3. &
\textbf{1.007} ($\Delta$ = 0.005) \newline
Deeper (\texttt{depth} 8$\to$10); \texttt{aspect} 64$\to$58; \texttt{head\_dim} 128$\to$64 (9 heads); window SSSL$\to$SSL; \texttt{matrix\_lr} $\uparrow$0.05; batch $2^{19}\to 2^{18}$. &
\textbf{1.004} ($\Delta$ = 0.008) \newline
Deeper (\texttt{depth} 8$\to$10); \texttt{embedding\_lr} $\uparrow$0.8; \texttt{matrix\_lr} $\uparrow$0.05; \texttt{weight\_decay} $\downarrow$0.15; \texttt{warmdown} 0.5$\to$0.7. &
\textbf{0.995} ($\Delta$ = 0.017) \newline
Deeper (\texttt{depth} 8$\to$10); batch $2^{19}\to 2^{18}$; \texttt{embedding\_lr} $\uparrow$0.8; \texttt{matrix\_lr} $\uparrow$0.05; \texttt{warmdown} 0.5$\to$0.7. \\
\midrule
2 &
\textbf{0.992} ($\Delta$ = 0.015) \newline
Wider (\texttt{aspect} 80$\to$96); restored value embeddings; \texttt{matrix\_lr} $\downarrow$0.038; \texttt{embedding\_lr} $\downarrow$0.55; \texttt{weight\_decay} $\uparrow$0.12; \texttt{ns\_steps} 3$\to$4. &
\textbf{0.990} ($\Delta$ = 0.017) \newline
Reverted \texttt{head\_dim} 64$\to$128; \texttt{aspect} 58$\to$64; window SSL$\to$SSSL; \texttt{embedding\_lr} $\uparrow$0.8; \texttt{weight\_decay} $\uparrow$0.3. &
\textbf{0.968} ($\Delta$ = 0.036) \newline
Embed init std 1.0$\to$0.5; \texttt{ns\_steps} 5$\to$4; \texttt{softcap} 15$\to$18; \texttt{aspect} 64$\to$56 (narrower); \textbf{focal loss} ($\gamma$=1.0) replacing CE. &
\textbf{0.995} ($\Delta\,<$\,0.001) \newline
\texttt{depth} 10$\to$12; batch $2^{18}\to 2^{17}$; short window $T/2\to T/4$; \texttt{ns\_steps} 5$\to$4; \texttt{embedding\_lr} $\uparrow$1.2; near-full cosine decay (\texttt{warmdown} 0.97). \\
\midrule
3 &
\textbf{0.987} ($\Delta$ = 0.005) \newline
Value embeds sparsified: every-2nd $\to$ every-3rd layer + last. \texttt{softcap} 18$\to$16. LR retuned (\texttt{matrix\_lr} 0.038$\to$0.04, \texttt{embedding\_lr} 0.55$\to$0.58). &
\textbf{0.990} ($\Delta\,<$\,0.001) \newline
\texttt{depth} 10$\to$8; \texttt{aspect} 64$\to$96 (wider); \texttt{embedding\_lr} $\uparrow$1.0; \texttt{weight\_decay} $\uparrow$0.35; short window halved ($T/2 \to T/4$); \texttt{softcap} 15$\to$20. &
--- &
\textbf{0.994} ($\Delta$ = 0.001) \newline
Train at \texttt{seq\_len}=1024 (half) for speed; \texttt{device\_batch} 64$\to$128; full cosine schedule; \texttt{final\_lr\_frac} $\downarrow$0.01. \\
\midrule
4 &
\textbf{0.986} ($\Delta$ = 0.001) \newline
Added learnable per-head attention output scale (\texttt{attn\_out\_scale}, init 1.0, lr = 0.1). Applied RMSNorm to attention output before scaling. &
\textbf{0.987} ($\Delta$ = 0.003) \newline
Batch $2^{18}\to 2^{17}$ (more optim steps); \texttt{device\_batch} 128$\to$64; \texttt{matrix\_lr} $\downarrow$0.035; \texttt{embedding\_lr} $\downarrow$0.8; \texttt{warmdown} 0.5$\to$0.7. &
--- &
\textbf{0.993} ($\Delta$ = 0.001) \newline
Reverted to full \texttt{seq\_len}=2048; \texttt{device\_batch} 128$\to$32; LRs reduced (\texttt{embedding} 1.2$\to$0.7, \texttt{matrix} 0.06$\to$0.035); \texttt{weight\_decay} $\downarrow$0.10. \\
\midrule
5 &
--- &
--- &
--- &
\textbf{0.988} ($\Delta$ = 0.005) \newline
\texttt{depth} 12$\to$10; \texttt{device\_batch} 32$\to$64; \texttt{x0\_lambdas} init $\uparrow$0.15; short window $T/8\to T/4$; LRs retuned up. \\
\midrule
6 &
--- &
--- &
--- &
\textbf{0.985} ($\Delta$ = 0.003) \newline
\texttt{mlp\_ratio} 4.0$\to$4.5 (wider MLP); \texttt{softcap} 20$\to$18; WSD schedule (\texttt{stable}=29\%, \texttt{warmdown}=70\%); \texttt{embedding\_lr} $\uparrow$0.9; \texttt{matrix\_lr} $\uparrow$0.055; \texttt{weight\_decay} $\downarrow$0.12. \\
\bottomrule
\end{tabularx}
\end{table}

\section{Conclusion}
\label{sec:conclusions}

We assessed agents' capability to design new neural architectures and training methods for foundation models. We did so through two complementary approaches, AIRA-Compose and AIRA-Design, spanning 12 new diverse agentic tasks derived from 3 distinct frameworks: Composer, LRA, and Autoresearch.

With AIRA-Compose, we deployed 11 agents across 340 24-hour and 300 60-hour runs, demonstrating their ability to autonomously formulate hypotheses on the optimal arrangement of computational primitives and construct original architectures. Unlike traditional methods constrained by rigid optimization objectives, these agents creatively and systematically navigate vast combinatorial search spaces. As a result, we introduce 14 novel architectures, AIRAformers and AIRAhybrids, which exhibit favorable downstream task performance, robust scaling properties, and promising loss--efficiency trade-offs.

Through AIRA-Design, we evaluated the ability of agents to design novel attention mechanisms and efficient training loops for a small language model. On the Long Range Arena (LRA) benchmark we deployed 20 agents, and show that Greedy Gemini 3 Pro and Greedy Opus 4.6 achieve a peak accuracy within $3\%$ of the current human SOTA. Strongest agents could efficiently tweak hyperparameters to their advantage and improve their designs. On the open-ended Autoresearch optimization task, we employed 7 agents and showed that augmenting top-performing ones with relevant literature and code repositories shifted their optimization strategies and improved convergence, enabling Greedy Opus 4.5 to surpass the published Autoresearch reference baseline with a final validation BPB of 0.968.

Our results suggest that agent-driven architecture search and design is not only feasible, but it can already yield highly competitive designs to power the next generation of foundation models and agents, ultimately achieving recursive self-improvement.

\textbf{Limitations and Future Work}

\textit{AIRA-Compose.} AIRA-Compose relies on small-scale proxy evaluation, which does not always faithfully predict large-scale performance. This also limits the number of primitive combinations that can be reliably explored, as translating configurations from a small to a large scale requires careful tuning. Testing AIRA-Compose on advanced harnesses such as $\text{AIRA}_2$ \citep{hambardzumyan2026aira_2}, which can assign multiple GPUs per agent, will allow us to perform NAS on larger models and reduce the proxy–target gap. The aggregation and extrapolation steps also remain non-agentic: future work will introduce agentic tasks that do not just search and evaluate, but also aggregate and scale autonomously. Moreover, our search relies on a single dataset per task; allowing agents to validate candidates across multiple datamixes would yield more robust base patterns. Finally, optimizing only over primitive arrangements effectively underutilizes the agents' capabilities. Enriching the search space with tunable hyperparameters or normalization choices would significantly expand the design space and fully exploit the LLMs' potential. Because AIRA-Compose is built as a modular, LLM-agnostic template, we are confident it can be extended to incorporate these enhancements.

\textit{AIRA-Design.} On the LRA benchmark, agents still fell short of producing paradigm-shifting scientific innovation in mechanistic design. The discovered architectures largely recombine and adapt ideas from prior work (e.g., Performer, Longformer, Conformer) rather than introducing new theoretical insights, suggesting that current agents excel at engineering-level synthesis and adaptation but not yet at genuine algorithmic innovation. One-shot agents failed to produce any valid submissions, confirming that iterative refinement and debugging are essential for low-level code generation tasks. The configurable setup generally degraded performance for weaker agents, which struggle when the design space is expanded. Additionally, LRA tasks expose a strong training bias toward PyTorch, which limits LLM proficiency in JAX/Flax---the framework required by the LRA benchmark. On the Autoresearch task, the greedy scaffold's full-file regeneration paradigm prevents surgical, single-variable edits, making causal attribution of improvements challenging and leading to compound modifications that are hard to interpret. Literature access does not uniformly help: some agents underperform with additional context, suggesting that effectively integrating prior knowledge into optimization strategies remains an open challenge. Future work could include: testing AIRA-Design tasks on more advanced harnesses with persistent state; employing agents with tool-use capabilities for interactive debugging; extending the LRA evaluation to more recent long-range benchmarks; and developing scaffolds that enable incremental, interpretable code modifications rather than full rewrites.

\bibliographystyle{plainnat}
\bibliography{paper}

@article{lupidi2026airs,
  title={{AIRS-Bench: a Suite of Tasks for Frontier AI Research Science Agents}},
  author={Lupidi, Alisia and Gauri, Bhavul and Foster, Thomas Simon and Omari, Bassel Al and Magka, Despoina and Pepe, Alberto and Audran-Reiss, Alexis and Aghamelu, Muna and Baldwin, Nicolas and Cipolina-Kun, Lucia and others},
  journal={arXiv preprint arXiv:2602.06855},
  year={2026}
}

@inproceedings{liao2026kernelevolvescalingagentickernel,
      title={{KernelEvolve: Scaling Agentic Kernel Coding for Heterogeneous AI Accelerators at Meta}}, 
      author={Gang Liao and Hongsen Qin and Ying Wang and Alicia Golden and Michael Kuchnik and Yavuz Yetim and Jia Jiunn Ang and Chunli Fu and Yihan He and Samuel Hsia and Zewei Jiang and Dianshi Li and Uladzimir Pashkevich and Varna Puvvada and Feng Shi and Matt Steiner and Ruichao Xiao and Nathan Yan and Xiayu Yu and Zhou Fang and Roman Levenstein and Kunming Ho and Haishan Zhu and Alec Hammond and Richard Li and Ajit Mathews and Kaustubh Gondkar and Abdul Zainul-Abedin and Ketan Singh and Hongtao Yu and Wenyuan Chi and Barney Huang and Sean Zhang and Noah Weller and Zach Marine and Wyatt Cook and Carole-Jean Wu and Gaoxiang Liu},
      year={2026},
      eprint={2512.23236},
      archivePrefix={arXiv},
      primaryClass={cs.LG},
      url={https://arxiv.org/abs/2512.23236}, 
}

@inproceedings{hammond2025agenticoperatorgenerationml,
      title={{Agentic Operator Generation for ML ASICs}}, 
      author={Alec M. Hammond and Aram Markosyan and Aman Dontula and Simon Mahns and Zacharias Fisches and Dmitrii Pedchenko and Keyur Muzumdar and Natacha Supper and Mark Saroufim and Joe Isaacson and Laura Wang and Warren Hunt and Kaustubh Gondkar and Roman Levenstein and Gabriel Synnaeve and Richard Li and Jacob Kahn and Ajit Mathews},
      year={2025},
      eprint={2512.10977},
      archivePrefix={arXiv},
      primaryClass={cs.DC},
      url={https://arxiv.org/abs/2512.10977}, 
}

@article{romera2024mathematical,
  title={{Mathematical discoveries from program search with large language models}},
  author={Romera-Paredes, Bernardino and Barekatain, Mohammadamin and Novikov, Alexander and Balog, Matej and Kumar, M Pawan and Dupont, Emilien and Ruiz, Francisco JR and Ellenberg, Jordan S and Wang, Pengming and Fawzi, Omar and others},
  journal={Nature},
  volume={625},
  number={7995},
  pages={468--475},
  year={2024},
  publisher={Nature Publishing Group UK London}
}

@article{diao2025climb,
  title={{Climb: Clustering-based iterative data mixture bootstrapping for language model pre-training}},
  author={Diao, Shizhe and Yang, Yu and Fu, Yonggan and Dong, Xin and Su, Dan and Kliegl, Markus and Chen, Zijia and Belcak, Peter and Suhara, Yoshi and Yin, Hongxu and others},
  journal={arXiv e-prints},
  pages={arXiv--2504},
  year={2025}
}

@article{borobia2026functional,
  title={{Functional Component Ablation Reveals Specialization Patterns in Hybrid Language Model Architectures}},
  author={Borobia, Hector and Segu{\'\i}-Mas, Elies and Tormo-Carb{\'o}, Guillermina},
  journal={arXiv preprint arXiv:2603.22473},
  year={2026}
}

@article{gemini2025v25,
  title={{Gemini 2.5: Pushing the frontier with advanced reasoning, multimodality, long context, and next generation agentic capabilities}},
  author={{Google Gemini Team}},
  journal={arXiv preprint arXiv:2507.06261},
  year={2025}
}

@article{gemini2024v15,
  title={{Gemini 1.5: Unlocking multimodal understanding across millions of tokens of context}},
  author={{Google Gemini Team} and Georgiev, Petko and Lei, Ving Ian and Burnell, Ryan and Bai, Libin and Gulati, Anmol and Tanzer, Garrett and Vincent, Damien and Pan, Zhufeng and Wang, Shibo and others},
  journal={arXiv preprint arXiv:2403.05530},
  year={2024}
}

@article{openai2025gptoss,
  title={{gpt-oss-120b \& gpt-oss-20b model card}},
  author={{OpenAI Team}},
  journal={arXiv preprint arXiv:2508.10925},
  year={2025}
}

@inproceedings{zoph2017neural,
  title={{Neural Architecture Search with Reinforcement Learning}},
  author={Zoph, Barret and Le, Quoc},
  booktitle={International Conference on Learning Representations},
  year={2017}
}

@article{gu2025jet,
  title={{Jet-nemotron: Efficient language model with post neural architecture search}},
  author={Gu, Yuxian and Hu, Qinghao and Yang, Shang and Xi, Haocheng and Chen, Junyu and Han, Song and Cai, Han},
  journal={arXiv preprint arXiv:2508.15884},
  year={2025}
}

@article{llama2024,
  title={{The Llama 3 Herd of Models}},
  author={{Llama Team}},
  journal={arXiv preprint arXiv:2407.21783},
  year={2024}
}

@article{dao2024transformers,
  title={{Transformers are ssms: Generalized models and efficient algorithms through structured state space duality}},
  author={Dao, Tri and Gu, Albert},
  journal={arXiv preprint arXiv:2405.21060},
  year={2024}
}

@article{olmohybrid,
  title={{Olmo-Hybrid: Architectural Improvements for Language Modeling}},
  author={{Allen Institute for AI}},
  year={2026},
  url={https://allenai.org/papers/olmo-hybrid}
}

@article{bae2025hybrid,
  title={{Hybrid architectures for language models: Systematic analysis and design insights}},
  author={Bae, Sangmin and Acun, Bilge and Habeeb, Haroun and Kim, Seungyeon and Lin, Chien-Yu and Luo, Liang and Wang, Junjie and Wu, Carole-Jean},
  journal={arXiv preprint arXiv:2510.04800},
  year={2025}
}

@article{hambardzumyan2026aira_2,
  title={{AIRA\_2: Overcoming Bottlenecks in AI Research Agents}},
  author={Hambardzumyan, Karen and Baldwin, Nicolas and Toledo, Edan and Hazra, Rishi and Kuchnik, Michael and Omari, Bassel Al and Foster, Thomas Simon and Protopopov, Anton and Gagnon-Audet, Jean-Christophe and Mediratta, Ishita and others},
  journal={arXiv preprint arXiv:2603.26499},
  year={2026}
}

@misc{leblond2023alphacode2,
  author = {Leblond, R{\'e}mi and others},
  title = {{AlphaCode 2 Technical Report}},
  year = {2023},
  url = {https://storage.googleapis.com/deepmind-media/AlphaCode2/AlphaCode2_Tech_Report.pdf},
  publisher = {Google DeepMind}
}

@article{nathani2025mlgym,
  title={{MLGym: A New Framework and Benchmark for Advancing AI Research Agents}},
  author={Nathani, Deepak and Madaan, Lovish and Roberts, Nicholas and Bashlykov, Nikolay and Menon, Ajay and Moens, Vincent and Budhiraja, Amar and Magka, Despoina and Vorotilov, Vladislav and Chaurasia, Gaurav and others},
  journal={arXiv preprint arXiv:2502.14499},
  year={2025}
}

@inproceedings{real2019regularized,
  title={{Regularized evolution for image classifier architecture search}},
  author={Real, Esteban and Aggarwal, Alok and Huang, Yanping and Le, Quoc V},
  booktitle={Proceedings of the aaai conference on artificial intelligence},
  volume={33},
  pages={4780--4789},
  year={2019}
}

@article{javaheripi2022litetransformersearch,
  title={{Litetransformersearch: Training-free neural architecture search for efficient language models}},
  author={Javaheripi, Mojan and de Rosa, Gustavo and Mukherjee, Subhabrata and Shah, Shital and Religa, Tomasz and Teodoro Mendes, Caio Cesar and Bubeck, Sebastien and Koushanfar, Farinaz and Dey, Debadeepta},
  journal={Advances in Neural Information Processing Systems},
  volume={35},
  pages={24254--24267},
  year={2022}
}

@inproceedings{jawahar2023automoe,
  title={{AutoMoE: Heterogeneous mixture-of-experts with adaptive computation for efficient neural machine translation}},
  author={Jawahar, Ganesh and Mukherjee, Subhabrata and Liu, Xiaodong and Kim, Young Jin and Abdul-Mageed, Muhammad and Lakshmanan, Laks and Hassan, Ahmed and Bubeck, Sebastien and Gao, Jianfeng and others},
  booktitle={Findings of the Association for Computational Linguistics: ACL 2023},
  pages={9116--9132},
  year={2023}
}

@article{white2023neural,
  title={{Neural architecture search: Insights from 1000 papers}},
  author={White, Colin and Safari, Mahmoud and Sukthanker, Rhea and Ru, Binxin and Elsken, Thomas and Zela, Arber and Dey, Debadeepta and Hutter, Frank},
  journal={arXiv preprint arXiv:2301.08727},
  year={2023}
}

@article{bercovich2025llama,
  title={{Llama-nemotron: Efficient reasoning models}},
  author={Bercovich, Akhiad and Levy, Itay and Golan, Izik and Dabbah, Mohammad and El-Yaniv, Ran and Puny, Omri and Galil, Ido and Moshe, Zach and Ronen, Tomer and Nabwani, Najeeb and others},
  journal={arXiv preprint arXiv:2505.00949},
  year={2025}
}

@article{elsken2019neural,
  title={{Neural architecture search: A survey}},
  author={Elsken, Thomas and Metzen, Jan Hendrik and Hutter, Frank},
  journal={Journal of Machine Learning Research},
  volume={20},
  number={55},
  pages={1--21},
  year={2019}
}

@inproceedings{zoph2018learning,
  title={{Learning transferable architectures for scalable image recognition}},
  author={Zoph, Barret and Vasudevan, Vijay and Shlens, Jonathon and Le, Quoc V},
  booktitle={Proceedings of the IEEE conference on computer vision and pattern recognition},
  pages={8697--8710},
  year={2018}
}

@misc{andrews2025arescalingagentenvironments,
      title={{ARE: Scaling Up Agent Environments and Evaluations}}, 
      author={Pierre Andrews and Amine Benhalloum and Gerard Moreno-Torres Bertran and Matteo Bettini and Amar Budhiraja and Ricardo Silveira Cabral and Virginie Do and Romain Froger and Emilien Garreau and Jean-Baptiste Gaya and Hugo Laurençon and Maxime Lecanu and Kunal Malkan and Dheeraj Mekala and Pierre Ménard and Grégoire Mialon and Ulyana Piterbarg and Mikhail Plekhanov and Mathieu Rita and Andrey Rusakov and Thomas Scialom and Vladislav Vorotilov and Mengjue Wang and Ian Yu},
      year={2025},
      eprint={2509.17158},
      archivePrefix={arXiv},
      primaryClass={cs.AI},
      url={https://arxiv.org/abs/2509.17158}, 
}

@article{schmidgall2025agent,
  title={{Agent laboratory: Using llm agents as research assistants}},
  author={Schmidgall, Samuel and Su, Yusheng and Wang, Ze and Sun, Ximeng and Wu, Jialian and Yu, Xiaodong and Liu, Jiang and Moor, Michael and Liu, Zicheng and Barsoum, Emad},
  journal={Findings of the Association for Computational Linguistics: EMNLP 2025},
  pages={5977--6043},
  year={2025},
  publisher={Association for Computational Linguistics}
}

@misc{li2024autokagglemultiagentframeworkautonomous,
  title         = {{AutoKaggle: A Multi-Agent Framework for Autonomous Data Science Competitions}},
  author        = {Ziming Li and Qianbo Zang and David Ma and Jiawei Guo and Tuney Zheng and Minghao Liu and Xinyao Niu and Yue Wang and Jian Yang and Jiaheng Liu and Wanjun Zhong and Wangchunshu Zhou and Wenhao Huang and Ge Zhang},
  year          = {2024},
  eprint        = {2410.20424},
  archiveprefix = {arXiv},
  primaryclass  = {cs.AI},
  url           = {https://arxiv.org/abs/2410.20424}
}

@misc{yamada2025aiscientistv2workshoplevelautomated,
      title={{The AI Scientist-v2: Workshop-Level Automated Scientific Discovery via Agentic Tree Search}}, 
      author={Yutaro Yamada and Robert Tjarko Lange and Cong Lu and Shengran Hu and Chris Lu and Jakob Foerster and Jeff Clune and David Ha},
      year={2025},
      eprint={2504.08066},
      archivePrefix={arXiv},
      primaryClass={cs.AI},
      url={https://arxiv.org/abs/2504.08066}, 
}

@article{starace2025paperbench,
  title={{PaperBench: Evaluating AI's Ability to Replicate AI Research}},
  author={Starace, Giulio and Jaffe, Oliver and Sherburn, Dane and Aung, James and Chan, Jun Shern and Maksin, Leon and Dias, Rachel and Mays, Evan and Kinsella, Benjamin and Thompson, Wyatt and others},
  journal={arXiv preprint arXiv:2504.01848},
  year={2025}
}

@article{novikov2025alphaevolve,
  title={{AlphaEvolve: A coding agent for scientific and algorithmic discovery}},
  author={Novikov, Alexander and V{\~u}, Ng{\^a}n and Eisenberger, Marvin and Dupont, Emilien and Huang, Po-Sen and Wagner, Adam Zsolt and Shirobokov, Sergey and Kozlovskii, Borislav and Ruiz, Francisco JR and Mehrabian, Abbas and others},
  journal={arXiv preprint arXiv:2506.13131},
  year={2025}
}

@article{xiang2025scireplicate,
  title={{SciReplicate-Bench: Benchmarking llms in agent-driven algorithmic reproduction from research papers}},
  author={Xiang, Yanzheng and Yan, Hanqi and Ouyang, Shuyin and Gui, Lin and He, Yulan},
  journal={arXiv preprint arXiv:2504.00255},
  year={2025}
}

@misc{jimenez2024swebenchlanguagemodelsresolve,
      title={{SWE-bench: Can Language Models Resolve Real-World GitHub Issues?}}, 
      author={Carlos E. Jimenez and John Yang and Alexander Wettig and Shunyu Yao and Kexin Pei and Ofir Press and Karthik Narasimhan},
      year={2024},
      eprint={2310.06770},
      archivePrefix={arXiv},
      primaryClass={cs.CL},
      url={https://arxiv.org/abs/2310.06770}, 
}

@misc{zhao2025automatedllmspeedrunningbenchmark,
      title={{The Automated LLM Speedrunning Benchmark: Reproducing NanoGPT Improvements}}, 
      author={Bingchen Zhao and Despoina Magka and Minqi Jiang and Xian Li and Roberta Raileanu and Tatiana Shavrina and Jean-Christophe Gagnon-Audet and Kelvin Niu and Shagun Sodhani and Michael Shvartsman and Andrei Lupu and Alisia Lupidi and Edan Toledo and Karen Hambardzumyan and Martin Josifoski and Thomas Foster and Lucia Cipolina-Kun and Abhishek Charnalia and Derek Dunfield and Alexander H. Miller and Oisin Mac Aodha and Jakob Foerster and Yoram Bachrach},
      year={2025},
      eprint={2506.22419},
      archivePrefix={arXiv},
      primaryClass={cs.AI},
      url={https://arxiv.org/abs/2506.22419}, 
}

@inproceedings{yin2025godel,
  title={{G{\"o}del agent: A self-referential agent framework for recursively self-improvement}},
  author={Yin, Xunjian and Wang, Xinyi and Pan, Liangming and Lin, Li and Wan, Xiaojun and Wang, William Yang},
  booktitle={Proceedings of the 63rd Annual Meeting of the Association for Computational Linguistics (Volume 1: Long Papers)},
  pages={27890--27913},
  year={2025}
}

@article{wang2024survey,
  title={{A survey on large language model based autonomous agents}},
  author={Wang, Lei and Ma, Chen and Feng, Xueyang and Zhang, Zeyu and Yang, Hao and Zhang, Jingsen and Chen, Zhiyuan and Tang, Jiakai and Chen, Xu and Lin, Yankai and others},
  journal={Frontiers of Computer Science},
  volume={18},
  number={6},
  pages={186345},
  year={2024},
  publisher={Springer}
}

@incollection{good1966speculations,
  title={{Speculations concerning the first ultraintelligent machine}},
  author={Good, Irving John},
  booktitle={Advances in computers},
  volume={6},
  pages={31--88},
  year={1966},
  publisher={Elsevier}
}

@phdthesis{schmidhuber1987evolutionary,
  title={{Evolutionary principles in self-referential learning, or on learning how to learn: the meta-meta-... hook}},
  author={Schmidhuber, J{\"u}rgen},
  year={1987},
  school={Technische Universit{\"a}t M{\"u}nchen}
}

@article{schmidhuber2003godel,
  title={{G{\"o}del machines: Self-referential universal problem solvers making provably optimal self-improvements}},
  author={Schmidhuber, J{\"u}rgen},
  journal={arXiv preprint cs.LO/0309048},
  year={2003}
}

@article{vaswani2017attention,
  title={{Attention is all you need}},
  author={Vaswani, Ashish and Shazeer, Noam and Parmar, Niki and Uszkoreit, Jakob and Jones, Llion and Gomez, Aidan N and Kaiser, {\L}ukasz and Polosukhin, Illia},
  journal={Advances in neural information processing systems},
  volume={30},
  year={2017}
}

@article{radford2018improving,
  title={{Improving language understanding by generative pre-training}},
  author={Radford, Alec and Narasimhan, Karthik and Salimans, Tim and Sutskever, Ilya and others},
  year={2018},
  publisher={San Francisco, CA, USA}
}

@inproceedings{devlin2019bert,
  title={{Bert: Pre-training of deep bidirectional transformers for language understanding}},
  author={Devlin, Jacob and Chang, Ming-Wei and Lee, Kenton and Toutanova, Kristina},
  booktitle={Proceedings of the 2019 conference of the North American chapter of the association for computational linguistics: human language technologies, volume 1 (long and short papers)},
  pages={4171--4186},
  year={2019}
}

@article{adler2024nemotron,
  title={{Nemotron-4 340b technical report}},
  author={Adler, Bo and Agarwal, Niket and Aithal, Ashwath and Anh, Dong H and Bhattacharya, Pallab and Brundyn, Annika and Casper, Jared and Catanzaro, Bryan and Clay, Sharon and Cohen, Jonathan and others},
  journal={arXiv preprint arXiv:2406.11704},
  year={2024}
}

@article{rahmani2025implicit,
  title={{Implicit language models are RNNs: balancing parallelization and expressivity}},
  author={Rahmani, Babak and Sch{\"o}ne, Mark and Kremer, Heiner and Falck, Fabian and Ballani, Hitesh and Gladrow, Jannes},
  journal={arXiv preprint arXiv:2502.07827},
  year={2025}
}

@inproceedings{peng2023rwkv,
  title={{Rwkv: Reinventing rnns for the transformer era}},
  author={Peng, Bo and Alcaide, Eric and Anthony, Quentin and Albalak, Alon and Arcadinho, Samuel and Biderman, Stella and Cao, Huanqi and Cheng, Xin and Chung, Michael and Derczynski, Leon and others},
  booktitle={Findings of the association for computational linguistics: EMNLP 2023},
  pages={14048--14077},
  year={2023}
}

@inproceedings{gu2024mamba,
  title={{Mamba: Linear-time sequence modeling with selective state spaces}},
  author={Gu, Albert and Dao, Tri},
  booktitle={First conference on language modeling},
  year={2024}
}

@article{yang2025qwen3,
  title={{Qwen3 technical report}},
  author={Yang, An and Li, Anfeng and Yang, Baosong and Zhang, Beichen and Hui, Binyuan and Zheng, Bo and Yu, Bowen and Gao, Chang and Huang, Chengen and Lv, Chenxu and others},
  journal={arXiv preprint arXiv:2505.09388},
  year={2025}
}

@article{tay2022efficient,
  title={{Efficient transformers: A survey}},
  author={Tay, Yi and Dehghani, Mostafa and Bahri, Dara and Metzler, Donald},
  journal={ACM Computing Surveys},
  volume={55},
  number={6},
  pages={1--28},
  year={2022},
  publisher={ACM New York, NY}
}

@article{pope2023efficiently,
  title={{Efficiently scaling transformer inference}},
  author={Pope, Reiner and Douglas, Sholto and Chowdhery, Aakanksha and Devlin, Jacob and Bradbury, James and Heek, Jonathan and Xiao, Kefan and Agrawal, Shivani and Dean, Jeff},
  journal={Proceedings of machine learning and systems},
  volume={5},
  pages={606--624},
  year={2023}
}

@article{sun2023retentive,
  title={{Retentive network: A successor to transformer for large language models}},
  author={Sun, Yutao and Dong, Li and Huang, Shaohan and Ma, Shuming and Xia, Yuqing and Xue, Jilong and Wang, Jianyong and Wei, Furu},
  journal={arXiv preprint arXiv:2307.08621},
  year={2023}
}

@article{lieber2024jamba,
  title={{Jamba: A hybrid transformer-mamba language model}},
  author={Lieber, Opher and Lenz, Barak and Bata, Hofit and Cohen, Gal and Osin, Jhonathan and Dalmedigos, Itay and Safahi, Erez and Meirom, Shaked and Belinkov, Yonatan and Shalev-Shwartz, Shai and others},
  journal={arXiv preprint arXiv:2403.19887},
  year={2024}
}

@inproceedings{singhal2025llama,
  title={{Llama-nemotron: Efficient reasoning models}},
  author={Singhal, Soumye and Zeng, Jiaqi and Bukharin, Alexander and Zhang, Yian and Shen, Gerald and Mahabaleshwarkar, Ameya Sunil and Kartal, Bilal and Suhara, Yoshi and Bercovich, Akhiad and Levy, Itay and others},
  booktitle={The Exploration in AI Today Workshop at ICML 2025},
  year={2025}
}

@article{ren2021comprehensive,
  title={{A comprehensive survey of neural architecture search: Challenges and solutions}},
  author={Ren, Pengzhen and Xiao, Yun and Chang, Xiaojun and Huang, Po-Yao and Li, Zhihui and Chen, Xiaojiang and Wang, Xin},
  journal={ACM Computing Surveys (CSUR)},
  volume={54},
  number={4},
  pages={1--34},
  year={2021},
  publisher={ACM New York, NY, USA}
}

@article{liu2021survey,
  title={A survey on evolutionary neural architecture search},
  author={Liu, Yuqiao and Sun, Yanan and Xue, Bing and Zhang, Mengjie and Yen, Gary G and Tan, Kay Chen},
  journal={IEEE transactions on neural networks and learning systems},
  volume={34},
  number={2},
  pages={550--570},
  year={2021},
  publisher={IEEE}
}

@article{tay2020long,
  title={{Long range arena: A benchmark for efficient transformers}},
  author={Tay, Yi and Dehghani, Mostafa and Abnar, Samira and Shen, Yikang and Bahri, Dara and Pham, Philip and Rao, Jinfeng and Yang, Liu and Ruder, Sebastian and Metzler, Donald},
  journal={arXiv preprint arXiv:2011.04006},
  year={2020}
}

@article{acun2025composer,
  title={{Composer: A search framework for hybrid neural architecture design}},
  author={Acun, Bilge and Sinha, Prasoon and Ardalani, Newsha and Bae, Sangmin and Golden, Alicia and Lin, Chien-Yu and Madhyastha, Meghana and Sun, Fei and Yadwadkar, Neeraja J and Wu, Carole-Jean},
  journal={arXiv preprint arXiv:2510.00379},
  year={2025}
}

@misc{toledo2025airesearchagentsmachine,
      title={{AI Research Agents for Machine Learning: Search, Exploration, and Generalization in MLE-bench}}, 
      author={Edan Toledo and Karen Hambardzumyan and Martin Josifoski and Rishi Hazra and Nicolas Baldwin and Alexis Audran-Reiss and Michael Kuchnik and Despoina Magka and Minqi Jiang and Alisia Maria Lupidi and Andrei Lupu and Roberta Raileanu and Kelvin Niu and Tatiana Shavrina and Jean-Christophe Gagnon-Audet and Michael Shvartsman and Shagun Sodhani and Alexander H. Miller and Abhishek Charnalia and Derek Dunfield and Carole-Jean Wu and Pontus Stenetorp and Nicola Cancedda and Jakob Nicolaus Foerster and Yoram Bachrach},
      year={2025},
      eprint={2507.02554},
      archivePrefix={arXiv},
      primaryClass={cs.AI},
      url={https://arxiv.org/abs/2507.02554}, 
}

@inproceedings{poli2024mechanistic,
  title={{Mechanistic Design and Scaling of Hybrid Architectures}},
  author={Poli, Michael and Thomas, Armin W and Nguyen, Eric and Ponnusamy, Pragaash and Deiseroth, Bj{\"o}rn and Kersting, Kristian and Suzuki, Taiji and Hie, Brian and Ermon, Stefano and Re, Christopher and others},
  booktitle={International Conference on Machine Learning},
  pages={40908--40950},
  year={2024},
  organization={PMLR}
}

@misc{karpathy2026autoresearch,
  author       = {Karpathy, Andrej},
  title        = {{Autoresearch}},
  year         = {2026},
  howpublished = {\url{https://github.com/karpathy/autoresearch}},
  note         = {GitHub repository},
}

@inproceedings{zhang2025memory,
  title={Memory Mosaics},
  author={Zhang, Jianyu and Nolte, Niklas and Sadhukhan, Ranajoy and Chen, Beidi and Bottou, Leon},
  booktitle={The Thirteenth International Conference on Learning Representations},
  year={2025}
}

@misc{anthropic2025claudecode,
  author       = {{Anthropic}},
  title        = {{Claude Code: An Agentic CLI for Repository-Level Engineering}},
  howpublished = {Technical Report},
  year         = {2025},
  url          = {https://www.anthropic.com/news/claude-code},
  note         = {Accessed: 2026-03-27}
}

@misc{google2026gemini31,
  author       = {{Google DeepMind}},
  title        = {{Gemini 3.1 Pro: A Smarter Model for Your Most Complex Tasks}},
  howpublished = {Google DeepMind Blog},
  year         = {2026},
  url          = {https://blog.google/innovation-and-ai/models-and-research/gemini-models/gemini-3-1-pro/},
  note         = {Accessed: 2026-03-27}
}

@article{li2024datacomp,
  title={{Datacomp-lm: In search of the next generation of training sets for language models}},
  author={Li, Jeffrey and Fang, Alex and Smyrnis, Georgios and Ivgi, Maor and Jordan, Matt and Gadre, Samir Yitzhak and Bansal, Hritik and Guha, Etash and Keh, Sedrick Scott and Arora, Kushal and others},
  journal={Advances in Neural Information Processing Systems},
  volume={37},
  pages={14200--14282},
  year={2024}
}

@misc{karpathy_dwarkesh_2025,
    author = {Karpathy, Andrej and Patel, Dwarkesh},
    title = {Andrej Karpathy --- AGI is still a decade away},
    howpublished = {The Dwarkesh Podcast},
    month = {Oct},
    year = {2025},
}

@article{ma2024megalodon,
  title={{Megalodon: Efficient llm pretraining and inference with unlimited context length}},
  author={Ma, Xuezhe and Yang, Xiaomeng and Xiong, Wenhan and Chen, Beidi and Yu, Lili and Zhang, Hao and May, Jonathan and Zettlemoyer, Luke and Levy, Omer and Zhou, Chunting},
  journal={Advances in Neural Information Processing Systems},
  volume={37},
  pages={71831--71854},
  year={2024}
}

@article{ren2024samba,
  title={{Samba: Simple hybrid state space models for efficient unlimited context language modeling}},
  author={Ren, Liliang and Liu, Yang and Lu, Yadong and Shen, Yelong and Liang, Chen and Chen, Weizhu},
  journal={arXiv preprint arXiv:2406.07522},
  year={2024}
}

@article{blakeman2025nemotron,
  title={{Nemotron-h: A family of accurate and efficient hybrid mamba-transformer models}},
  author={Blakeman, Aaron and Basant, Aarti and Khattar, Abhinav and Renduchintala, Adithya and Bercovich, Akhiad and Ficek, Aleksander and Bjorlin, Alexis and Taghibakhshi, Ali and Deshmukh, Amala Sanjay and Mahabaleshwarkar, Ameya Sunil and others},
  journal={arXiv preprint arXiv:2504.03624},
  year={2025}
}

@misc{nemotron3super2026,
  title={{Nemotron 3 Super: Maximum Compute Efficiency for Complex Multi-Agent Applications}},
  author={Alexiuk, Chris and Patel, Chintan and {NVIDIA}},
  year={2026},
  url={https://developer.nvidia.com/blog}
}

@article{basant2025nvidia,
  title={{Nvidia nemotron nano 2: An accurate and efficient hybrid mamba-transformer reasoning model}},
  author={Basant, Aarti and Khairnar, Abhijit and Paithankar, Abhijit and Khattar, Abhinav and Renduchintala, Adithya and Malte, Aditya and Bercovich, Akhiad and Hazare, Akshay and Rico, Alejandra and Ficek, Aleksander and others},
  journal={arXiv preprint arXiv:2508.14444},
  year={2025}
}

@article{dong2024hymba,
  title={{Hymba: A Hybrid Mamba-Transformer Architecture for Small Language Models}},
  author={Dong, Xin and others},
  journal={arXiv preprint arXiv:2411.13676},
  year={2024}
}

@article{yang2024gated,
  title={{Gated DeltaNet: A Linear Attention Alternative to SSMs}},
  author={Yang, Yufeng and others},
  journal={arXiv preprint arXiv:2412.06464},
  year={2024}
}

@misc{nanochat,
  author = {Andrej Karpathy},
  title = {{nanochat: The best ChatGPT that \$100 can buy}},
  year = {2025},
  publisher = {GitHub},
  url = {https://github.com/karpathy/nanochat}
}

@article{seo2025paper2code,
  title={{Paper2code: Automating code generation from scientific papers in machine learning}},
  author={Seo, Minju and Baek, Jinheon and Lee, Seongyun and Hwang, Sung Ju},
  journal={arXiv preprint arXiv:2504.17192},
  year={2025}
}

@article{liu2023sophia,
  title={{Sophia: A scalable stochastic second-order optimizer for language model pre-training}},
  author={Liu, Hong and Li, Zhiyuan and Hall, David and Liang, Percy and Ma, Tengyu},
  journal={arXiv preprint arXiv:2305.14342},
  year={2023}
}

@article{defazio2024road,
  title={{The road less scheduled}},
  author={Defazio, Aaron and Yang, Xingyu and Mehta, Harsh and Mishchenko, Konstantin and Khaled, Ahmed and Cutkosky, Ashok},
  journal={Advances in Neural Information Processing Systems},
  volume={37},
  pages={9974--10007},
  year={2024}
}

@article{yang2021tuning,
  title={{Tuning large neural networks via zero-shot hyperparameter transfer}},
  author={Yang, Ge and Hu, Edward and Babuschkin, Igor and Sidor, Szymon and Liu, Xiaodong and Farhi, David and Ryder, Nick and Pachocki, Jakub and Chen, Weizhu and Gao, Jianfeng},
  journal={Advances in Neural Information Processing Systems},
  volume={34},
  pages={17084--17097},
  year={2021}
}

@article{lee2024grokfast,
  title={{Grokfast: Accelerated grokking by amplifying slow gradients}},
  author={Lee, Jaerin and Kang, Bong Gyun and Kim, Kihoon and Lee, Kyoung Mu},
  journal={arXiv preprint arXiv:2405.20233},
  year={2024}
}

@inproceedings{pagliardini2024ademamix,
  title={{AdEMAMix: Better and Faster Training with Older Gradients}},
  author={Pagliardini, Matteo and Ablin, Pierre and Grangier, David},
  booktitle={OPT 2024: Optimization for Machine Learning},
  year={2024}
}

@article{he2026autoresearch,
  title={{The AutoResearch Moment: From Experimenter to Research Director}},
  author={He, Chaoyue and Zhou, Xin and Wang, Di and Xu, Hong and Liu, Wei and Miao, Chunyan},
  year={2026},
  publisher={Preprints}
}

@article{hagele2024scaling,
  title={{Scaling laws and compute-optimal training beyond fixed training durations}},
  author={H{\"a}gele, Alexander and Bakouch, Elie and Kosson, Atli and Allal, Loubna B and Von Werra, Leandro and Jaggi, Martin},
  journal={Advances in Neural Information Processing Systems},
  volume={37},
  pages={76232--76264},
  year={2024}
}

@article{tissue2024scaling,
  title={{Scaling law with learning rate annealing}},
  author={Tissue, Howe and Wang, Venus and Wang, Lu},
  journal={arXiv preprint arXiv:2408.11029},
  year={2024}
}

@article{li2025predictable,
  title={{Predictable Scale: Part I, Step Law--Optimal Hyperparameter Scaling Law in Large Language Model Pretraining}},
  author={Li, Houyi and Zheng, Wenzhen and Wang, Qiufeng and Zhang, Hanshan and Wang, Zili and Xuyang, Shijie and Fan, Yuantao and Ding, Zhenyu and Wang, Haoying and Ding, Ning and others},
  journal={arXiv preprint arXiv:2503.04715},
  year={2025}
}

@inproceedings{abnar2025parameters,
  title={{Parameters vs FLOPs: Scaling Laws for Optimal Sparsity for Mixture-of-Experts Language Models}},
  author={Abnar, Samira and Shah, Harshay and Busbridge, Dan and El-Nouby, Alaaeldin and Susskind, Joshua M and Thilak, Vimal},
  booktitle={International Conference on Machine Learning},
  pages={204--230},
  year={2025},
  organization={PMLR}
}

@article{islamov2026role,
  title={{On the Role of Batch Size in Stochastic Conditional Gradient Methods}},
  author={Islamov, Rustem and Machacek, Roman and Lucchi, Aurelien and Silveti-Falls, Antonio and Gorbunov, Eduard and Cevher, Volkan},
  journal={arXiv preprint arXiv:2603.21191},
  year={2026}
}

@article{liu2025muon,
  title={{Muon is scalable for llm training}},
  author={Liu, Jingyuan and Su, Jianlin and Yao, Xingcheng and Jiang, Zhejun and Lai, Guokun and Du, Yulun and Qin, Yidao and Xu, Weixin and Lu, Enzhe and Yan, Junjie and others},
  journal={arXiv preprint arXiv:2502.16982},
  year={2025}
}

@misc{modded_nanogpt_2024,
  author       = {Keller Jordan and Jeremy Bernstein and Brendan Rappazzo and
                  @fernbear.bsky.social and Boza Vlado and You Jiacheng and
                  Franz Cesista and Braden Koszarsky and @Grad62304977},
  title        = {{modded-nanogpt: Speedrunning the NanoGPT baseline}},
  year         = {2024},
  url          = {https://github.com/KellerJordan/modded-nanogpt}
}

@article{hu2024minicpm,
  title={{MiniCPM: Unveiling the potential of small language models with scalable training strategies}},
  author={Hu, Shengding and Tu, Yuge and Han, Xu and He, Chaoqun and Cui, Ganqu and Long, Xiang and Zheng, Zhi and Fang, Yewei and Huang, Yuxiang and Zhao, Weilin and others},
  journal={arXiv preprint arXiv:2404.06395},
  year={2024}
}

@article{hsu2024liger,
  title={{Liger kernel: Efficient triton kernels for llm training}},
  author={Hsu, Pin-Lun and Dai, Yun and Kothapalli, Vignesh and Song, Qingquan and Tang, Shao and Zhu, Siyu and Shimizu, Steven and Sahni, Shivam and Ning, Haowen and Chen, Yanning},
  journal={arXiv preprint arXiv:2410.10989},
  year={2024}
}

@inproceedings{kim2024solar,
  title={{Solar 10.7 b: Scaling large language models with simple yet effective depth up-scaling}},
  author={Kim, Sanghoon and Kim, Dahyun and Park, Chanjun and Lee, Wonsung and Song, Wonho and Kim, Yunsu and Kim, Hyeonwoo and Kim, Yungi and Lee, Hyeonju and Kim, Jihoo and others},
  booktitle={Proceedings of the 2024 Conference of the North American Chapter of the Association for Computational Linguistics: Human Language Technologies (Volume 6: Industry Track)},
  pages={23--35},
  year={2024}
}

@inproceedings{geiping2023cramming,
  title={{Cramming: Training a Language Model on a single GPU in one day.}},
  author={Geiping, Jonas and Goldstein, Tom},
  booktitle={International Conference on Machine Learning},
  pages={11117--11143},
  year={2023},
  organization={PMLR}
}

@article{xi2024coat,
  title={{COAT: Compressing optimizer states and activation for memory-efficient fp8 training}},
  author={Xi, Haocheng and Cai, Han and Zhu, Ligeng and Lu, Yao and Keutzer, Kurt and Chen, Jianfei and Han, Song},
  journal={arXiv preprint arXiv:2410.19313},
  year={2024}
}

@article{jain2026autoresearch,
  title={{AutoResearch-RL: Perpetual Self-Evaluating Reinforcement Learning Agents for Autonomous Neural Architecture Discovery}},
  author={Jain, Nilesh and Yadav, Rohit and Kotian, Sagar and AI, Claude},
  journal={arXiv preprint arXiv:2603.07300},
  year={2026}
}

@article{yang2023spectral,
  title={{A spectral condition for feature learning}},
  author={Yang, Greg and Simon, James B and Bernstein, Jeremy},
  journal={arXiv preprint arXiv:2310.17813},
  year={2023}
}

@article{narayan2025mu,
  title={{$\mu$nit Scaling: Simple and Scalable FP8 LLM Training}},
  author={Narayan, Saaketh and Gupta, Abhay and Paul, Mansheej and Blalock, Davis},
  journal={arXiv preprint arXiv:2502.05967},
  year={2025}
}

@article{peng2023yarn,
  title={{YaRN: Efficient context window extension of large language models}},
  author={Peng, Bowen and Quesnelle, Jeffrey and Fan, Honglu and Shippole, Enrico},
  journal={arXiv preprint arXiv:2309.00071},
  year={2023}
}

@article{huang2024ultra,
  title={{Ultra-sparse memory network}},
  author={Huang, Zihao and Min, Qiyang and Huang, Hongzhi and Zhu, Defa and Zeng, Yutao and Guo, Ran and Zhou, Xun},
  journal={arXiv preprint arXiv:2411.12364},
  year={2024}
}

@article{behrouz2024titans,
  title={{Titans: Learning to memorize at test time}},
  author={Behrouz, Ali and Zhong, Peilin and Mirrokni, Vahab},
  journal={arXiv preprint arXiv:2501.00663},
  year={2024}
}

@article{tan2024scaling,
  title={{Scaling stick-breaking attention: An efficient implementation and in-depth study}},
  author={Tan, Shawn and Yang, Songlin and Courville, Aaron and Panda, Rameswar and Shen, Yikang},
  journal={arXiv preprint arXiv:2410.17980},
  year={2024}
}

@article{xie2023residual,
  title={{Residual: Transformer with dual residual connections}},
  author={Xie, Shufang and Zhang, Huishuai and Guo, Junliang and Tan, Xu and Bian, Jiang and Awadalla, Hany Hassan and Menezes, Arul and Qin, Tao and Yan, Rui},
  journal={arXiv preprint arXiv:2304.14802},
  year={2023}
}

@article{zhang2024relu,
  title={{ReLU$^2$ Wins: Discovering Efficient Activation Functions for Sparse LLMs}},
  author={Zhang, Zhengyan and Song, Yixin and Yu, Guanghui and Han, Xu and Lin, Yankai and Xiao, Chaojun and Song, Chenyang and Liu, Zhiyuan and Mi, Zeyu and Sun, Maosong},
  journal={arXiv preprint arXiv:2402.03804},
  year={2024}
}

@article{bondarenko2023quantizable,
  title={{Quantizable transformers: Removing outliers by helping attention heads do nothing}},
  author={Bondarenko, Yelysei and Nagel, Markus and Blankevoort, Tijmen},
  journal={Advances in Neural Information Processing Systems},
  volume={36},
  pages={75067--75096},
  year={2023}
}

@article{liu2021pay,
  title={{Pay attention to MLPs}},
  author={Liu, Hanxiao and Dai, Zihang and So, David and Le, Quoc V},
  journal={Advances in neural information processing systems},
  volume={34},
  pages={9204--9215},
  year={2021}
}

@inproceedings{yuan2025native,
  title={{Native sparse attention: Hardware-aligned and natively trainable sparse attention}},
  author={Yuan, Jingyang and Gao, Huazuo and Dai, Damai and Luo, Junyu and Zhao, Liang and Zhang, Zhengyan and Xie, Zhenda and Wei, Yuxing and Wang, Lean and Xiao, Zhiping and others},
  booktitle={Proceedings of the 63rd Annual Meeting of the Association for Computational Linguistics (Volume 1: Long Papers)},
  pages={23078--23097},
  year={2025}
}

@inproceedings{liu2024mobilellm,
  title={{MobileLLM: Optimizing sub-billion parameter language models for on-device use cases}},
  author={Liu, Zechun and Zhao, Changsheng and Iandola, Forrest and Lai, Chen and Tian, Yuandong and Fedorov, Igor and Xiong, Yunyang and Chang, Ernie and Shi, Yangyang and Krishnamoorthi, Raghuraman and others},
  booktitle={Forty-first International Conference on Machine Learning},
  year={2024}
}

@article{lu2025moba,
  title={{MoBA: Mixture of block attention for long-context LLMs}},
  author={Lu, Enzhe and Jiang, Zhejun and Liu, Jingyuan and Du, Yulun and Jiang, Tao and Hong, Chao and Liu, Shaowei and He, Weiran and Yuan, Enming and Wang, Yuzhi and others},
  journal={arXiv preprint arXiv:2502.13189},
  year={2025}
}

@article{dong2024flex,
  title={{Flex attention: A programming model for generating optimized attention kernels}},
  author={Dong, Juechu and Feng, Boyuan and Guessous, Driss and Liang, Yanbo and He, Horace},
  journal={arXiv preprint arXiv:2412.05496},
  volume={2},
  number={3},
  pages={4},
  year={2024}
}

@article{gloeckle2024better,
  title={{Better \& faster large language models via multi-token prediction}},
  author={Gloeckle, Fabian and Idrissi, Badr Youbi and Rozi{\`e}re, Baptiste and Lopez-Paz, David and Synnaeve, Gabriel},
  journal={arXiv preprint arXiv:2404.19737},
  year={2024}
}

@inproceedings{yedifferential,
  title={{Differential Transformer}},
  author={Ye, Tianzhu and Dong, Li and Xia, Yuqing and Sun, Yutao and Zhu, Yi and Huang, Gao and Wei, Furu},
  booktitle={The Thirteenth International Conference on Learning Representations},
  year={2025}
}

@article{lin2025forgetting,
  title={{Forgetting transformer: Softmax attention with a forget gate}},
  author={Lin, Zhixuan and Nikishin, Evgenii and He, Xu Owen and Courville, Aaron},
  journal={arXiv preprint arXiv:2503.02130},
  year={2025}
}

@article{xu2024kv,
  title={{KV shifting attention enhances language modeling}},
  author={Xu, Mingyu and Cheng, Wei and Wang, Bingning and Chen, Weipeng},
  journal={arXiv preprint arXiv:2411.19574},
  year={2024}
}

@article{chen2024mixture,
  title={{Mixture of hidden-dimensions transformer}},
  author={Chen, Yilong and Shang, Junyuan and Zhang, Zhengyu and Sheng, Jiawei and Liu, Tingwen and Wang, Shuohuan and Sun, Yu and Wu, Hua and Wang, Haifeng},
  journal={arXiv preprint arXiv:2412.05644},
  year={2024}
}

@inproceedings{ainslie2023gqa,
  title={Gqa: Training generalized multi-query transformer models from multi-head checkpoints},
  author={Ainslie, Joshua and Lee-Thorp, James and De Jong, Michiel and Zemlyanskiy, Yury and Lebr{\'o}n, Federico and Sanghai, Sumit},
  booktitle={Proceedings of the 2023 Conference on Empirical Methods in Natural Language Processing},
  pages={4895--4901},
  year={2023}
}

@article{shazeer2020glu,
  title={Glu variants improve transformer},
  author={Shazeer, Noam},
  journal={arXiv preprint arXiv:2002.05202},
  year={2020}
}

@article{carbonneaux2025cwm,
  title={{CWM: An open-weights LLM for research on code generation with world models}},
  author={Carbonneaux, Quentin and Cohen, Gal and Gehring, Jonas and Kahn, Jacob and Kossen, Jannik and Kreuk, Felix and McMilin, Emily and Meyer, Michel and Wei, Yuxiang and Zhang, David and others},
  journal={arXiv preprint arXiv:2510.02387},
  year={2025}
}

\appendix
\newpage
\section*{Appendix}
\label{app:appendixA}

\section{Related Work}
\label{app:appendix_related}

\textbf{Hybrid Language Models.} Transformers are the backbone of modern LLMs \citep{llama2024,gemini2024v15,gemini2025v25, openai2025gptoss}, but the quadratic cost of self-attention limits long-context processing. To address this, the field is moving towards hybrid architectures \citep{dao2024transformers,ren2024samba,gu2024mamba}.  These models interleave diverse computational primitives (e.g., Attention, SSMs) to balance efficiency with expressiveness \citep{bae2025hybrid}. SSMs like Mamba-2 \citep{dao2024transformers}, Gated DeltaNet (GDN) \citep{yang2024gated}, Samba \citep{ren2024samba}, and Hymba \citep{dong2024hymba} represent a compelling alternative because of their linear complexity, striking a balance of performance and cost-efficiency while still successfully capturing complex data dynamics \citep{rahmani2025implicit}. The interleaving of SSM and Attention blocks does not act as a Transformer fallback, but it serves as the language modeling backbone of hybrid architectures \citep{borobia2026functional}, yielding models that surpass the theoretical limits of either primitive alone \citep{olmohybrid}. This paradigm shift powers several SOTA models, including Qwen3-Next \citep{yang2025qwen3}, Nemotron-H \citep{blakeman2025nemotron}, Nemotron Nano 2 \citep{basant2025nvidia} and Nemotron 3 Super \citep{nemotron3super2026}.

\textbf{Neural Architecture Search.} Hybrid models design implies a vast combinatorial search space, and manual tuning is not sufficient, hence researchers rely on Neural Architecture Search (NAS) \citep{elsken2019neural,liu2021survey,white2023neural}. Historically, NAS has been studied for smaller models using reinforcement learning \citep{zoph2017neural} or evolutionary algorithms \citep{real2019regularized} to combine small ``cells'' into full architectures \citep{zoph2018learning}. Applying traditional NAS methods to LLMs is prohibitive, and most attempts rely on approximations. AutoMOE \citep{jawahar2023automoe} utilizes a supernet approach, which trains an over-parameterized network to carve smaller sub-networks from it. LiteTransformerSearch \citep{javaheripi2022litetransformersearch} employs zero-cost performance proxies, estimating a model's fully-trained accuracy based on untrained properties like initial gradient flow and parameter count. Recent works have introduced the Post Neural Architecture Search (PostNAS) paradigm, which prunes existing pretrained models or replaces standard attention blocks with linear variants \citep{bercovich2025llama,gu2025jet}. Mechanistic design principles are used to analytically derive the optimal allocation of diverse layers \citep{poli2024mechanistic}, while empirical search frameworks like Composer \citep{acun2025composer} bypass computational bottlenecks using small scale proxies to identify block arrangements that reliably perform well at scale.

\textbf{Agents towards Recursive Self-Improvement.} Advances in agentic reasoning have enabled LLMs to function as autonomous researchers, capable of scientific and algorithmic discovery \citep{novikov2025alphaevolve,anthropic2025claudecode,google2026gemini31}. More benchmarks are emerging to evaluate agentic capabilities: AIRS-Bench \citep{lupidi2026airs}, assessing the full research cycle pipeline, MLGym \citep{nathani2025mlgym} and MLE-bench \citep{toledo2025airesearchagentsmachine}, focusing on engineering tasks, and PaperBench \citep{starace2025paperbench}, targeting scientific paper reproduction. An immediate next step in agentic research is represented by agents that iteratively discover and implement superior versions of themselves \citep{good1966speculations, schmidhuber1987evolutionary, yin2025godel}. Recent efforts include the Automated LLM Speedrunning Benchmark \citep{zhao2025automatedllmspeedrunningbenchmark}, which challenges agents to minimize training time to a target loss, and Autoresearch \citep{karpathy2026autoresearch}, which automates model code and training loop optimization. These agentic workflows are facilitated by \textit{nanochat} \citep{nanochat}, a minimal, hackable environment spanning the entire LLM pipeline.

\newpage
\section{Small Scale Ranking}
\label{app:ranking}

\begin{figure}[!htbp]
    \centering
    \includegraphics[width=\linewidth]{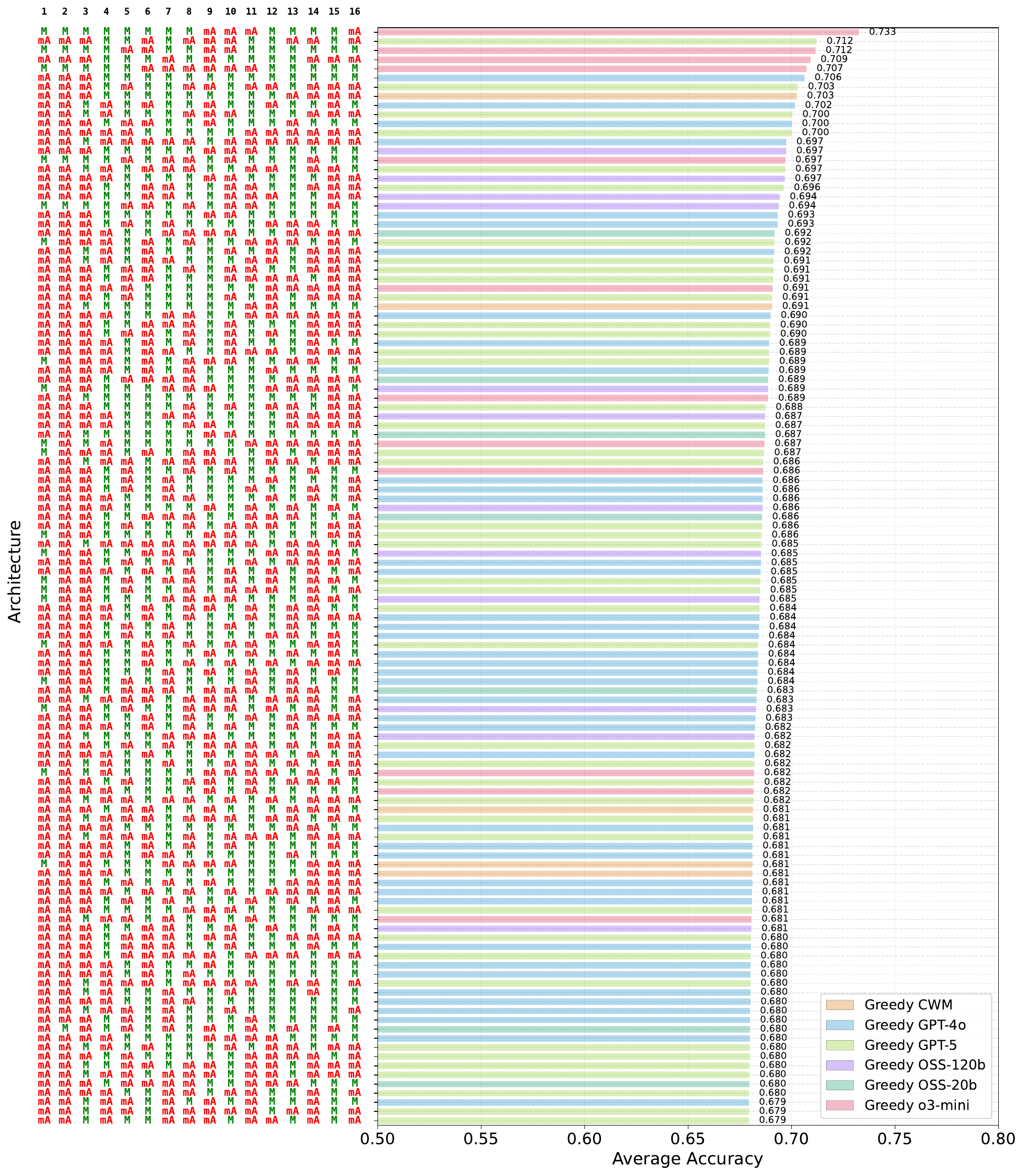}
    \caption{\textbf{Agent-discovered Architecture Ranking on MAD 2-Primitives (M, mA).} Top 120 small scale architectures discovered during greedy search, ranked by test performance. Each row displays the primitive sequence (M in green, mA in red) alongside its accuracy. Bar colors indicate the discovering agent. }
    \label{fig:2primranking}
\end{figure}

\begin{figure}[!htbp]
    \centering
    \includegraphics[width=\linewidth]{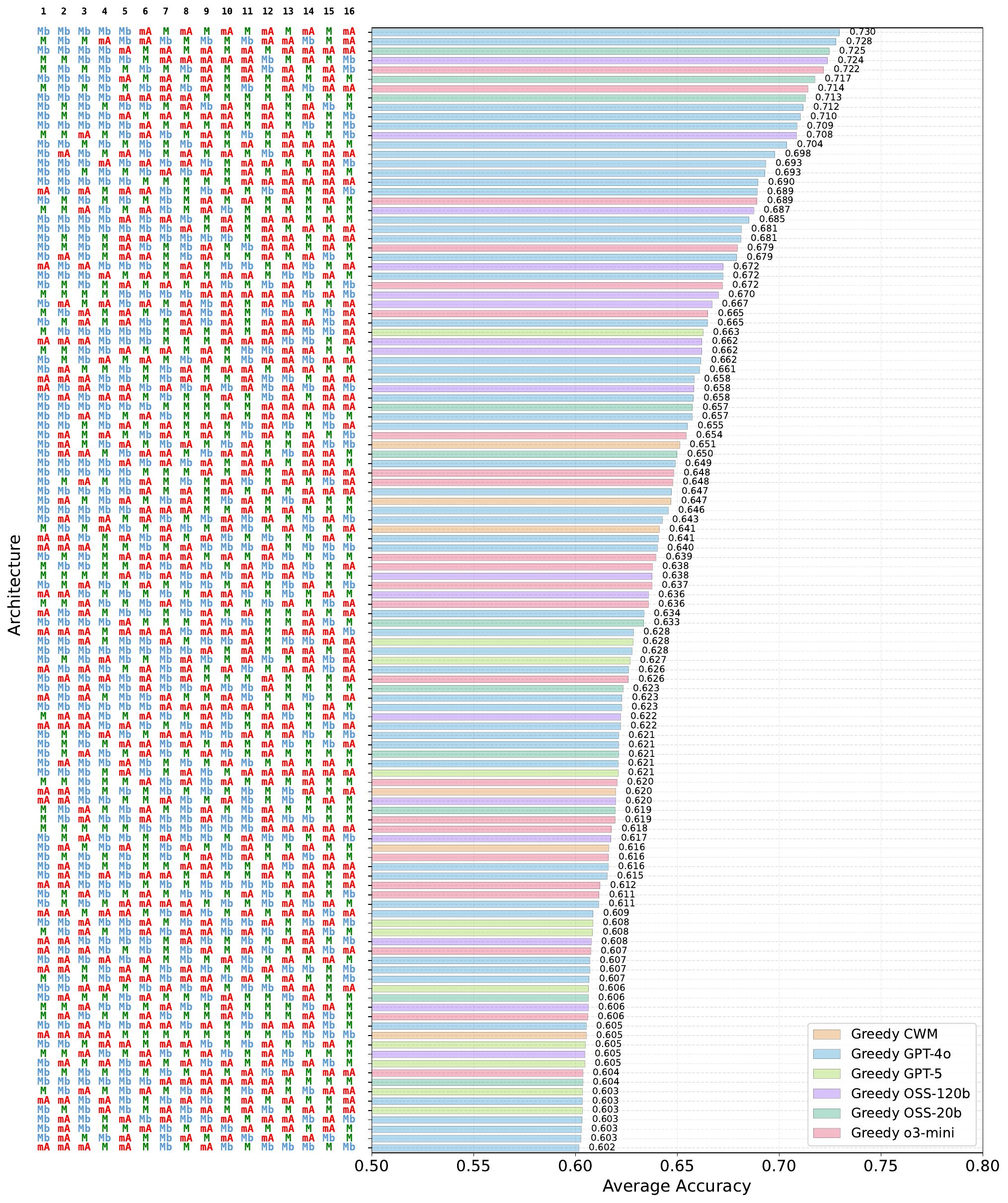}
    \caption{\textbf{Agent-discovered Architecture Ranking on MAD 3-Primitives (M, mA, Mb).} Top 120 small scale architectures discovered during greedy search, ranked by test performance. Each row displays the primitive sequence (M in green, mA in red, Mb in blue) alongside its accuracy. Bar colors indicate the discovering agent. }
    \label{fig:3primranking}
\end{figure}

\newpage
\section{Additional Details}
\label{app:additional}

\begin{table}[h]
\centering
\small
\caption{Environment requirements for AIRA-Compose tasks}
\label{tab:envreq}
\begin{tabular}{ll ll ll}
\toprule
\textbf{Package} & \textbf{Version} & \textbf{Package} & \textbf{Version} & \textbf{Package} & \textbf{Version} \\
\midrule
\texttt{torch} & \texttt{2.6.0} & \texttt{einops} & -- & \texttt{transformers} & -- \\
\texttt{torchvision} & \texttt{0.21.0} & \texttt{seaborn} & -- & \texttt{tokenizers} & -- \\
\texttt{torchaudio} & \texttt{2.6.0} & \texttt{torchmetrics} & -- & \texttt{huggingface-hub} & -- \\
\texttt{ninja} & -- & \texttt{pytorch-lightning} & -- & \texttt{safetensors} & -- \\
\texttt{datasets} & \texttt{4.0.0} & \texttt{opt-einsum} & -- & \texttt{filelock} & -- \\
\texttt{scikit-learn} & -- & \texttt{ray} & -- & \texttt{fsspec} & -- \\
\texttt{multiprocess} & -- & \texttt{scipy} & -- & \texttt{triton} & \texttt{3.1.0} \\
\texttt{ax-platform} & -- & \texttt{absl-py} & -- & \texttt{flash-attn} & \texttt{2.8.3} \\
\texttt{submitit} & -- & \texttt{tiktoken} & -- & \texttt{causal-conv1d} & \texttt{1.5.3.post1} \\
\texttt{blobfile} & -- & \texttt{pyyaml} & -- & \texttt{mamba-ssm} & \texttt{2.2.6.post3} \\
\texttt{numpy} & -- & \texttt{tqdm} & -- & & \\
\texttt{pandas} & -- & \texttt{packaging} & -- & & \\
\bottomrule
\end{tabular}
\end{table}

\begin{figure}[!htbp]
    \centering

    \begin{subfigure}{\linewidth}
        \centering
        \includegraphics[width=\linewidth]{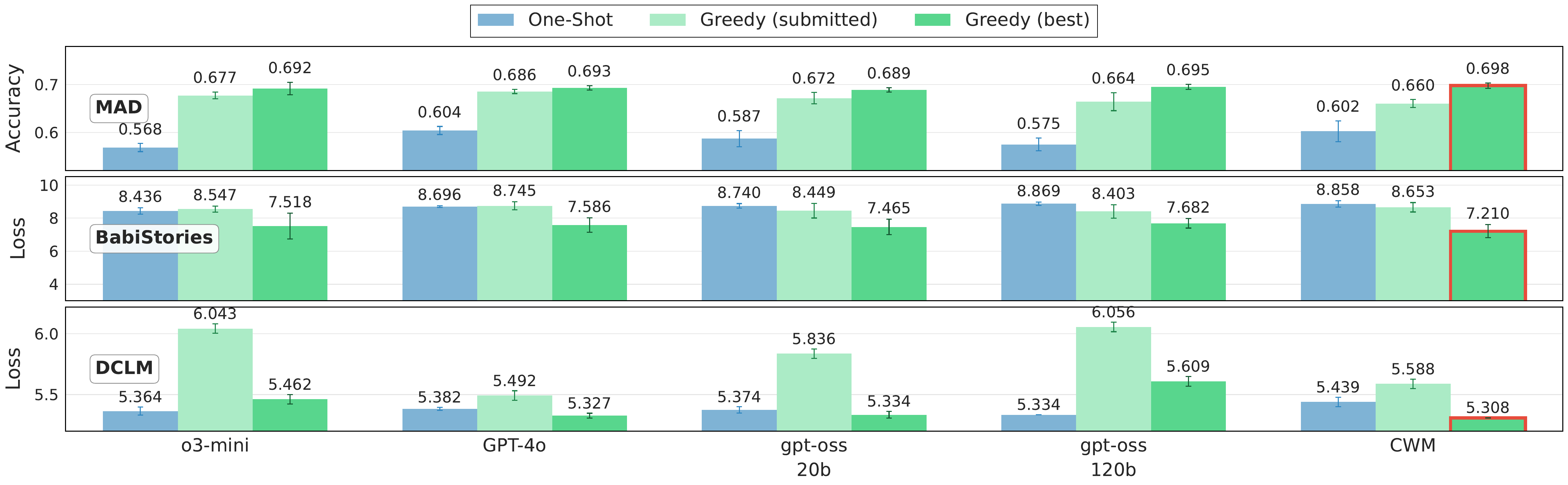}
    \end{subfigure}
    \vspace{-2em}
    \caption{\textbf{Comparison of LLM agents on the 2-Primitives architecture search task across three datasets.} Performance of 11 agents, \{o3-mini, GPT-4o, gpt-oss-20b, gpt-oss-120b, and CWM\} $\times$ \{one-shot, greedy\} and greedy GPT-5. For greedy agents, we distinguish between \textit{submitted} solutions (selected based on best validation fitness) and \textit{best} solutions (achieving the highest test performance across all iterations). Top panel: MAD dataset (accuracy, higher is better); middle and bottom panels: BabiStories and DCLM datasets (loss, lower is better). The 2-Primitives search space is constrained to two primitives (M, mA). Error bars represent 95\% confidence intervals. Best performing agent is highlighted in red.}

    \label{fig:2prim_appendix}
\end{figure}

\begin{figure}[!htbp]
    \centering
    \includegraphics[width=\linewidth]{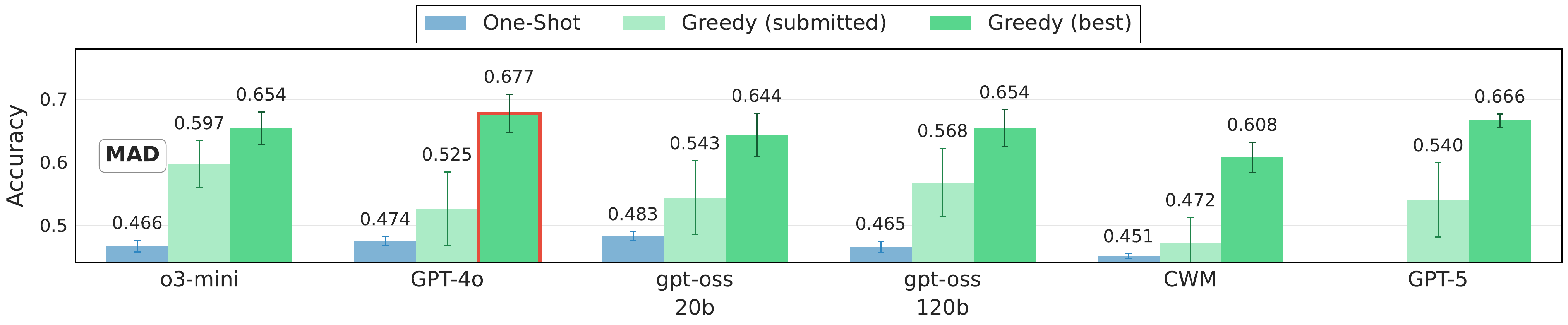}
    \caption{\textbf{Comparison of LLM agents on the 3-Primitives architecture search task on MAD dataset.} Performance of 10 agents, \{o3-mini, GPT-4o, gpt-oss-20b, gpt-oss-120b, and CWM\} $\times$ \{one-shot, greedy\}. For greedy agents, we distinguish between \textit{submitted} solutions (selected based on best validation fitness) and \textit{best} solutions (achieving the highest test performance across all iterations). The 3-Primitives search space is constrained to three primitives (M, mA, Mb). Error bars represent 95\% confidence intervals. Best performing agent is highlighted in red.}
    \label{fig:3primmadacc}
\end{figure}

\begin{figure}[!htbp]
    \centering
    \includegraphics[width=\linewidth]{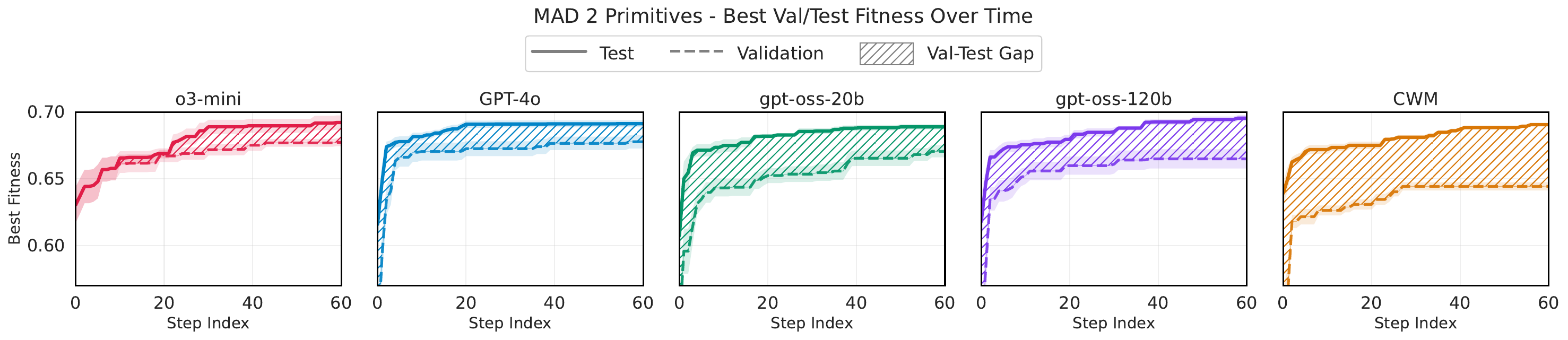}
    \caption{\textbf{Running Maximum Fitness Trajectories for MAD 3 Primitives (M, mA)}. Best validation (dashed line) and test (solid line) fitness achieved up to each step index, averaged across runs, for the greedy agents: o3-mini, GPT-4o, gpt-oss-20b, gpt-oss-120b, CWM. Shaded regions indicate $95\%$ confidence intervals, and hatched areas highlight the generalization gap between validation and test fitness. }
    \label{fig:valtrajectory}
\end{figure}

\begin{figure}[!htbp]
    \centering
    \includegraphics[width=\linewidth]{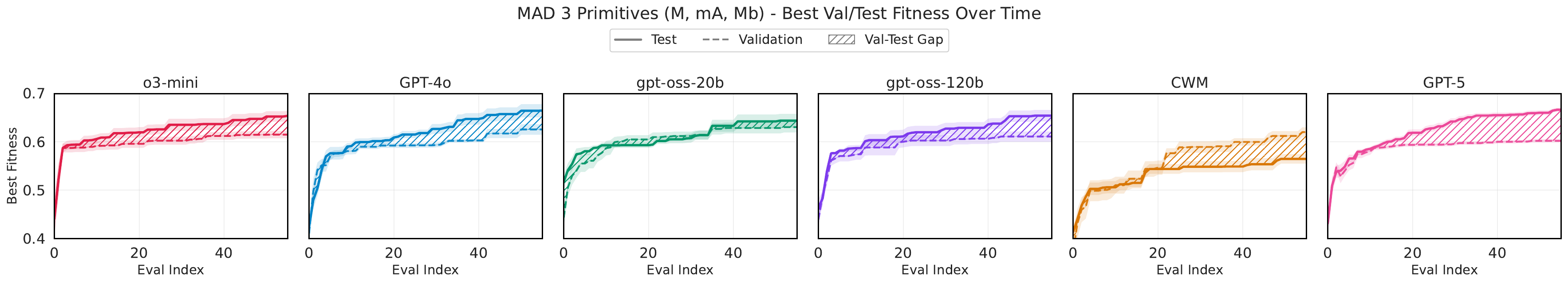}
    \caption{\textbf{Running Maximum Fitness Trajectories for MAD 3 Primitives (M, mA, Mb)}. Best validation (dashed line) and test (solid line) fitness achieved up to each step index, averaged across runs, for the greedy agents: o3-mini, GPT-4o, gpt-oss-20b, gpt-oss-120b, CWM and GPT-5. Shaded regions indicate $95\%$ confidence intervals, and hatched areas highlight the generalization gap between validation and test fitness. }
    \label{fig:trajectories_3prim}
\end{figure}

\begin{figure}[!htbp]
\centering
\includegraphics[width=\textwidth]{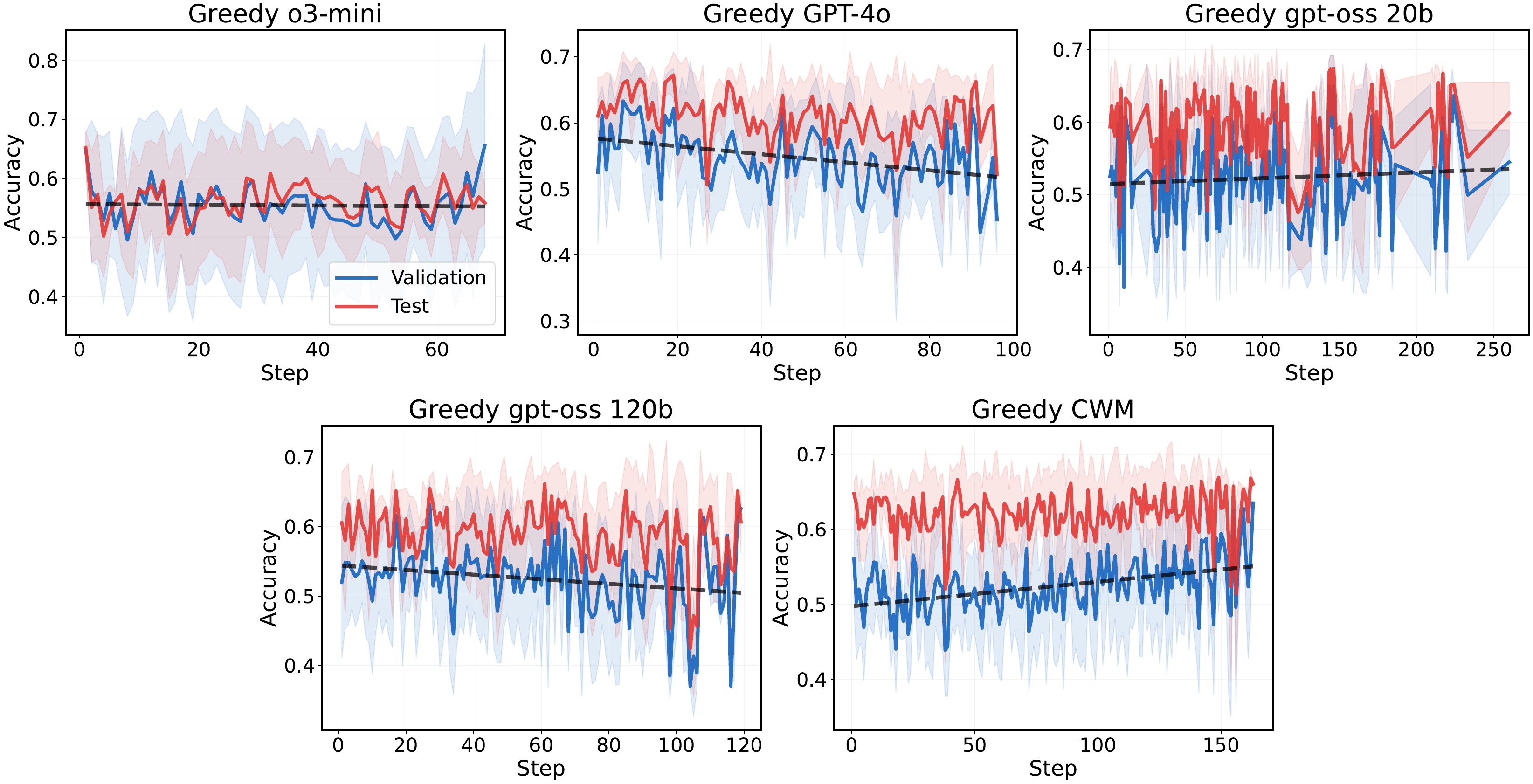}
\caption{\textbf{Averaged exploration trajectories for the 2-primitive (M, mA) greedy search on the MAD benchmark}. Each panel shows the mean validation (blue) and test (red) accuracy per step across 10 seeds, with $\pm 1$ standard deviation shading and validation accuracy linear trend lines. The gap between validation and test accuracy reflects the generalization challenge of agent-designed architectures under the proxy evaluation used for greedy selection.}
\label{fig:trajectories_2prim_avg}
\end{figure}

\begin{figure}[!htbp]
\centering
\includegraphics[width=\textwidth]{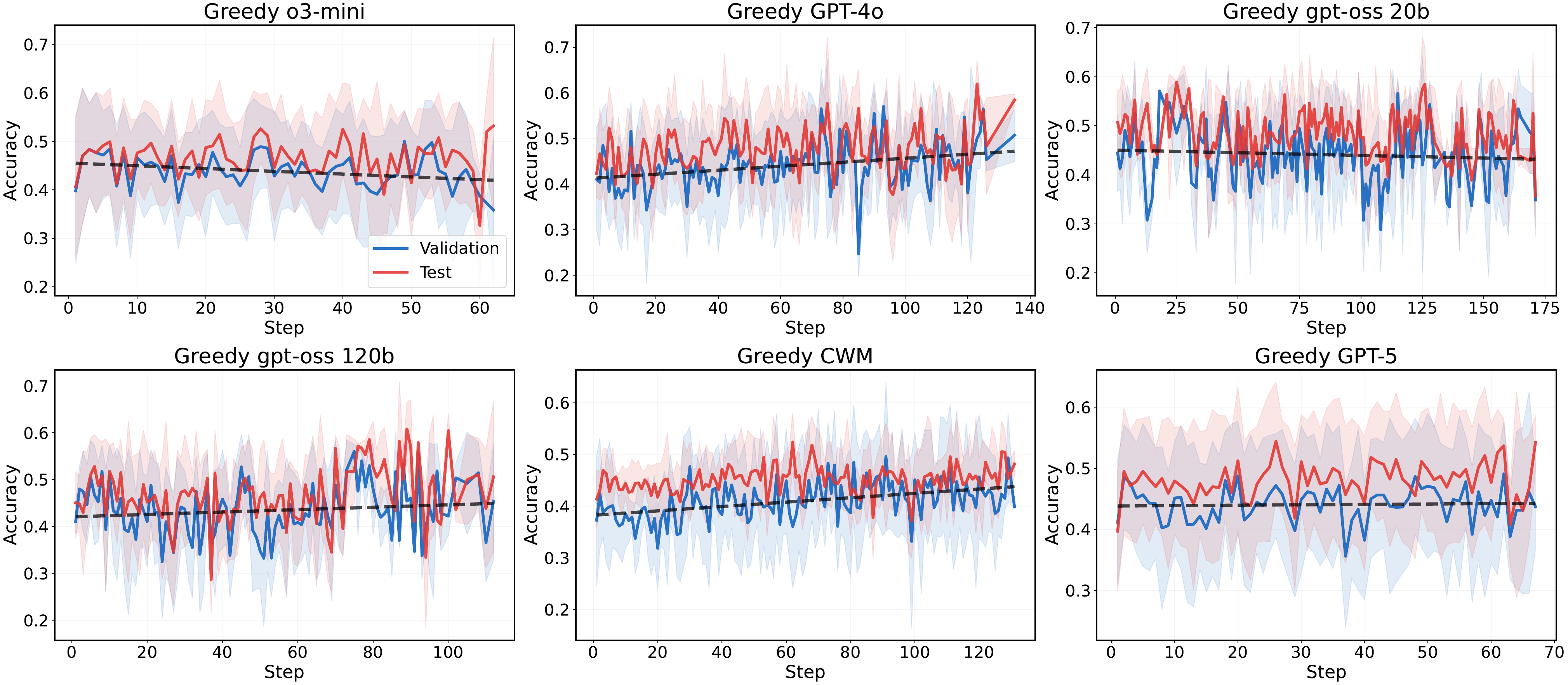}
\caption{\textbf{Averaged exploration trajectories for the 3-primitive (M, mA, Mb) greedy search on the MAD benchmark}. Each panel shows the mean validation (blue) and test (red) accuracy per step across 10 seeds, with $\pm 1$ standard deviation shading and validation accuracy linear trend lines. The gap between validation and test accuracy reflects the generalization challenge of agent-designed architectures under the proxy evaluation used for greedy selection.}
\label{fig:trajectories_3prim_avg}
\end{figure}

\begin{table*}[htbp]
\centering
\caption{Per-task DCLM Core Score breakdown at 1B scale (isotoken budget, 37.5B tokens). Metric per task in parentheses. \textbf{Bold} = best, \underline{underline} = second best per column \textit{within each primitive group}. Last column: DCLM Core Score = unweighted average (\%).}
\label{tab:dclm_pertask}
\resizebox{\textwidth}{!}{%
\begin{tabular}{@{}lcccccccccccccc|c@{}}
\toprule
\textbf{Model} & \rotatebox{90}{COPA (Acc.)} & \rotatebox{90}{CoQA (F1 Score)} & \rotatebox{90}{LAMBADA (Acc.)} & \rotatebox{90}{OBQA (Norm. Acc.)} & \rotatebox{90}{WinoGr. (Acc.)} & \rotatebox{90}{XWino. (Acc.)} & \rotatebox{90}{AGIEval (Norm. Acc.)} & \rotatebox{90}{ARC-C (Norm. Acc.)} & \rotatebox{90}{ARC-E (Norm. Acc.)} & \rotatebox{90}{BoolQ (Acc.)} & \rotatebox{90}{CSQA (Acc.)} & \rotatebox{90}{HSwag (Norm. Acc.)} & \rotatebox{90}{PIQA (Norm. Acc.)} & \rotatebox{90}{SQuADv2 (F1 Score)} & \rotatebox{90}{\textbf{Average}} \\
\midrule
\multicolumn{16}{l}{\textit{2-Primitive (M, mA)}} \\
\midrule
Llama 3.2 & \underline{78.0} & 45.4 & 46.8 & 33.4 & 56.4 & 76.7 & \underline{20.9} & 33.1 & 64.3 & 37.9 & 20.1 & 53.3 & 73.3 & 17.1 & 46.9 \\
Composite (St.) & 75.7 & 43.1 & 43.0 & 33.5 & 57.3 & 76.7 & 20.0 & 34.7 & 65.1 & 37.9 & \textbf{20.6} & 54.7 & 73.1 & 16.6 & 46.6 \\
Composite (Str.) & \textbf{78.3} & 44.6 & 44.5 & 34.5 & 56.2 & 78.5 & 20.3 & \textbf{36.2} & 67.8 & 37.9 & 20.3 & 55.7 & 72.5 & 14.3 & 47.3 \\
\midrule
AIRAformer-A (Str.) & 75.7 & 50.4 & 47.2 & 34.6 & 57.7 & \textbf{80.4} & \textbf{21.0} & \underline{35.9} & 67.9 & 37.9 & 19.8 & 57.3 & \textbf{73.8} & 19.1 & 48.5 \\
AIRAformer-B (St.) & 75.7 & 50.6 & 46.2 & 34.3 & 57.2 & 80.1 & 18.7 & 35.5 & 67.7 & 37.9 & 19.9 & \underline{57.4} & \underline{73.5} & \underline{19.5} & 48.1 \\
AIRAformer-C (St.) & 77.7 & \underline{53.2} & 49.0 & \textbf{35.3} & \underline{58.3} & 80.2 & 20.3 & 35.3 & 67.2 & 37.9 & 18.9 & 56.6 & 73.4 & \textbf{20.2} & \underline{48.8} \\
AIRAformer-C (Str.) & 75.0 & 52.1 & \textbf{49.3} & 33.5 & 56.6 & 78.8 & 18.6 & 34.7 & 67.2 & 37.9 & \textbf{20.6} & 57.0 & 73.1 & 19.2 & 48.1 \\
AIRAformer-D (St.) & 74.3 & \textbf{53.9} & \textbf{49.3} & 33.7 & 57.7 & 80.3 & 19.1 & 35.5 & \underline{68.0} & 37.9 & 19.0 & \underline{57.4} & 73.4 & 18.5 & 48.4 \\
AIRAformer-D (Str.) & 76.7 & 52.7 & \underline{49.2} & \underline{35.0} & \textbf{59.0} & 79.6 & 18.3 & 35.5 & \textbf{68.3} & 37.9 & \underline{20.5} & \textbf{58.1} & \underline{73.5} & \textbf{20.2} & \textbf{48.9} \\
\midrule
\multicolumn{16}{l}{\textit{3-Primitive (M, mA, Mb)}} \\
\midrule
Nemotron-H (Approx.) & \underline{79.0} & 49.0 & 49.1 & 36.2 & 56.2 & 81.3 & \textbf{23.0} & 34.6 & 66.2 & 37.9 & 19.4 & 57.2 & \underline{74.3} & 15.1 & 48.5 \\
Nemotron-2 (Approx.) & 76.0 & 47.6 & 48.6 & 35.8 & 58.3 & 81.1 & 20.9 & 35.7 & 66.8 & 38.0 & 21.1 & 57.9 & 73.0 & 16.6 & 48.4 \\
Mamba (Mb+M) & \underline{79.0} & 34.4 & 45.2 & 35.6 & 57.7 & 80.1 & 16.1 & 35.6 & 64.8 & 37.9 & 19.8 & 55.9 & 73.6 & 9.6 & 46.1 \\
Composer (2Mb-M-3A) & \underline{79.0} & \underline{53.2} & \underline{51.8} & 34.6 & 58.3 & 81.3 & 19.6 & 35.9 & 67.3 & 37.9 & \underline{22.0} & 57.6 & 73.5 & \underline{18.2} & \textbf{49.3} \\
\midrule
AIRAhybrid-A (Str.) & \textbf{82.0} & 40.6 & 47.6 & 35.2 & 57.2 & 79.3 & 19.6 & 34.7 & \underline{67.5} & 37.9 & \textbf{22.9} & 57.2 & 72.7 & 9.9 & 47.5 \\
AIRAhybrid-B (St.) & 78.0 & 50.5 & \textbf{51.9} & 36.4 & \textbf{59.9} & \textbf{82.6} & 18.3 & \underline{36.0} & 66.2 & 37.9 & 19.2 & \textbf{58.5} & 73.6 & 16.6 & 49.0 \\
AIRAhybrid-B (Str.) & 76.0 & 49.6 & 50.5 & 37.0 & 58.7 & 81.8 & 21.3 & \textbf{36.8} & 66.6 & 37.9 & 21.5 & \underline{58.4} & \textbf{75.0} & 16.1 & \underline{49.1} \\
AIRAhybrid-C (Str.) & 76.0 & \textbf{53.5} & 51.2 & 35.6 & 57.9 & 80.6 & \underline{22.6} & 34.6 & 65.8 & 37.9 & 19.1 & 57.0 & 73.1 & \textbf{19.6} & 48.9 \\
AIRAhybrid-D (St.) & \underline{79.0} & 49.1 & 51.4 & 35.8 & 58.3 & 81.8 & 16.1 & 35.1 & 66.8 & 37.9 & 19.2 & 57.2 & 73.2 & 16.0 & 48.4 \\
AIRAhybrid-D (Str.) & 74.0 & 50.8 & 50.6 & \underline{37.2} & 57.9 & 81.5 & 17.4 & \textbf{36.8} & \textbf{67.7} & 37.9 & 20.0 & 57.8 & 73.8 & 16.0 & 48.5 \\
AIRAhybrid-E (St.) & 78.0 & 46.9 & 51.7 & 36.2 & \underline{59.7} & \underline{82.1} & 20.9 & \textbf{36.8} & 67.3 & 37.9 & 21.1 & 57.8 & 72.7 & 14.3 & 48.8 \\
AIRAhybrid-E (Str.) & 74.0 & 48.3 & 51.5 & \textbf{37.4} & 59.0 & 81.7 & \textbf{23.0} & 34.0 & 66.8 & 37.9 & 19.6 & 58.0 & 73.7 & 16.0 & 48.6 \\
\bottomrule
\end{tabular}%
}
\end{table*}

\begin{figure}[!htbp]
    \centering
    \includegraphics[width=\linewidth]{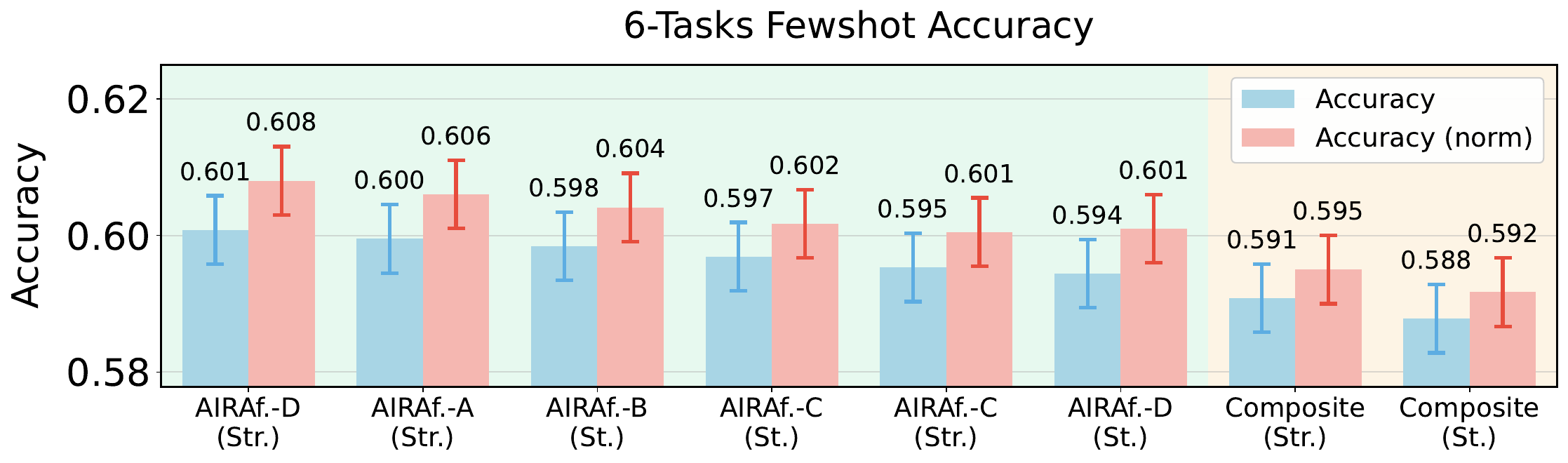}
    \caption{\textbf{Downstream evaluation of AIRAformer architectures at 1B scale}. 0-shot accuracy on 6 commonsense reasoning tasks (ARC-C, ARC-E, HellaSwag, PIQA, SciQ, WinoGrande), showing both raw and normalized accuracy across three seeds.}
    \label{fig:downstream_eval}
\end{figure}

Table~\ref{tab:dclm_pertask} expands the aggregate DCLM Core Scores reported in Tables~\ref{tab:pretraining_results} and~\ref{tab:pretraining_results_hybrid_bolded} by providing the full per-task breakdown across all 14 evaluation tasks. The Composer-found (2Mb-M-3A) architecture achieves the highest overall average (49.3\%), while the AIRAhybrid models show complementary strengths: AIRAhybrid-B variants achieve the best scores on WinoGrande (59.9), XWinograd (82.6), HellaSwag (58.5), and PIQA (75.0), while AIRAhybrid-E (Stretched) leads on OpenBookQA (37.4) and ties for best on AGIEval (23.0). BoolQ remains essentially at chance level ($\sim$37.9\%) across all architectures, indicating that 1B-scale models with our training setup uniformly fail under 10-shot evaluation regardless of architecture. Among the 2-primitive models, the AIRAformer variants consistently outperform the Composite and Llama baselines on CoQA, SQuADv2, and HellaSwag, with AIRAformer-D (Stretched) achieving the highest average (48.9\%).

\begin{figure}[!htbp]
    \centering
    \includegraphics[width=\linewidth]{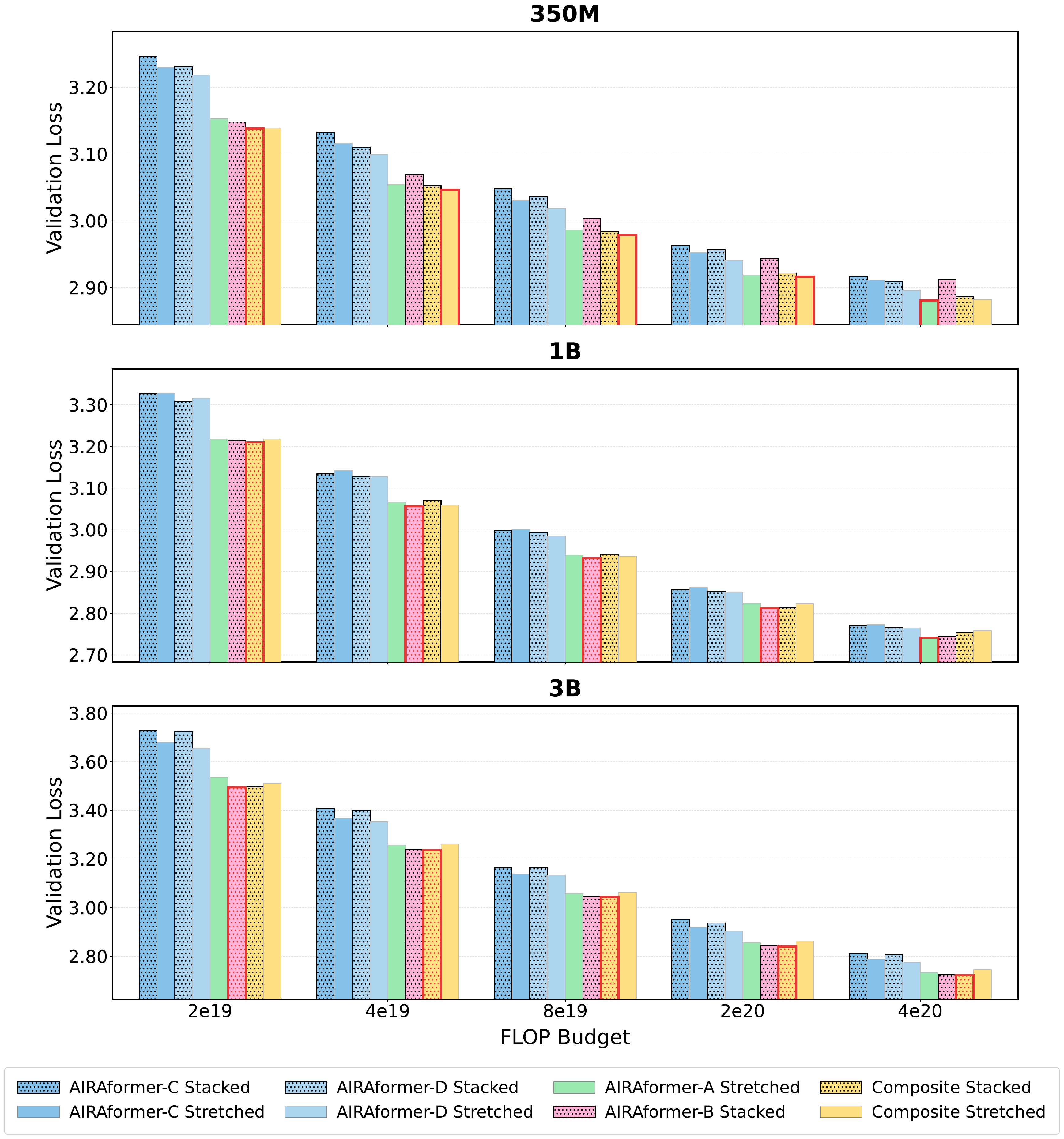}
    \caption{\textbf{IsoFLOP comparison of validation loss across model sizes.} Each subplot shows validation loss for 8 architectures (6 AIRAformer variants and 2 Composite baselines) at a given model size (350M, 1B, 3B), evaluated across 5 FLOP budgets ranging from $2\times10^{19}$ to $4\times10^{20}$. Bars with red borders indicate the architecture achieving the lowest validation loss at each FLOP budget. }
    \label{fig:vallossisoflop}
\end{figure}

\begin{figure}[!htbp]
    \centering
    \includegraphics[width=\linewidth]{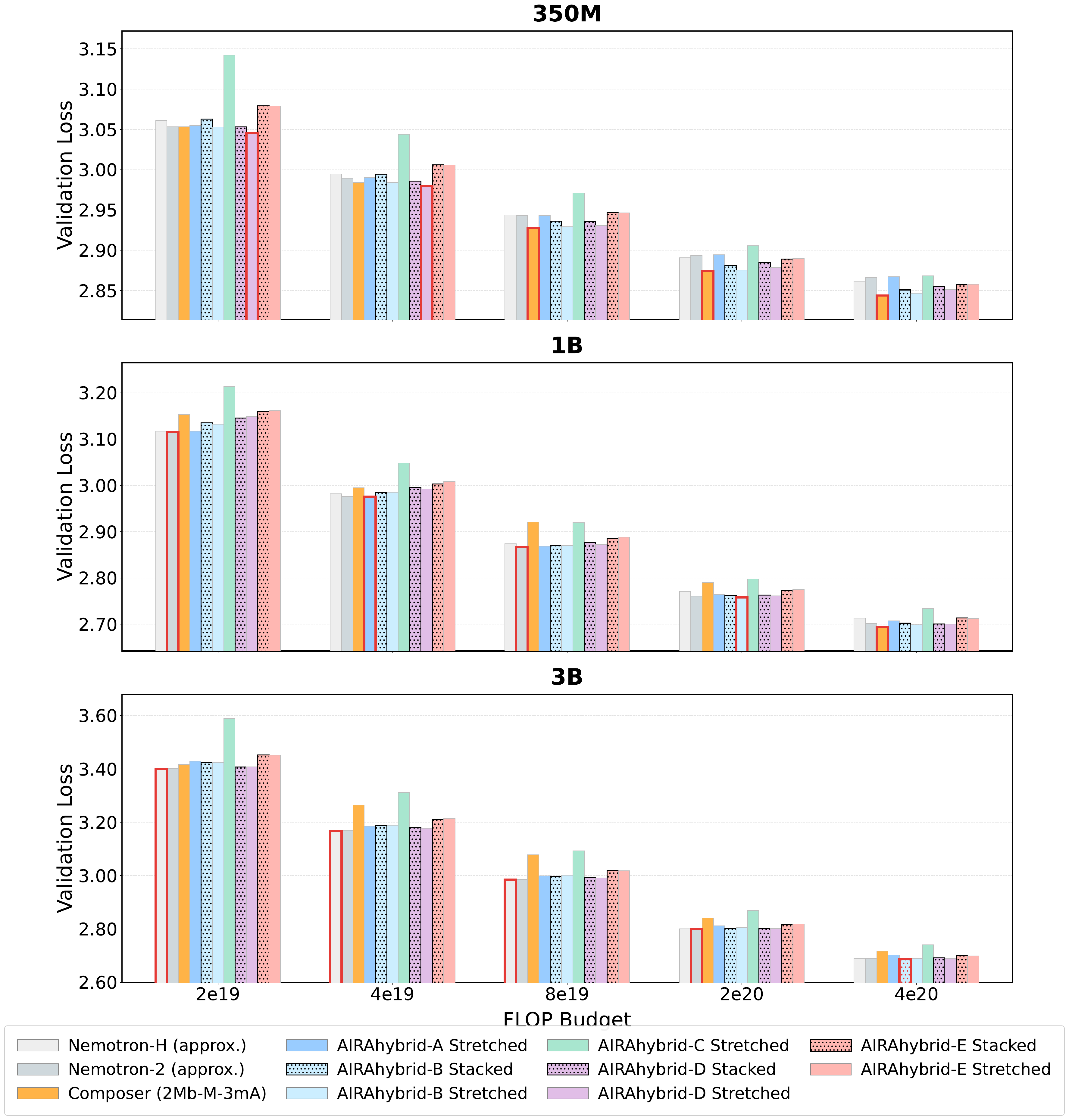}
    \caption{\textbf{IsoFLOP comparison of validation loss across model sizes, 3-primitive case.} Each subplot shows validation loss for 11 architectures (8 AIRAhybrids, 2 baselines and 1 Composer-found baseline) at a given model size (350M, 1B, 3B), evaluated across 5 FLOP budgets ranging from $2\times10^{19}$ to $4\times10^{20}$. Bars with red borders indicate the architecture achieving the lowest validation loss at each FLOP budget. }
    \label{fig:vallossisoflop3b}
\end{figure}

Figures~\ref{fig:vallossisoflop}--\ref{fig:vallossisoflop3b} present the isoFLOP validation loss across three model sizes (350M, 1B, and 3B parameters) and five FLOP budgets for both the 2-primitive and 3-primitive settings. In the 2-primitive case, balanced architectures consistently outperform attention-heavy ones: AIRAformer-A Stretched, AIRAformer-B Stacked, and the Composite baselines achieve lower validation loss than AIRAformer-C and AIRAformer-D (which have A/M ratios exceeding 2) across all scales and budgets. For example, at the 1B scale with $4\times10^{20}$ FLOPs, AIRAformer-A Stretched achieves a validation loss of 2.7415 compared to 2.7647--2.7732 for the attention-heavy variants.

\begin{figure}[!htbp]
    \centering
    \includegraphics[width=\linewidth]{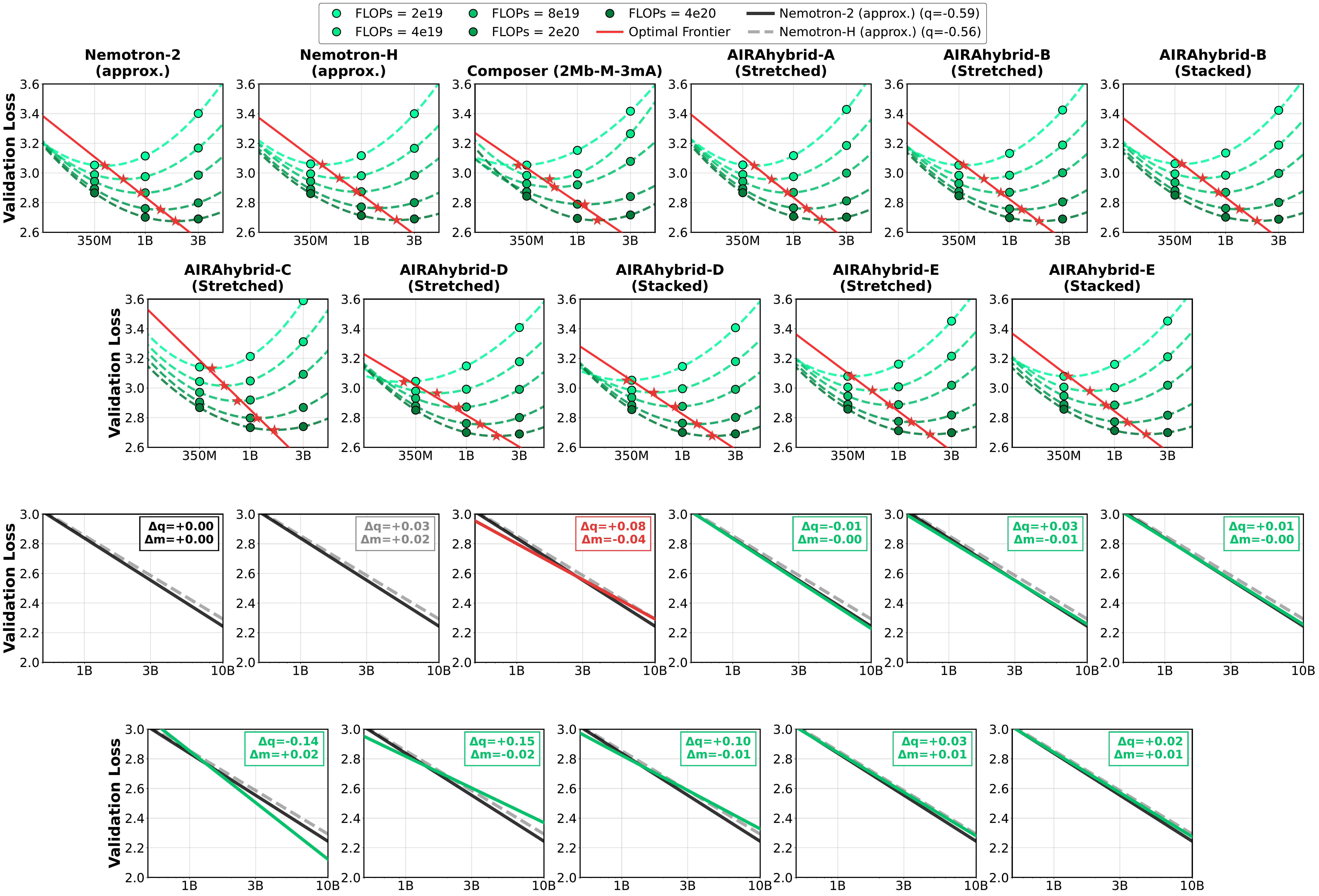}
    \caption{\textbf{IsoFLOP scaling curves and optimal frontier comparison (M, mA, Mb), full results.} \textbf{Top:} Validation loss versus model size for each architecture across 5 FLOPs budgets. \textbf{Bottom:} Comparison of each architecture's optimal frontier against the approximated Nemotron-2 and Nemotron-H.}
    \label{fig:mb_frontier_full}
\end{figure}

In the 3-primitive hybrid setting, we observe a similar pattern: architectures that distribute compute more evenly across Mamba, MLP, and attention primitives---such as AIRAhybrid-C Stretched and the Composer (2Mb-M-3A) model---consistently reach the lowest validation losses, while Mamba-only configurations (AIRAhybrid-A) and attention-dominated hybrids underperform.

\newpage
\section{Aggregation and Scaling Patterns}
\label{app:scaling}

\subsection{Aggregation methods}

In the two-primitive space:

\begin{itemize}
    \item \textbf{AIRAformer-A} (base pattern: \texttt{(A + M) + (2A + 2M) + 4 $\times$ (A + M) + 2M}): Obtained from the top-20 performing architectures on each of the 3 datasets. Within each dataset, the architectures are weighted using a strong exponential decay inversely proportional to their rank. Finally, the 3 datasets are given equal weight in the overall aggregation.

    \item \textbf{AIRAformer-B} (base pattern: \texttt{2A + 5 $\times$ (A + M) + 4M}): Obtained via $N_1$ aggregation on the top-400 architectures across all 5 agents on the MAD dataset.
    \item \textbf{AIRAformer-C} (base pattern: \texttt{(2A + M) + 3 $\times$ (A + M) + (2A + M) + 4A)}: Obtained via $N_1$ aggregation on the top cluster after $k$-means clustering (using 3 clusters) on architectures derived from 20 seeds search of greedy GPT-5 agent on the MAD dataset.

    \item \textbf{AIRAformer-D} (base pattern: \texttt{5 $\times$ (2A + M) + A}):  Obtained via $N_2$ aggregation on the top cluster after $k$-means clustering (using 3 clusters) on architectures derived from 20 seeds search of greedy GPT-5 agent on the MAD dataset.

\end{itemize}

In the three-primitive space:

\begin{itemize}
\item \textbf{AIRAhybrid-A} (base pattern: \texttt{2Mb + M + 11Mb + 2M}): Obtained via $N_0$ aggregation on the top cluster after $k$-means clustering (using 3 clusters) on the architectures found by greedy GPT-5.

\item \textbf{AIRAhybrid-B} (base pattern: \texttt{3 $\times$ (2Mb + M + A) + 2Mb + 2M}): Obtained via $N_1$ aggregation on the top cluster after $k$-means clustering (using 3 clusters) on the architectures found by greedy GPT-5.

\item \textbf{AIRAhybrid-C} (base pattern: \texttt{2Mb + A + 2 $\times$ (Mb + A) + M + 2 $\times$ (A + Mb + A + M)}): Obtained via $N_2$ aggregation on the top cluster after $k$-means clustering (using 3 clusters) on the architectures found by greedy GPT-5.

\item \textbf{AIRAhybrid-D} (base pattern: \texttt{2Mb + M + 2 $\times$ (Mb + M) + A + M + Mb + M + A + M + Mb + M + A}): Obtained via $N_1$ aggregation on the top cluster after $k$-means clustering (using 3 clusters) on the architectures found by all 6 agents.

\item \textbf{AIRAhybrid-E} (base pattern: \texttt{5 $\times$ (Mb + M + A) + M}): Obtained via $N_2$ aggregation on the top cluster after $k$-means clustering (using 3 clusters) on the architectures found by all 6 agents.
\end{itemize}

\subsection{Scaling patterns}

We provide a breakdown of the primitives and their implementation at small scale (16-layer proxy) and large scale (350M, 1B, 3B) below. At small scale, all primitives share a model dimension $d = 128$. At large scale, the model dimension $d$, number of attention heads $n_h$, head dimension $d_h$, and hidden dimensions are configured per scale as detailed in the IsoFLOP methodology section below.

\begin{itemize}

\item \textbf{MLP (M):} A feed-forward network. At small scale, a standard two-layer MLP with ReLU activation and hidden dimension $h_{\text{mlp}} = 258$. At large scale, a SwiGLU variant~\citep{shazeer2020glu} with gated projections: gate and up projections $d \to h_{\text{mlp}}$ and a down projection $h_{\text{mlp}} \to d$, where $h_{\text{mlp}}$ is computed from $d$ and a scale-dependent expansion factor (see below).

\item \textbf{Multi-head Attention (mA):} Standard multi-head causal self-attention~\citep{vaswani2017attention}. At small scale, we use $n_h = 16$ query heads with head dimension $d_h = 8$ and $n_{\text{kv}} = 16$ key-value heads (i.e., multi-head attention without grouping). At large scale, grouped-query attention (GQA)~\citep{ainslie2023gqa} is employed with $n_{\text{kv}} = 8$ key-value heads shared across a larger number of query heads, yielding a total KV dimension of $d_{\text{kv}} = n_{\text{kv}} \times d_h$.

\item \textbf{Mamba (Mb):} A selective state-space model (SSM) based on Mamba-2~\citep{dao2024transformers}, which processes sequences with linear complexity through data-dependent gating and a hardware-efficient selective scan. At small scale, the SSM uses an expansion factor $e_{\text{ssm}} = 1.25$ (yielding SSM hidden dimension $d_{\text{ssm}} = e_{\text{ssm}} \cdot d = 160$), a state dimension $n_s = 4$, a convolution kernel size $k = 4$, and $n_g = 1$ group. At large scale, the SSM hidden dimension is instead computed as $d_{\text{ssm}} = \texttt{multiple\_of}(\lfloor \frac{2}{3} \cdot 3d \rfloor, 256)$, with a larger state dimension $n_s$ and SSM head dimension $d_h^{\text{ssm}} = 64$ (see below for per-scale values).

\end{itemize}

\begin{table*}[!htbp]
\centering
\caption{2-primitive architecture configurations across parameter scales. We report the full layer pattern at each scale, total depth (L), parameter counts (with / without embeddings, in billions), and the approximated ratio of Attention to MLP blocks (A:M).}
\label{tab:architecture_scaling_full}
\small
\begin{tabular}{c c c c c c}
\toprule
\textbf{Architecture} & \textbf{Scale} & \textbf{L} & \textbf{Params (B)} & \textbf{A:M} & \textbf{Pattern} \\
\midrule
Llama 3.2
 & \textbf{350M} & 28 & 0.35 / 0.55 & 1.0:1 & \texttt{14$\times$(A-M)} \\
 & \textbf{1B}   & 32 & 0.97 / 1.23 & 1.0:1 & \texttt{16$\times$(A-M)} \\
 & \textbf{3B}   & 56 & 2.82 / 3.21 & 1.0:1 & \texttt{28$\times$(A-M)} \\
\addlinespace
Composite (Stacked)
 & \textbf{350M} & 24 & 0.35 / 0.55 & 0.5:1 & \texttt{4$\times$(2A-4M)} \\
 & \textbf{1B}   & 27 & 1.00 / 1.26 & 0.5:1 & \texttt{4$\times$(2A-4M)-A-2M} \\
 & \textbf{3B}   & 51 & 3.00 / 3.39 & 0.5:1 & \texttt{8$\times$(2A-4M)-A-2M} \\
\addlinespace
Composite (Stretched)
 & \textbf{350M} & 26 & 0.39 / 0.59 & 0.4:1 & \texttt{3A-8M-3A-5M-2A-5M} \\
 & \textbf{1B}   & 29 & 1.06 / 1.32 & 0.5:1 & \texttt{4A-9M-4A-5M-2A-5M} \\
 & \textbf{3B}   & 54 & 3.17 / 3.56 & 0.5:1 & \texttt{7A-16M-7A-10M-4A-10M} \\
\midrule
AIRAformer-A (Stretched)
 & \textbf{350M} & 26 & 0.34 / 0.54 & 0.8:1 & \texttt{2$\times$(A-M-2A-2M-3$\times$(A-M)-A-3M)} \\
 & \textbf{1B}   & 32 & 1.05 / 1.31 & 0.8:1 & \texttt{2$\times$(A-M-2A-2M)-$4\times$(2A-2M)-4M} \\
 & \textbf{3B}   & 56 & 2.97 / 3.36 & 0.8:1 & \texttt{2$\times$(A-M-4A-4M-6$\times$(A-M)-2A-6M)} \\
\addlinespace
AIRAformer-B (Stacked)
 & \textbf{350M} & 23 & 0.35 / 0.55 & 0.4:1 & \texttt{(3A-4$\times$(M-A)-5M)-7M} \\
 & \textbf{1B}   & 32 & 1.05 / 1.31 & 0.8:1 & \texttt{2$\times$(3A-4$\times$(M-A)-5M)} \\
 & \textbf{3B}   & 53 & 2.94 / 3.33 & 0.7:1 & \texttt{3$\times$(3A-4$\times$(M-A)-5M)-5M} \\
\addlinespace
AIRAformer-C (Stacked)
 & \textbf{350M} & 34 & 0.34 / 0.54 & 2.4:1 & \texttt{2$\times$(2$\times$(2A-M)-3$\times$(A-M)-4A)-2A} \\
 & \textbf{1B}   & 48 & 1.10 / 1.36 & 2.2:1 & \texttt{3$\times$(2$\times$(2A-M)-3$\times$(A-M)-4A)} \\
 & \textbf{3B}   & 76 & 2.92 / 3.31 & 2.4:1 & \texttt{5$\times$(2$\times$(2A-M)-3$\times$(A-M)-4A)-6A} \\
\addlinespace
AIRAformer-C (Stretched)
 & \textbf{350M} & 34 & 0.35 / 0.55 & 3.0:1 & \texttt{4A-2M-2A-2M-2A-2M-2A-8A} \\
 & \textbf{1B}   & 48 & 1.10 / 1.36 & 3.0:1 & \texttt{6A-3M-3A-3M-3A-3M-3A-3M-6A-3M-12A} \\
 & \textbf{3B}   & 70 & 2.92 / 3.31 & 2.4:1 & \texttt{10A-5M-5A-5M-5A-5M-5A-20A-3A} \\
\addlinespace
AIRAformer-D (Stacked)
 & \textbf{350M} & 34 & 0.34 / 0.54 & 2.4:1 & \texttt{2$\times$(5$\times$(2A-M)-A)-2A} \\
 & \textbf{1B}   & 48 & 1.10 / 1.36 & 2.2:1 & \texttt{3$\times$(5$\times$(2A-M)-A)} \\
 & \textbf{3B}   & 76 & 2.92 / 3.31 & 2.4:1 & \texttt{5$\times$(5$\times$(2A-M)-A)-6A} \\
\addlinespace
AIRAformer-D (Stretched)
 & \textbf{350M} & 34 & 0.35 / 0.55 & 2.0:1 & \texttt{4A-2M-4A-2M-4A-2M-4A-2M-4A-2M-2A-M} \\
 & \textbf{1B}   & 48 & 1.10 / 1.36 & 2.2:1 & \texttt{6A-3M-6A-3M-6A-3M-6A-3M-6A-3M-3A} \\
 & \textbf{3B}   & 70 & 2.92 / 3.31 & 2.0:1 & \texttt{10A-5M-10A-5M-10A-5M-10A-5M-10A-5M-5A-3M} \\
\bottomrule
\end{tabular}
\end{table*}

\begin{figure}[!htbp]
    \centering
    \includegraphics[width=\linewidth]{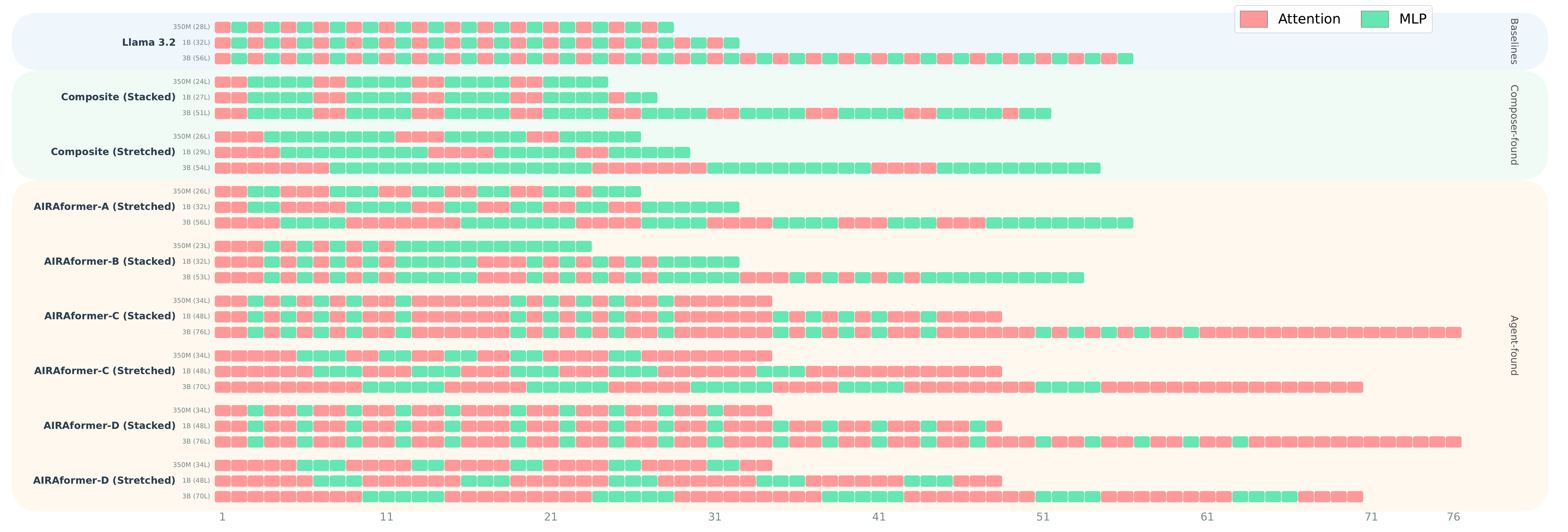}
    \vspace{-2.5em}
    \caption{\textbf{Architecture layer patterns for all discovered and baseline models across three parameter scales (350M, 1B, 3B), 2-primitive case}. Each row within an architecture shows the sequence of Attention (red) and MLP (green) layers. Llama 3.2 uses a uniform 1:1 alternating pattern. Composite architectures are MLP-heavy (5A+11M base ratio). AIRAformer-A and AIRAformer-B (7A+9M base ratio) were obtained via custom and $N_1$ aggregation across multiple  models, respectively. AIRAformer-C and AIRAformer-D (11A+5M base ratio) were discovered via $N_1$ and $N_2$ aggregation on the top cluster after 3-means clustering on GPT-5  rankings. Stacked variants repeat the base pattern, while Stretched variants proportionally scale each contiguous group of layers.}
    \label{fig:allarchitectures}
\end{figure}

\begin{table*}[!htbp]
\centering
\caption{3-primitive (M, mA, Mb) architecture configurations across parameter scales. We report the full layer pattern at each scale, total depth (L), parameter counts (with / without embeddings, in billions), and the number of Attention (A), MLP (M), and Mamba (Mb) blocks.}
\label{tab:architecture_scaling_full_3prims}
\small
\begin{tabular}{c c c c c c c c}
\toprule
\textbf{Architecture} & \textbf{Scale} & \textbf{L} & \textbf{A} & \textbf{M} & \textbf{Mb} & \textbf{Params (B)} & \textbf{Pattern} \\
\midrule
Nemotron-H & \textbf{350M} & 18 & 2 & 8 & 8 & 0.36 / 0.55 & \texttt{Mb-M-A-5$\times$(M-Mb)-A-M-Mb-M-Mb} \\
(Approx.) & \textbf{1B} & 28 & 3 & 13 & 12 & 1.00 / 1.26 & \texttt{Mb-M-2$\times$(A-4$\times$(M-Mb)-M)-A-M-Mb-M} \\
 & \textbf{3B} & 37 & 4 & 17 & 16 & 2.96 / 3.36 & \texttt{Mb-M-2$\times$(A-4$\times$(M-Mb)-M)-A-2$\times$(M-Mb)-M-A-M-3$\times$(Mb-M)} \\
\addlinespace
Nemotron-2 & \textbf{350M} & 17 & 2 & 7 & 8 & 0.33 / 0.52 & \texttt{3$\times$(Mb-M)-Mb-A-4$\times$(M-Mb)-A-M-Mb} \\
(Approx.) & \textbf{1B} & 29 & 3 & 13 & 13 & 1.02 / 1.28 & \texttt{2$\times$(3$\times$(Mb-M)-Mb-A-M)-Mb-M-Mb-A-5$\times$(M-Mb)-M} \\
 & \textbf{3B} & 37 & 4 & 16 & 17 & 2.91 / 3.30 & \texttt{2$\times$(3$\times$(Mb-M)-Mb-A-M)-Mb-M-Mb-A-7$\times$(M-Mb)-A-M-Mb} \\
\addlinespace
Mamba & \textbf{350M} & 17 & 0 & 8 & 9 & 0.36 / 0.56 & \texttt{8$\times$(Mb-M)-Mb} \\
(Mb+M)  & \textbf{1B} & 26 & 0 & 13 & 13 & 0.99 / 1.25 & \texttt{13$\times$(Mb-M)} \\
 & \textbf{3B} & 35 & 0 & 17 & 18 & 2.98 / 3.37 & \texttt{17$\times$(Mb-M)-Mb} \\

 \addlinespace
Composer & \textbf{350M} & 26 & 12 & 4 & 10 & 0.34 / 0.53 & \texttt{4$\times$(2Mb-M-3A)-2Mb} \\
(2Mb-M-3A) & \textbf{1B} & 45 & 21 & 8 & 16 & 1.04 / 1.30 & \texttt{7$\times$(2Mb-M-3A)-2Mb-M} \\
& \textbf{3B} & 57 & 27 & 10 & 20 & 2.98 / 3.38 & \texttt{9$\times$(2Mb-M-3A)-2Mb-M} \\
\addlinespace
\midrule


AIRAhybrid-A & \textbf{350M} & 20 & 0 & 3 & 17 & 0.33 / 0.53 & \texttt{2Mb-M-14Mb-2M-Mb} \\
(Stretched) & \textbf{1B} & 32 & 0 & 6 & 26 & 0.97 / 1.24 & \texttt{4Mb-2M-22Mb-4M} \\
 & \textbf{3B} & 44 & 0 & 8 & 36 & 3.01 / 3.41 & \texttt{6Mb-3M-30Mb-5M} \\
\addlinespace
AIRAhybrid-B & \textbf{350M} & 20 & 4 & 6 & 10 & 0.34 / 0.54 & \texttt{3$\times$(2Mb-M-A)-3$\times$(Mb-M)-A-Mb} \\
(Stretched) & \textbf{1B} & 32 & 6 & 10 & 16 & 0.98 / 1.24 & \texttt{3$\times$(4Mb-2M-2A)-2$\times$(2Mb-2M)} \\
 & \textbf{3B} & 43 & 9 & 13 & 21 & 2.93 / 3.32 & \texttt{3$\times$(5Mb-3M-3A)-3Mb-3M-3Mb-M} \\
\addlinespace
AIRAhybrid-B & \textbf{350M} & 20 & 5 & 7 & 8 & 0.35 / 0.54 & \texttt{3$\times$(Mb-M-A)-2$\times$(3$\times$(Mb-M)-A)-Mb} \\
(Stacked) & \textbf{1B} & 32 & 6 & 10 & 16 & 0.98 / 1.24 & \texttt{2$\times$(3$\times$(2Mb-M-A)-2$\times$(Mb-M))} \\
 & \textbf{3B} & 43 & 7 & 12 & 24 & 2.94 / 3.33 & \texttt{2$\times$(3$\times$(3Mb-M-A)-2$\times$(Mb-M))-Mb-M-A-Mb-M} \\
\addlinespace
AIRAhybrid-C & \textbf{350M} & 29 & 16 & 3 & 10 & 0.33 / 0.53 & \texttt{4Mb-2A-Mb-3$\times$(2A-Mb-2A-M)-Mb-A-Mb-A} \\
(Stretched) & \textbf{1B} & 40 & 21 & 6 & 13 & 0.86 / 1.12 & \texttt{5Mb-3A-2Mb-3$\times$(3A-2Mb-3A-2M)} \\
 & \textbf{3B} & 61 & 35 & 6 & 20 & 2.73 / 3.12 & \texttt{8Mb-5A-3Mb-2$\times$(5A-3Mb-5A-3M)-5A-3Mb-5A} \\
\addlinespace
AIRAhybrid-D & \textbf{350M} & 18 & 3 & 8 & 7 & 0.35 / 0.54 & \texttt{2Mb-M-Mb-3$\times$(M-Mb-M-A)-Mb-M} \\
(Stretched) & \textbf{1B} & 32 & 6 & 14 & 12 & 1.08 / 1.34 & \texttt{4Mb-2M-2Mb-3$\times$(2M-2Mb-2M-2A)} \\
 & \textbf{3B} & 39 & 6 & 17 & 16 & 3.01 / 3.41 & \texttt{5Mb-2M-2Mb-3$\times$(2M-2Mb-2M-2A)-3$\times$(Mb-M)} \\
\addlinespace
AIRAhybrid-D & \textbf{350M} & 18 & 3 & 8 & 7 & 0.35 / 0.54 & \texttt{Mb-M-Mb-3$\times$(M-Mb-M-A)-Mb-M-Mb} \\
(Stacked) & \textbf{1B} & 32 & 6 & 14 & 12 & 1.08 / 1.34 & \texttt{2$\times$(2Mb-3$\times$(M-Mb-M-A))} \\
 & \textbf{3B} & 39 & 7 & 17 & 15 & 2.98 / 3.37 & \texttt{2$\times$(2Mb-3$\times$(M-Mb-M-A))-3$\times$(Mb-M)-A} \\
\addlinespace
AIRAhybrid-E & \textbf{350M} & 20 & 6 & 7 & 7 & 0.34 / 0.54 & \texttt{5$\times$(Mb-M-A)-M-Mb-M-A-Mb} \\
(Stretched) & \textbf{1B} & 32 & 10 & 12 & 10 & 0.97 / 1.23 & \texttt{5$\times$(2Mb-2M-2A)-2M} \\
 & \textbf{3B} & 44 & 14 & 15 & 15 & 2.93 / 3.32 & \texttt{4$\times$(3Mb-3M-3A)-3Mb-3M-2A} \\
\addlinespace
AIRAhybrid-E & \textbf{350M} & 20 & 6 & 7 & 7 & 0.34 / 0.54 & \texttt{5$\times$(Mb-M-A)-M-Mb-M-A-Mb} \\
(Stacked) & \textbf{1B} & 32 & 10 & 12 & 10 & 0.97 / 1.23 & \texttt{2$\times$(5$\times$(Mb-M-A)-M)} \\
 & \textbf{3B} & 44 & 14 & 16 & 14 & 2.98 / 3.38 & \texttt{2$\times$(5$\times$(Mb-M-A)-M)-4$\times$(Mb-M-A)} \\
\bottomrule
\end{tabular}
\end{table*}

\begin{figure}[!htbp]
    \centering
    \includegraphics[width=\linewidth]{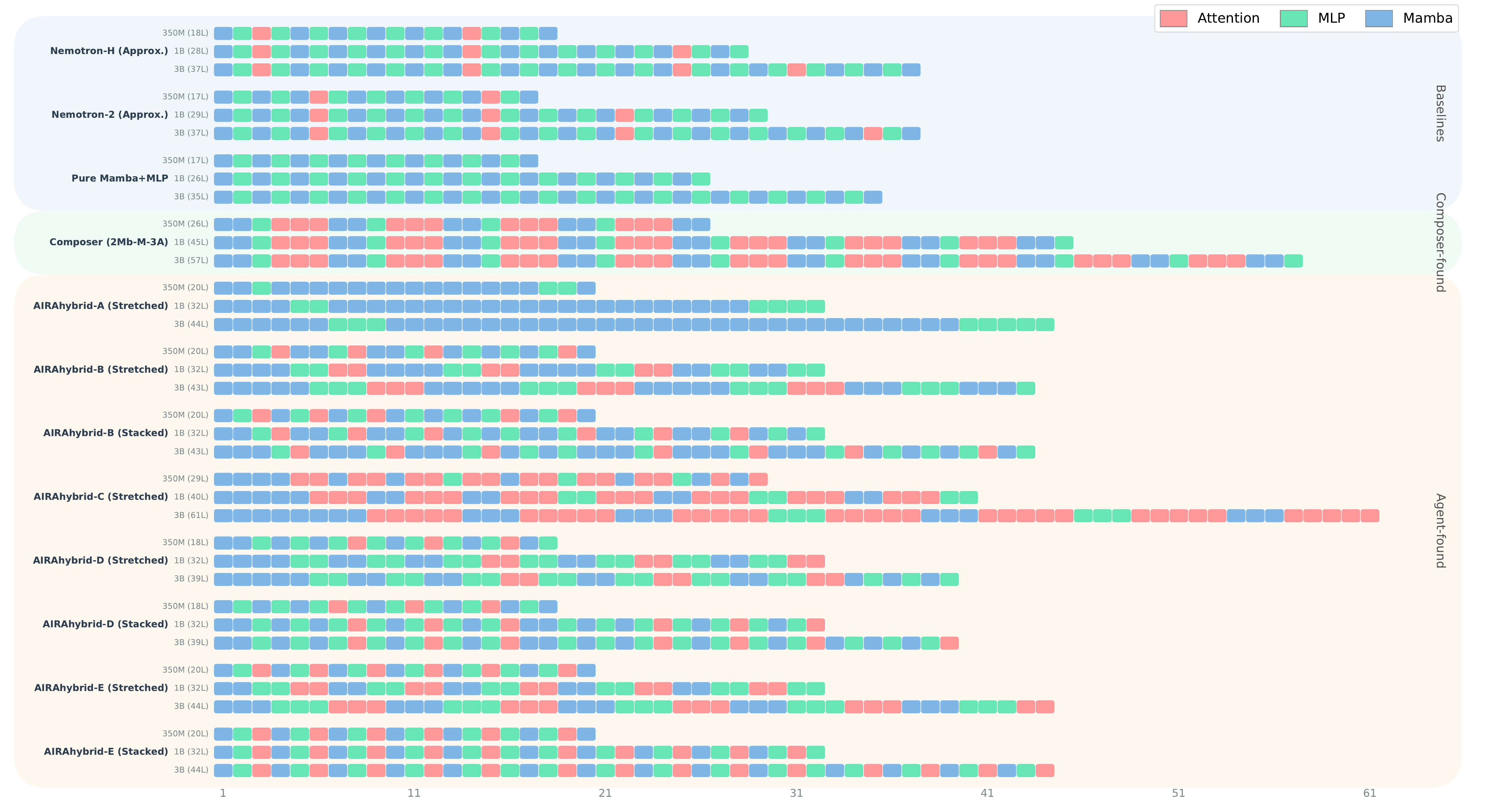}
    \vspace{-2.5em}
    \caption{\textbf{Architecture layer patterns for all discovered and baseline models across three parameter scales (350M, 1B, 3B), 3-primitive case (M, mA, Mb)}. Each row within an architecture shows the sequence of Attention (red), MLP (green) and Mamba (blue) layers. We compare our 8 agent-discovered architectures (AIRAhybrid A-E) with 2 baselines, Nemotron-H and Nemotron-2, approximated to the desired scales, and 1 Composer-found architecture.}
    \label{fig:allarchitectures3prims}
\end{figure}

\subsection{IsoFLOP Budget Methodology}

We compute the training floating-point operations (FLOPs) per token per layer for each primitive using the standard convention. Each linear operation of shape $(M, K) \times (K, N)$ costs $2MKN$ FLOPs in the forward pass. This is multiplied by a factor of 3 for training (accounting for one forward and two backward passes), yielding $6 \cdot d_{\text{in}} \cdot d_{\text{out}}$ FLOPs per projection per token.

\textbf{Model Configurations.}
All three model scales share a sequence length $s = 8{,}192$, $n_{\text{kv}} = 8$ key-value heads, and a constant batch size of $524{,}288$ tokens per step. The batch size is derived via $\text{DP} \times \text{local\_bs} \times \text{acc\_steps} \times s$ (e.g., $16 \times 4 \times 1 \times 8{,}192$ at the 1B scale).

For the \textbf{2-primitive experiments}:
\begin{itemize}
  \item \textbf{350M:} $d = 1536$, $n_h = 24$, $d_h = 64$, $d_{\text{kv}} = 512$
  \item \textbf{1B:} $d = 2048$, $n_h = 32$, $d_h = 64$, $d_{\text{kv}} = 512$
  \item \textbf{3B:} $d = 3072$, $n_h = 24$, $d_h = 128$, $d_{\text{kv}} = 1024$
\end{itemize}

For the \textbf{3-primitive hybrid experiments}:
\begin{itemize}
  \item \textbf{350M:} $d = 1536$, $n_h = 24$, $d_h = 64$, $d_{\text{kv}} = 512$
  \item \textbf{1B:} $d = 2048$, $n_h = 32$, $d_h = 64$, $d_{\text{kv}} = 512$
  \item \textbf{3B:} $d = 3072$, $n_h = 32$, $d_h = 96$, $d_{\text{kv}} = 768$
\end{itemize}

For the \textbf{2-primitive (Attention + MLP)} experiments, the SwiGLU MLP hidden dimension $h$ is computed as $h = \texttt{multiple\_of}(\lfloor \frac{2}{3} \cdot 4d \cdot f \rfloor, 1024)$. We use an expansion factor of $f = 1.0$ at the 350M and 3B scales, and $f = 1.4$ at the 1B scale. This yields $h = 4{,}096$ (350M), $h = 8{,}192$ (1B), and $h = 8{,}192$ (3B).

For the \textbf{3-primitive (Attention + MLP + Mamba)} experiments, the expansion factor is fixed at $f = 1.4$ across all scales, resulting in $h = 6{,}144$ (350M), $h = 8{,}192$ (1B), and $h = 12{,}288$ (3B). The Mamba2 SSM additionally utilizes a state dimension $n_s = 128$ (350M and 1B) or $n_s = 256$ (3B), a convolution kernel size $k = 4$, $n_g = 1$ group, and an SSM head dimension $d_h^{\text{ssm}} = 64$. The SSM hidden dimension is defined as $d_{\text{ssm}} = \texttt{multiple\_of}(\lfloor \frac{2}{3} \cdot 3d \rfloor, 256)$, yielding $d_{\text{ssm}} = 3{,}072$ (350M), $d_{\text{ssm}} = 4{,}096$ (1B), and $d_{\text{ssm}} = 6{,}144$ (3B). Consequently, the number of SSM heads is $n_{\text{ssm}} = d_{\text{ssm}} / d_h^{\text{ssm}}$, which equates to 48, 64, and 96 heads, respectively.


Embeddings are tied with vocabulary size 128{,}256.


\textbf{Per-Primitive FLOPs per Token (Training).}
For \textbf{Grouped-Query Attention}, the computational cost comprises four linear projections---$Q$ ($d \to d$), $K$ ($d \to d_{\text{kv}}$), $V$ ($d \to d_{\text{kv}}$), and $O$ ($d \to d$)---alongside the quadratic attention computation ($QK^\top$ and $\text{softmax} \cdot V$):
\begin{equation}
  F_{\text{Attn}} = 6(2d^2 + 2d \cdot d_{\text{kv}}) + 12sd
\end{equation}

For the \textbf{SwiGLU MLP}, the three projections (gate and up: $d \to h$; down: $h \to d$) result in:
\begin{equation}
  F_{\text{MLP}} = 6 \cdot 3 \cdot d \cdot h = 18dh
\end{equation}

For the \textbf{Mamba2 SSM}, the cost decomposes into four components:
(i) an input projection $d \to d_{\text{in\_proj}}$ where $d_{\text{in\_proj}} = 2d_{\text{ssm}} + 2n_g n_s + n_{\text{ssm}}$;
(ii) a depthwise \texttt{conv1d} over $d_{\text{conv}} = d_{\text{ssm}} + 2n_g n_s$ channels with kernel size $k$;
(iii) the selective scan, costing $2 \cdot n_{\text{ssm}} \cdot d_h^{\text{ssm}} \cdot n_s$ per token; and
(iv) an output projection $d_{\text{ssm}} \to d$.
This yields the following total cost:
\begin{equation}
  F_{\text{SSM}} = 6(d \cdot d_{\text{in\_proj}} + d_{\text{conv}} \cdot k + 2n_{\text{ssm}} d_h^{\text{ssm}} n_s + d_{\text{ssm}} \cdot d)
\end{equation}





\textbf{Training Steps from FLOP Budget.}
Given an architecture with $L_{\text{Attn}}$ attention layers, $L_{\text{MLP}}$ MLP layers, and $L_{\text{SSM}}$ Mamba layers layers, the total FLOPs per training step are:
\begin{equation}
  C_{\text{step}} = (L_{\text{Attn}} \cdot F_{\text{Attn}} + L_{\text{MLP}} \cdot F_{\text{MLP}} + L_{\text{SSM}} \cdot F_{\text{SSM}}) \times B
\end{equation}
where $B = 524{,}288$ is the constant batch size in tokens.

The number of training steps $T$ for a target FLOP budget $\mathcal{F}$ is calculated as $T = \lfloor \mathcal{F} / C_{\text{step}} \rfloor$.

For example, at the 1B scale, the Composite (Stacked) architecture ($L_{\text{Attn}} = 10$, $L_{\text{MLP}} = 19$) yields:
\begin{equation}
  C_{\text{step}} = (10 \cdot 263{,}192{,}576 + 19 \cdot 301{,}989{,}888) \times 524{,}288 \approx 4.39 \times 10^{15} \text{ FLOPs/step}
\end{equation}
This gives $T = \lfloor 2 \times 10^{19} / (4.39 \times 10^{15}) \rfloor = 4{,}552$ steps for a budget of $2 \times 10^{19}$ FLOPs (matching Table \ref{tab:iso_flop_steps}).

Similarly, for the 1B hybrid AIRAhybrid-B Stretched model ($L_{\text{SSM}} = 16$, $L_{\text{MLP}} = 10$, $L_{\text{Attn}} = 6$):
\begin{equation}
  C_{\text{step}} = (6 \cdot 263{,}192{,}576 + 10 \cdot 301{,}989{,}888 + 16 \cdot 161{,}333{,}696) \times 524{,}288 \approx 3.77 \times 10^{15} \text{ FLOPs/step}
\end{equation}
This yields $T = \lfloor 2 \times 10^{19} / (3.77 \times 10^{15}) \rfloor = 5{,}308$ steps at a budget of $2 \times 10^{19}$ FLOPs (matching Table \ref{tab:iso_flop_steps_3prim}).

The full set of training steps for all architectures and budgets is reported in Tables \ref{tab:iso_flop_steps}--\ref{tab:iso_flop_steps_3prim}.

\begin{table*}[!htbp]
\centering
\caption{Training steps required to reach target isoFLOP budgets. We include the count of Attention (A) and MLP (M) layers to contextualize the varying computational intensity per step. Calculations are based on a constant batch size of 524,288 tokens per step. Model base configurations: 350M ($d=1536$, $h=4096$), 1B ($d=2048$, $h=8192$), 3B ($d=3072$, $h=8192$).}
\label{tab:iso_flop_steps}
\small
\begin{tabular}{ll cc ccccc}
\toprule
 &  &  &  & \multicolumn{5}{c}{\textbf{Target FLOP Budget}} \\
\cmidrule(l){5-9}
\textbf{Scale} & \textbf{Architecture} & \textbf{A} & \textbf{M} & \textbf{2e19} & \textbf{4e19} & \textbf{8e19} & \textbf{2e20} & \textbf{4e20} \\
\midrule

\multirow{8}{*}{\textbf{350M}}
 & Composite (Stacked) & 8 & 16 & 11,484 & 22,967 & 45,934 & 114,835 & 229,670 \\
 & Composite (Stretched) & 8 & 18 & 10,751 & 21,501 & 43,002 & 107,505 & 215,011 \\
 & AIRAformer-A (Stretched) & 12 & 14 & 9,907 & 19,815 & 39,629 & 99,073 & 198,147 \\
 & AIRAformer-B (Stacked) & 7 & 16 & 12,175 & 24,351 & 48,701 & 121,753 & 243,506 \\
 & AIRAformer-C (Stacked) & 24 & 10 & 6,737 & 13,474 & 26,948 & 67,370 & 134,740 \\
 & AIRAformer-C (Stretched) & 23 & 11 & 6,828 & 13,656 & 27,312 & 68,280 & 136,561 \\
 & AIRAformer-D (Stacked) & 24 & 10 & 6,737 & 13,474 & 26,948 & 67,370 & 134,740 \\
 & AIRAformer-D (Stretched) & 23 & 11 & 6,828 & 13,656 & 27,312 & 68,280 & 136,561 \\
\midrule

\multirow{8}{*}{\textbf{1B}}
 & Composite (Stacked) & 10 & 19 & 4,552 & 9,104 & 18,208 & 45,520 & 91,041 \\
 & Composite (Stretched) & 10 & 19 & 4,552 & 9,104 & 18,208 & 45,520 & 91,041 \\
 & AIRAformer-A (Stretched) & 14 & 18 & 4,176 & 8,352 & 16,703 & 41,758 & 83,517 \\
 & AIRAformer-B (Stacked) & 14 & 18 & 4,176 & 8,352 & 16,703 & 41,758 & 83,517 \\
 & AIRAformer-C (Stacked) & 33 & 15 & 2,879 & 5,758 & 11,516 & 28,791 & 57,581 \\
 & AIRAformer-C (Stretched) & 33 & 15 & 2,879 & 5,758 & 11,516 & 28,791 & 57,581 \\
 & AIRAformer-D (Stacked) & 33 & 15 & 2,879 & 5,758 & 11,516 & 28,791 & 57,581 \\
 & AIRAformer-D (Stretched) & 33 & 15 & 2,879 & 5,758 & 11,516 & 28,791 & 57,581 \\
\midrule

\multirow{8}{*}{\textbf{3B}}
 & Composite (Stacked) & 17 & 34 & 1,651 & 3,302 & 6,605 & 16,512 & 33,024 \\
 & Composite (Stretched) & 18 & 36 & 1,559 & 3,119 & 6,238 & 15,595 & 31,190 \\
 & AIRAformer-A (Stretched) & 25 & 31 & 1,504 & 3,008 & 6,015 & 15,038 & 30,076 \\
 & AIRAformer-B (Stacked) & 21 & 32 & 1,589 & 3,178 & 6,356 & 15,889 & 31,778 \\
 & AIRAformer-C (Stacked) & 56 & 20 & 1,108 & 2,216 & 4,432 & 11,081 & 22,161 \\
 & AIRAformer-C (Stretched) & 47 & 23 & 1,203 & 2,406 & 4,812 & 12,030 & 24,061 \\
 & AIRAformer-D (Stacked) & 56 & 20 & 1,108 & 2,216 & 4,432 & 11,081 & 22,161 \\
 & AIRAformer-D (Stretched) & 47 & 23 & 1,203 & 2,406 & 4,812 & 12,030 & 24,061 \\
\bottomrule
\end{tabular}
\end{table*}

\begin{table*}[!htbp]
\centering
\caption{Training steps required to reach target isoFLOP budgets across 11 architectures and 3 parameter scales. We report the count of Mamba (Mb), MLP (M), and Attention (A) layers to contextualize computational intensity. Calculations assume a constant batch size of 524,288 tokens per step.}
\label{tab:iso_flop_steps_3prim}
\small
\begin{tabular}{ll ccc ccccc}
\toprule
 &  &  &  &  & \multicolumn{5}{c}{\textbf{Target FLOP Budget}} \\
\cmidrule(l){6-10}
\textbf{Scale} & \textbf{Architecture} & \textbf{Mb} & \textbf{M} & \textbf{A} & \textbf{2e19} & \textbf{4e19} & \textbf{8e19} & \textbf{2e20} & \textbf{4e20} \\
\midrule

\multirow{11}{*}{\textbf{350M}}
 & Nemotron-H (Approx.) & 8 & 8 & 2 & 15,402 & 30,804 & 61,608 & 154,022 & 308,044 \\
 & Nemotron-2 (Approx.) & 8 & 8 & 1 & 16,672 & 33,345 & 66,691 & 166,727 & 333,455 \\
 & Composer (2Mb-M-3A) & 10 & 4 & 12 & 9,857 & 19,715 & 39,430 & 98,577 & 197,154 \\
 & AIRAhybrid-A (Stretched) & 16 & 4 & 0 & 17,660 & 35,320 & 70,641 & 176,603 & 353,207 \\
 & AIRAhybrid-B (Stretched) & 10 & 6 & 4 & 14,130 & 28,261 & 56,523 & 141,309 & 282,618 \\
 & AIRAhybrid-B (Stacked) & 10 & 6 & 4 & 14,130 & 28,261 & 56,523 & 141,309 & 282,618 \\
 & AIRAhybrid-C (Stretched) & 9 & 4 & 16 & 8,416 & 16,833 & 33,667 & 84,168 & 168,337 \\
 & AIRAhybrid-D (Stretched) & 7 & 8 & 3 & 14,826 & 29,652 & 59,305 & 148,262 & 296,525 \\
 & AIRAhybrid-D (Stacked) & 7 & 8 & 3 & 14,826 & 29,652 & 59,305 & 148,262 & 296,525 \\
 & AIRAhybrid-E (Stretched) & 6 & 8 & 6 & 12,521 & 25,042 & 50,084 & 125,210 & 250,421 \\
 & AIRAhybrid-E (Stacked) & 6 & 8 & 6 & 12,521 & 25,042 & 50,084 & 125,210 & 250,421 \\
\midrule

\multirow{11}{*}{\textbf{1B}}
 & Nemotron-H (Approx.) & 12 & 13 & 3 & 5,732 & 11,465 & 22,930 & 57,325 & 114,650 \\
 & Nemotron-2 (Approx.) & 13 & 13 & 3 & 5,596 & 11,193 & 22,387 & 55,968 & 111,937 \\
 & Composer (M-2Mb-2A-M) & 10 & 12 & 10 & 4,841 & 9,682 & 19,365 & 48,412 & 96,825 \\
 & AIRAhybrid-A (Stretched) & 26 & 6 & 0 & 6,351 & 12,702 & 25,404 & 63,511 & 127,022 \\
 & AIRAhybrid-B (Stretched) & 16 & 10 & 6 & 5,308 & 10,616 & 21,232 & 53,081 & 106,162 \\
 & AIRAhybrid-B (Stacked) & 16 & 10 & 6 & 5,308 & 10,616 & 21,232 & 53,081 & 106,162 \\
 & AIRAhybrid-C (Stretched) & 13 & 6 & 21 & 4,033 & 8,066 & 16,132 & 40,332 & 80,664 \\
 & AIRAhybrid-D (Stretched) & 12 & 14 & 6 & 4,922 & 9,845 & 19,690 & 49,227 & 98,454 \\
 & AIRAhybrid-D (Stacked) & 12 & 14 & 6 & 4,922 & 9,845 & 19,690 & 49,227 & 98,454 \\
 & AIRAhybrid-E (Stretched) & 10 & 12 & 10 & 4,841 & 9,682 & 19,365 & 48,412 & 96,825 \\
 & AIRAhybrid-E (Stacked) & 10 & 12 & 10 & 4,841 & 9,682 & 19,365 & 48,412 & 96,825 \\
\midrule

\multirow{11}{*}{\textbf{3B}}
 & Nemotron-H (Approx.) & 16 & 17 & 4 & 1,978 & 3,956 & 7,913 & 19,782 & 39,565 \\
 & Nemotron-2 (Approx.) & 17 & 17 & 3 & 1,986 & 3,973 & 7,947 & 19,868 & 39,737 \\
 & Composer (2Mb-M-3A) & 20 & 10 & 27 & 1,443 & 2,887 & 5,774 & 14,435 & 28,871 \\
 & AIRAhybrid-A (Stretched) & 36 & 8 & 0 & 2,033 & 4,067 & 8,135 & 20,339 & 40,679 \\
 & AIRAhybrid-B (Stretched) & 22 & 13 & 8 & 1,852 & 3,704 & 7,408 & 18,520 & 37,041 \\
 & AIRAhybrid-B (Stacked) & 22 & 13 & 8 & 1,852 & 3,704 & 7,408 & 18,520 & 37,041 \\
 & AIRAhybrid-C (Stretched) & 20 & 9 & 32 & 1,361 & 2,723 & 5,447 & 13,618 & 27,237 \\
 & AIRAhybrid-D (Stretched) & 15 & 17 & 7 & 1,881 & 3,763 & 7,527 & 18,817 & 37,635 \\
 & AIRAhybrid-D (Stacked) & 15 & 17 & 7 & 1,881 & 3,763 & 7,527 & 18,817 & 37,635 \\
 & AIRAhybrid-E (Stretched) & 14 & 16 & 14 & 1,703 & 3,407 & 6,814 & 17,035 & 34,070 \\
 & AIRAhybrid-E (Stacked) & 14 & 16 & 14 & 1,703 & 3,407 & 6,814 & 17,035 & 34,070 \\
\bottomrule
\end{tabular}
\end{table*}

\newpage
\section{AIRA-Design: LRA - Task Details}
\label{app:configurable}

\begin{table}[!htbp]
\centering
\caption{Tunable hyperparameters in the 3 \textit{Configurable} tasks and their default values employed in their \textit{Non-Configurable} counterparts.}
\label{tab:model_config_defaults}
\small
\begin{tabular}{llcccl}
\toprule
\textbf{Variable in \texttt{model.py}} & \textbf{Config Key} & \textbf{Text} & \textbf{ListOps} & \textbf{Retrieval} & \textbf{Description} \\
\midrule
\texttt{SEQ\_LEN / MAX\_LENGTH} & \texttt{max\_length} & 1000 & 2000 & 4000 & Maximum seq. len.\\
\texttt{LR / LEARNING\_RATE} & \texttt{learning\_rate} & 0.0001 & 0.0001 & 0.0001 & Learning rate \\
\texttt{BATCH\_SIZE} & \texttt{batch\_size} & 16 & 16 & 8 & Batch size \\
\texttt{NUM\_TRAIN\_STEPS / TRAIN\_STEPS} & \texttt{num\_train\_steps} & 20000 & 5000 & 5000 & Training steps \\
\texttt{NUM\_LAYERS} & \texttt{num\_layers} & 6 & 4 & 4 & Transformer layers \\
\texttt{NUM\_HEADS} & \texttt{num\_heads} & 8 & 8 & 4 & Attention heads \\
\texttt{EMB\_DIM / EMBED\_DIM} & \texttt{emb\_dim} & 512 & 512 & 128 & Embedding dim. \\
\texttt{QKV\_DIM} & \texttt{qkv\_dim} & 512 & 512 & 128 & QKV dimension \\
\texttt{MLP\_DIM} & \texttt{mlp\_dim} & 2048 & 1024 & 512 & MLP hidden dim. \\
\texttt{WEIGHT\_DECAY} & \texttt{weight\_decay} & 0.01 & 0.01 & 0.01 & Weight decay \\
\texttt{WARMUP\_STEPS / WARMUP} & \texttt{warmup} & 500 & 1000 & 8000 & LR warmup steps \\
\texttt{EVAL\_FREQUENCY / EVAL\_FREQ} & \texttt{eval\_frequency} & 50 & 50 & 50 & Eval frequency \\
\texttt{RANDOM\_SEED / SEED} & \texttt{random\_seed} & 0 & 0 & 0 & Random seed \\
\bottomrule
\end{tabular}
\end{table}

Agents conducted extensive hyperparameter tuning across the 3 \textit{Configurable} version of the tasks, generating a total of 7,537 valid evaluation steps (3,189 for ListOps, 2,642 for Text, and 1,706 for Retrieval) across all greedy agents. We report a statistics of the attempted tuning in Table~\ref{tab:config_exploration}.

\begin{table*}[!htbp]
\centering
\small
\caption{Hyperparameter exploration across the three configurable LRA tasks. We report the range of values explored and the count of unique values evaluated by the agents for each configuration parameter.}
\label{tab:config_exploration}
\begin{tabular}{l cc cc cc}
\toprule
& \multicolumn{2}{c}{\textbf{ListOps}} & \multicolumn{2}{c}{\textbf{Text}} & \multicolumn{2}{c}{\textbf{Retrieval}} \\
\cmidrule(lr){2-3} \cmidrule(lr){4-5} \cmidrule(lr){6-7}
\textbf{Parameter} & \textbf{Range Explored} & \textbf{Unique} & \textbf{Range Explored} & \textbf{Unique} & \textbf{Range Explored} & \textbf{Unique} \\
\midrule
\texttt{SEQ\_LEN} & 200 -- 4096 & 8 & 32 -- 4096 & 18 & 64 -- 8192 & 9 \\
\texttt{LR} & 5e-05 -- 4e-03 & 18 & 1e-05 -- 0.01 & 21 & 3e-05 -- 5e-03 & 18 \\
\texttt{BATCH\_SIZE} & 2 -- 128 & 10 & 1 -- 64 & 9 & 1 -- 32 & 6 \\
\texttt{NUM\_TRAIN\_STEPS} & 10 -- 20000 & 32 & 2 -- 30000 & 34 & 5 -- 15000 & 27 \\
\texttt{NUM\_LAYERS} & 2 -- 12 & 9 & 1 -- 12 & 8 & 2 -- 8 & 6 \\
\texttt{NUM\_HEADS} & 1 -- 16 & 7 & 1 -- 8 & 5 & 1 -- 16 & 6 \\
\texttt{EMB\_DIM} & 32 -- 512 & 12 & 8 -- 512 & 10 & 64 -- 512 & 7 \\
\texttt{QKV\_DIM} & 32 -- 512 & 11 & 8 -- 512 & 11 & 32 -- 512 & 8 \\
\texttt{MLP\_DIM} & 128 -- 2048 & 11 & 16 -- 2048 & 13 & 64 -- 2048 & 9 \\
\texttt{WEIGHT\_DECAY} & 0.01 -- 0.1 & 6 & 0.01 -- 0.1 & 5 & 1e-04 -- 0.05 & 5 \\
\texttt{WARMUP\_STEPS} & 50 -- 2000 & 18 & 5 -- 4000 & 23 & 5 -- 3000 & 22 \\
\texttt{EVAL\_FREQUENCY} & 5 -- 1000 & 9 & 50 -- 4000 & 10 & 10 -- 1000 & 12 \\
\bottomrule
\end{tabular}
\end{table*}

\begin{table}[htbp]
    \centering
    \renewcommand{\arraystretch}{1.4}
    \footnotesize
    \caption{Breakdown of the best generated solutions for the 6 LRA AIRA-Design tasks, in their Configurable and Non-Configurable setups, detailing the generating agent, achieved score, and a brief summary of the architecture. For the Non-Configurable setup, we report the configs that deviate from default ones.}
    \label{tab:lra_summaries}
    \begin{tabularx}{\textwidth}{@{}llc>{\bfseries}l >{\raggedright\arraybackslash}p{2.0cm} X@{}}
        \toprule
        \textbf{Task} & \textbf{Setup} & \textbf{Score} & \textbf{Agent} & \textbf{Architecture} & \textbf{Summary} \\
        \midrule

        \multirow{4}{*}{\shortstack[l]{\textbf{ListOps}\\(SOTA: 0.6379)}}
        & Non-Config. & 0.4415 & \color{RoyalBlue}{Gemini 3.1 Pro} & Bi-SVR &

        \textit{Bidirectional Selective Vector Recurrence:} Projects input to $(r, k, v, a_f, a_b)$ with SiLU activations; computes element-wise $kv = k \odot v$ and data-dependent decays $\sigma(a + 2)$. Forward and backward states are accumulated via \texttt{lax.associative\_scan} with the recurrence $(g_1 g_2,\; x_1 g_2 + x_2)$. The two directions are summed, GroupNorm-ed, and gated by receptance $r$. A depthwise 1D conv provides local mixing before projection. \\
        \cmidrule{2-6}
        & Config. & 0.5050 & \color{RedOrange}{Opus 4.6} &
        DWConv + BiLinAttn \newline \normalfont\scriptsize\ttfamily NUM\_LAYERS:6, EMB\_DIM:192, MLP\_DIM:384, LR:1e-3, BATCH:64 &
        Three sub-layers per block: (1)~a large-kernel depthwise conv gated by a learned sigmoid, (2)~global non-causal linear attention with an $\text{ELU}(x)+1$ feature map computing $\phi(Q)(\phi(K)^\top V)$, and (3)~a SiLU-gated FFN ($\text{SiLU}(\text{Dense}(x)) \odot \text{Dense}(x)$). Learned positional embeddings; MEAN pooling. \\
        \midrule

        \multirow{4}{*}{\shortstack[l]{\textbf{Text}\\(SOTA: 0.9053)}}
        & Non-Config. & 0.8400 & \color{Blue}{Gemini 3 Pro} & XCTA + SwiGLU &
        \textit{Cross-Covariance Text Attention:} Computes a $d \times d$ cross-covariance matrix $C = K^\top Q$ (after L2-normalizing $Q, K$ over the sequence axis) instead of the $L \times L$ token attention matrix, achieving $O(L \cdot d^2)$ complexity. A learned per-head temperature $\tau$ scales $C$ before softmax. Uses convolutional positional encoding (depthwise 1D conv, kernel~5) and a SwiGLU FFN: $\text{SiLU}(\text{gate}) \odot \text{val}$ from a single $2 \times$ projection. \\
        \cmidrule{2-6}
        & Config. & 0.8792 & \color{Blue}{Gemini 3 Pro} &
        Bi-SSM \newline \normalfont\scriptsize\ttfamily NUM\_LAYERS:4, EMB\_DIM:128, MLP\_DIM:256, LR:5e-4, BATCH:16 &
        \textit{Bidirectional State Space Model:} Each direction runs a \texttt{ParallelLinearRNN} via associative scan: a depthwise 1D conv (kernel~3) followed by SiLU feeds into input-dependent forget ($\sigma(\cdot + 1)$) and input gates. Forward and backward hidden states are concatenated and modulated by a SiLU gate. No explicit positional embeddings---position is implicitly captured by the recurrent structure. \\
        \midrule

        \multirow{4}{*}{\shortstack[l]{\textbf{Retrieval}\\(SOTA: 0.8176)}}
        & Non-Config. & 0.7361 & \color{Magenta}{Opus 4.5} & Linear Attn + Bilinear MLP &
        \textit{Linear Attention Dual Encoder:} Replaces softmax attention with a $\phi(x) = \text{ELU}(x)+1$ linear approximation: $\phi(Q)(\phi(K)^\top V) / (\phi(Q) \cdot \mathbf{1}^\top \phi(K))$. The MLP uses a bilinear GELU form: $\text{GELU}(W_g x) \odot \text{GELU}(W_v x)$. A shared-weight dual encoder processes both documents; pooled representations are combined via NLI interaction $[\mathbf{h}_1;\,\mathbf{h}_2;\,\mathbf{h}_1 \odot \mathbf{h}_2;\,\mathbf{h}_1 - \mathbf{h}_2]$. \\
        \cmidrule{2-6}
        & Config. & 0.7943 & \color{RedOrange}{Opus 4.6} &
        Hybrid Local-Global Attn \newline \normalfont\scriptsize\ttfamily SEQ\_LEN:4000, NUM\_LAYERS:4, EMB\_DIM:192, MLP\_DIM:512, LR:3e-4, BATCH:8 &
        \textit{Hybrid Local-Global Attention:} Combines non-overlapping chunked softmax attention ($W$=256 tokens) with global linear attention (ELU+1). A per-position sigmoid gate $\sigma(Wx)$ blends: $\sigma \cdot \mathbf{y}_{\text{local}} + (1-\sigma) \cdot \mathbf{y}_{\text{global}}$. Memory is $O(nW + nd^2)$. The encoder returns both hidden states and padding masks for masked mean pooling in the dual-encoder NLI head. \\
        \bottomrule
    \end{tabularx}
\end{table}

\newpage
\section{AIRA-Design: Autoresearch - Task Details}
\label{app:autoresearch}

\begin{table}[!htbp]
\centering
\caption{Python requirements comparison between the Autoresearch implementation in \cite{karpathy2026autoresearch} and the two \airadojo{}~tasks, both with and without literature.}
\label{tab:autoresearchreq}
\begin{tabular}{cc}
\toprule
\textbf{Autoresearch \citep{karpathy2026autoresearch}} & \textbf{AutoregressiveLanguageModellingAutoresearchBPB} \\
\midrule
\texttt{torch==2.9.1 (cu128)} & \texttt{torch==2.6.0 (cu126)} \\
\texttt{numpy>=2.2.6} & \texttt{numpy>=2.2.6} \\
\texttt{pyarrow>=21.0.0} & \texttt{pyarrow>=21.0.0} \\
\texttt{requests>=2.32.0} & \texttt{requests>=2.32.0} \\
\texttt{rustbpe>=0.1.0} & \texttt{rustbpe>=0.1.0} \\
\texttt{tiktoken>=0.11.0} & \texttt{tiktoken>=0.11.0} \\
\texttt{kernels>=0.11.7} & \texttt{triton==3.1.0} \\
\texttt{matplotlib>=3.10.8} & \texttt{flash-attn==2.8.3} \\
\texttt{pandas>=2.3.3} & \texttt{ninja} \\
\bottomrule
\end{tabular}
\end{table}

We employ the pipeline of \cite{seo2025paper2code} to provide agents with useful context for the Autoresearch task. The \texttt{AutoregressiveLanguageModellingWithLiteratureAutoresearchBPB} task differs from the base version by one additional folder, \texttt{pwc/}, containing a \texttt{task\_info.jsonl} file with structured information extracted from 41 curated research papers. Each entry includes the paper title, task description, input/output specifications, a detailed pipeline description, datasets used, model architecture details, training hyperparameters, and experiment metadata, extracted by GPT-5 from the papers' \texttt{.tex} files. The papers are organized into three categories: Architecture Improvements (20 papers), Training Strategies (17 papers), and Optimizers (5 papers), as detailed in Tables~\ref{tab:literature_architecture}--\ref{tab:literature_optimizers}. For 14 of the 41 papers, working GitHub repositories are additionally provided in \texttt{pwc/code/}. Agents are instructed to consult the structured summaries to identify promising techniques, browse the reference code for implementation details, and prioritize low-cost modifications with high impact on convergence speed within the fixed 5-minute training budget. Of the 41 articles provided, 2 were published in 2021, 7 in 2023, 21 in 2024, 8 in 2025 and 3 in 2026.

\small
\begin{longtable}{p{4.5cm}p{10cm}c}
\caption{Architecture Improvement papers included in the literature review of the Autoresearch task (20 papers).} \label{tab:literature_architecture} \\
\toprule
\textbf{Paper} & \textbf{Relevance} & \textbf{Code} \\
\midrule
\endfirsthead
\endhead

Better Faster LLMs \citep{gloeckle2024better} & Predicts multiple future tokens per position during training. & \color{black}{\ding{55}} \\ \addlinespace
Differential Transformer \citep{yedifferential} & Replaces softmax attention with differential attention to cancel noise. & \color{Emerald}{\ding{51}} \\ \addlinespace
FlexAttention \citep{dong2024flex} & A PyTorch API for implementing custom attention patterns at FlashAttention speed. & \color{black}{\ding{55}} \\ \addlinespace
Forgetting Transformer \citep{lin2025forgetting} & Replaces standard softmax attention with a gated variant. & \color{black}{\ding{55}} \\ \addlinespace
Gated Delta Networks \citep{yang2024gated} & Combines a data-dependent decay gate with the delta update rule. & \color{black}{\ding{55}} \\ \addlinespace
KV Shifting Attention \citep{xu2024kv} & Modifies the attention module to shift key/value positions. & \color{black}{\ding{55}} \\ \addlinespace
Mixture of Hidden Dimensions Transformer \citep{chen2024mixture} & Selectively activates hidden sub-dimensions across layers. & \color{black}{\ding{55}} \\ \addlinespace
MoBA \citep{lu2025moba} & A drop-in, dynamically sparse self-attention mechanism. & \color{Emerald}{\ding{51}} \\ \addlinespace
MobileLLM \citep{liu2024mobilellm} & Trains compact decoder-only Transformer models optimized for on-device usage. & \color{Emerald}{\ding{51}} \\ \addlinespace
Native Sparse Attention \citep{yuan2025native} & Hardware-aligned sparse attention patterns natively trainable end-to-end. & \color{Emerald}{\ding{51}} \\ \addlinespace
gMLPs \citep{liu2021pay} & Implements gated MLPs as a drop-in replacement for self-attention. & \color{black}{\ding{55}} \\ \addlinespace
Quantizable Transformers \citep{bondarenko2023quantizable} & Removes activation outliers to enable cleaner quantization. & \color{black}{\ding{55}} \\ \addlinespace
ReLU$^2$ Wins \citep{zhang2024relu} & Shows squared ReLU produces naturally sparse hidden representations. & \color{black}{\ding{55}} \\ \addlinespace
ResiDual \citep{xie2023residual} & Dual-residual architecture combining Post-LN and Pre-LN strengths. & \color{black}{\ding{55}} \\ \addlinespace
RWKV-7 ``Goose'' \citep{peng2023rwkv} & Recurrent architecture for dynamic, expressive state evolution. & \color{black}{\ding{55}} \\ \addlinespace
Stick-breaking Attention \citep{tan2024scaling} & Replaces softmax with a sequential stick-breaking process. & \color{Emerald}{\ding{51}} \\ \addlinespace
Titans \citep{behrouz2024titans} & Augments attention with a neural long-term memory module. & \color{black}{\ding{55}} \\ \addlinespace
Ultra-Sparse Memory \citep{huang2024ultra} & Replaces MLP layers with ultra-sparse memory layers. & \color{black}{\ding{55}} \\ \addlinespace
YaRN \citep{peng2023yarn} & Extends effective context window via modified RoPE interpolation. & \color{black}{\ding{55}} \\ \addlinespace
$\mu$nit Scaling \citep{narayan2025mu} & Unit-scaled initialization for stable FP8 training. & \color{black}{\ding{55}} \\
\bottomrule
\end{longtable}

\small
\begin{longtable}{p{4.5cm}p{10.0cm}c}
\caption{Training Strategy papers included in the literature review of the Autoresearch task (16 papers).} \label{tab:literature_training} \\
\toprule
\textbf{Paper} & \textbf{Relevance} & \textbf{Code} \\
\midrule
\endfirsthead

A Spectral Condition for Feature Learning \citep{yang2023spectral} & Establishes theoretical conditions for feature learning in wide MLPs. Informs LR and width scaling choices. & \color{black}{\ding{55}} \\ \addlinespace

AutoResearch-RL: Perpetual Self-Evaluating RL Agents \citep{jain2026autoresearch} & Implements an agent that edits training scripts to optimize rewards. Discovered configurations matching hand-tuned baselines. & \color{black}{\ding{55}} \\ \addlinespace

COAT: Compressing Optimizer States and Activations \citep{xi2024coat} & Quantizes both optimizer states and activations to FP8; achieves near-lossless performance with reduced memory. & \color{Emerald}{\ding{51}} \\ \addlinespace

Cramming: Training a Language Model on a Single GPU \citep{geiping2023cramming} & Trains BERT-style models in 24 hours on one GPU. Relevant to fixed-budget optimization. & \color{Emerald}{\ding{51}} \\ \addlinespace

Depth Up-Scaling (SOLAR 10.7B) \citep{kim2024solar} & Scales Transformers by duplicating and fine-tuning middle layers to avoid training from scratch. & \color{black}{\ding{55}} \\ \addlinespace

Liger Kernel: Efficient Triton Kernels \citep{hsu2024liger} & Drop-in fused Triton kernels (CrossEntropy, RMSNorm) that are 3$\times$ faster with lower memory. & \color{Emerald}{\ding{51}} \\ \addlinespace

LLM Speedrunner \citep{zhao2025automatedllmspeedrunningbenchmark} & An agentic approach that modifies scripts to achieve lower validation cross-entropy for NanoGPT. & \color{Emerald}{\ding{51}} \\ \addlinespace

MiniCPM: Small Language Models \citep{hu2024minicpm} & Uses Warmup-Stable-Decay (WSD) schedule to make 2.4B models perform like 7B models. & \color{black}{\ding{55}} \\ \addlinespace

modded-nanogpt \citep{modded_nanogpt_2024} & The gold standard for speedruns; uses Muon, ReLU$^2$, and sliding window attention. & \color{Emerald}{\ding{51}} \\ \addlinespace

Muon is Scalable (Moonlight) \citep{liu2025muon} & Validates the Muon optimizer on 16B-parameter MoE models up to 5.7T tokens. & \color{Emerald}{\ding{51}} \\ \addlinespace

Batch Size in Stochastic Conditional Gradient \citep{islamov2026role} & Shows validation loss is batch-independent once $B$ and $S$ are large enough. & \color{black}{\ding{55}} \\ \addlinespace

Parameters vs FLOPs (MoE) \citep{abnar2025parameters} & Derives scaling laws for optimal sparsity in mixture-of-experts models. & \color{black}{\ding{55}} \\ \addlinespace

Optimal Hyperparameter Scaling Law \citep{li2025predictable} & Introduces the "Step Law" to predict optimal LR and batch size via power-law functions. & \color{black}{\ding{55}} \\ \addlinespace

Scaling Law with LR Annealing \citep{tissue2024scaling} & Proposes a specific loss curve fit ($R^2 > 0.999$) for cooldown schedules. & \color{black}{\ding{55}} \\ \addlinespace

Scaling Laws Beyond Fixed Durations \citep{hagele2024scaling} & Uses constant LR plus short cooldown to allow flexibility in training duration. & \color{black}{\ding{55}} \\ \addlinespace

The AutoResearch Moment \citep{he2026autoresearch} & Discusses the transition to autonomous research loops via \texttt{program\_md} specifications. & \color{black}{\ding{55}} \\
\bottomrule
\end{longtable}

\begin{table}[!htbp]
\centering
\small
\caption{Optimizer papers included in the literature review of the Autoresearch task (5 papers).}
\label{tab:literature_optimizers}
\begin{tabular}{p{4.5cm}p{10.0cm}c}
\toprule
\textbf{Paper} & \textbf{Relevance} & \textbf{Code} \\
\midrule
AdEMAMix: Adaptive EMA Mixture Optimizer \citep{pagliardini2024ademamix} & Augments AdamW with a second, very slow momentum ($\beta_3 \approx 0.9999$) to leverage old gradients while maintaining reactivity via a fast momentum. A 1.3B model matches AdamW validation loss in nearly half the tokens. & \color{black}{\ding{55}} \\ \addlinespace
Grokfast \citep{lee2024grokfast} & An optimizer-agnostic, 5-line gradient filtering method. Inserts between \texttt{loss.backward()} and \texttt{optimizer.step()} with no changes to the optimizer itself. & \color{Emerald}{\ding{51}} \\ \addlinespace
Maximal Update Parametrization ($\mu$P) \citep{yang2021tuning} & Enables transfer of optimal hyperparameters from small proxy models to large targets by scaling per-layer according to width. & \color{Emerald}{\ding{51}} \\ \addlinespace
Schedule-Free Optimization \citep{defazio2024road} & A drop-in replacement that eliminates the need for a pre-specified LR schedule. Maintains three parameter sequences (base $z$, evaluation $x$, gradient location $y$). & \color{Emerald}{\ding{51}} \\ \addlinespace
Sophia: A Scalable Second-Order Optimizer \citep{liu2023sophia} & Uses diagonal Hessian estimates with per-coordinate update clipping. Achieves $\sim$2$\times$ speedup over AdamW across GPT-2 and GPT-NeoX models. & \color{black}{\ding{55}} \\
\bottomrule
\end{tabular}
\end{table}

\newpage
\section{AIRA-Design: Autoresearch - Additional Results}
\label{app:appendix_autoresearch}

\begin{figure}[!htbp]
    \centering
    \includegraphics[width=\linewidth]{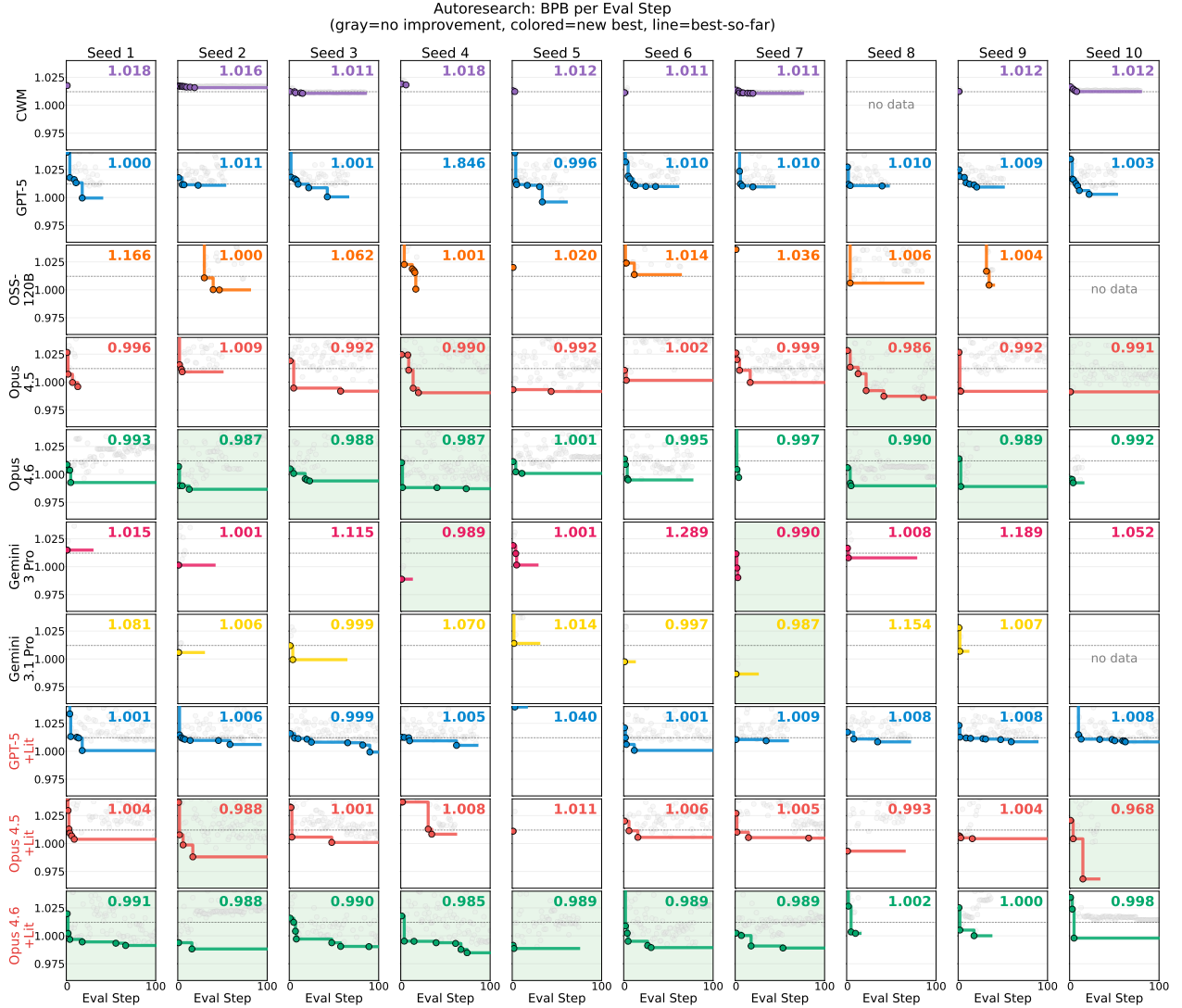}
        \caption{\textbf{Autoresearch: BPB trajectories across 10 agents and 10 seeds.} Each subplot shows the validation bits-per-byte at each evaluation step for a single agent--seed combination. Colored dots mark steps that set a new cumulative best (improvements), gray dots indicate non-improving steps, and the colored step line tracks the best-so-far BPB. The dashed line denotes the RADv1 baseline ($\text{BPB}=1.0121$). Green-shaded cells indicate seeds where the agent improved over the baseline by at least 0.0205 BPB, which is the improvement obtained in \cite{karpathy2026autoresearch}. Each subplot is labelled with the final fitness value achieved by the agent.}

    \label{fig:rawtrajectories}
\end{figure}

We perform a structural feature analysis across all agents and 100 seeds. For each improvement step (i.e., each step that achieves a new cumulative-best BPB), we extract the generated \texttt{train.py} code, parse it into a structured feature vector covering 27 dimensions (MLP activation type, model depth, head dimension, window pattern, learning rates, regularization toggles, and others) and compare it against the previous best to identify which features changed. Because agents regenerate the entire file at each step, most improvement steps involve multiple simultaneous feature changes, making causal attribution challenging. We report the results at three levels of granularity: per-feature (Table~\ref{tab:autoresearch_perfeature}), per-boolean-toggle (Table~\ref{tab:autoresearch_boolean}), and per-category (Table~\ref{tab:autoresearch_categories}).

\begin{table}[!htbp]
\centering
\caption{Per-feature BPB improvement aggregated across all agents. For each improvement step with positive $\Delta$BPB over the previous best, we extract which features changed. ``Count'' is the number of such steps where the feature changed; ``Solo'' is the number where it was the \emph{only} feature that changed.}
\label{tab:autoresearch_perfeature}
\small
\begin{tabular}{llcccccc}
\toprule
\textbf{Feature} & \textbf{Category} & \textbf{Count} & \textbf{Solo} & \textbf{Med.\,$\Delta$} & \textbf{Mean\,$\Delta$} & \textbf{Best\,$\Delta$} & \textbf{Solo Med.\,$\Delta$} \\
\midrule
\texttt{parallel\_residual}  & Embeddings / heads  &   2 & 0 & 0.0583 & 0.0583 & 0.0888 & --- \\
\texttt{label\_smoothing}    & Regularization      &   7 & 0 & 0.0291 & 0.0726 & 0.3240 & --- \\
\texttt{n\_kv\_head}         & Depth / width       &  11 & 1 & 0.0278 & 0.0595 & 0.2614 & 0.0609 \\
\texttt{attn\_dropout}       & Attention pattern   &   5 & 1 & 0.0113 & 0.0126 & 0.0291 & 0.0018 \\
\texttt{value\_embeds}       & Embeddings / heads  &  17 & 0 & 0.0103 & 0.0421 & 0.3240 & --- \\
\texttt{window\_pattern}     & Attention pattern   &  49 & 0 & 0.0096 & 0.0302 & 0.3240 & --- \\
\texttt{z\_loss}             & Regularization      &  12 & 0 & 0.0094 & 0.0397 & 0.3240 & --- \\
\texttt{partial\_rope}       & Attention pattern   &   2 & 1 & 0.0083 & 0.0083 & 0.0137 & 0.0137 \\
\texttt{learned\_prefix}     & Embeddings / heads  &   2 & 1 & 0.0075 & 0.0075 & 0.0137 & 0.0012 \\
\texttt{head\_dim}           & Depth / width       &  22 & 1 & 0.0060 & 0.0202 & 0.1786 & 0.0024 \\
\texttt{depth}               & Depth / width       &  53 & 0 & 0.0058 & 0.0306 & 0.5226 & --- \\
\texttt{mlp\_type}           & MLP / activation    &  48 & 8 & 0.0050 & 0.0351 & 0.5226 & 0.0031 \\
\texttt{batch\_size}         & Batch / curriculum  &  40 & 3 & 0.0043 & 0.0347 & 0.5226 & 0.0000 \\
\texttt{matrix\_lr}          & Learning rates      &  58 & 1 & 0.0040 & 0.0168 & 0.3240 & 0.0000 \\
\texttt{grad\_clipping}      & Regularization      &  16 & 1 & 0.0038 & 0.0114 & 0.0418 & 0.0007 \\
\texttt{embedding\_lr}       & Learning rates      &  57 & 1 & 0.0038 & 0.0161 & 0.3240 & 0.0005 \\
\texttt{curriculum}          & Batch / curriculum  &   2 & 0 & 0.0035 & 0.0035 & 0.0061 & --- \\
\texttt{weight\_decay}       & Learning rates      &  46 & 0 & 0.0034 & 0.0170 & 0.3240 & --- \\
\texttt{dual\_rope\_bases}   & Attention pattern   &   1 & 0 & 0.0034 & 0.0034 & 0.0034 & --- \\
\texttt{ema}                 & Regularization      &  10 & 0 & 0.0033 & 0.0076 & 0.0299 & --- \\
\texttt{aux\_heads}          & Embeddings / heads  &   4 & 0 & 0.0030 & 0.0061 & 0.0179 & --- \\
\texttt{token\_corruption}   & Regularization      &   2 & 0 & 0.0028 & 0.0028 & 0.0048 & --- \\
\texttt{pos\_loss\_weight}   & Batch / curriculum  &   2 & 0 & 0.0024 & 0.0024 & 0.0037 & --- \\
\texttt{weight\_tying}       & Embeddings / heads  &  11 & 0 & 0.0018 & 0.0466 & 0.3240 & --- \\
\texttt{attn\_out\_norm}     & Other               &   1 & 1 & 0.0012 & 0.0012 & 0.0012 & 0.0012 \\
\texttt{agc}                 & Regularization      &   2 & 1 & 0.0010 & 0.0010 & 0.0013 & 0.0007 \\
\texttt{attn\_temperature}   & Attention pattern   &   3 & 0 & 0.0009 & 0.0024 & 0.0062 & --- \\
\bottomrule
\end{tabular}
\end{table}

\begin{table}[!htbp]
\centering
\caption{Impact of adding vs.\ removing boolean features. Only improvement steps with positive $\Delta$BPB are included.}
\label{tab:autoresearch_boolean}
\small
\begin{tabular}{lrrrrl}
\toprule
\textbf{Feature} & \textbf{Add\,\#} & \textbf{Add med.\,$\Delta$} & \textbf{Rem\,\#} & \textbf{Rem med.\,$\Delta$} & \textbf{Verdict} \\
\midrule
\texttt{agc}                 & 1 & 0.0013 & 1 & 0.0007 & add helps more \\
\texttt{attn\_dropout}       & 2 & 0.0154 & 3 & 0.0113 & add helps more \\
\texttt{attn\_out\_norm}     & 1 & 0.0012 & 0 & ---    & add only \\
\texttt{attn\_temperature}   & 2 & 0.0036 & 1 & 0.0002 & add helps more \\
\texttt{aux\_heads}          & 2 & 0.0113 & 2 & 0.0009 & add helps more \\
\texttt{curriculum}          & 1 & 0.0061 & 1 & 0.0008 & add helps more \\
\texttt{dual\_rope\_bases}   & 1 & 0.0034 & 0 & ---    & add only \\
\texttt{ema}                 & 4 & 0.0148 & 6 & 0.0023 & add helps more \\
\texttt{grad\_clipping}      & 7 & 0.0015 & 9 & 0.0040 & remove helps more \\
\texttt{label\_smoothing}    & 2 & 0.1829 & 5 & 0.0228 & add helps more \\
\texttt{learned\_prefix}     & 1 & 0.0012 & 1 & 0.0137 & remove helps more \\
\texttt{parallel\_residual}  & 1 & 0.0888 & 1 & 0.0278 & add helps more \\
\texttt{partial\_rope}       & 1 & 0.0028 & 1 & 0.0137 & remove helps more \\
\texttt{pos\_loss\_weight}   & 1 & 0.0011 & 1 & 0.0037 & remove helps more \\
\texttt{token\_corruption}   & 1 & 0.0008 & 1 & 0.0048 & remove helps more \\
\texttt{value\_embeds}       & 11 & 0.0136 & 6 & 0.0006 & add helps more \\
\texttt{weight\_tying}       & 5 & 0.0018 & 6 & 0.0025 & remove helps more \\
\texttt{z\_loss}             & 5 & 0.0048 & 7 & 0.0127 & remove helps more \\
\bottomrule
\end{tabular}
\end{table}

\begin{table}[!htbp]
\centering
\caption{Category-level summary of BPB improvements. ``\% of improvements'' is the fraction of all positive-$\Delta$ steps involving at least one feature from that category. Categories are sorted by decreasing median $\Delta$.}
\label{tab:autoresearch_categories}
\small
\begin{tabular}{lrrrrr}
\toprule
\textbf{Category} & \textbf{Count} & \textbf{\% of improvements} & \textbf{Med.\,$\Delta$} & \textbf{Mean\,$\Delta$} & \textbf{Best\,$\Delta$} \\
\midrule
Attention pattern        & 59 & 37\% & 0.0087 & 0.0266 & 0.3240 \\
Depth / width            & 74 & 46\% & 0.0072 & 0.0259 & 0.5226 \\
MLP / activation         & 48 & 30\% & 0.0050 & 0.0351 & 0.5226 \\
Embeddings / heads       & 34 & 21\% & 0.0044 & 0.0312 & 0.3240 \\
Regularization           & 31 & 19\% & 0.0040 & 0.0214 & 0.3240 \\
Batch / curriculum       & 42 & 26\% & 0.0039 & 0.0332 & 0.5226 \\
Learning rates           & 78 & 49\% & 0.0036 & 0.0142 & 0.3240 \\
\bottomrule
\end{tabular}
\end{table}

Across 156 total improvement steps, depth and width changes are the most frequent (46\% of steps) with a median $\Delta$ of 0.007, while attention pattern changes yield the highest median $\Delta$ (0.009) but appear in fewer steps (37\%). Learning rate modifications co-occur with nearly half of all improvements (49\%), reflecting the tendency of agents to adjust hyperparameters alongside architectural changes. Among MLP type transitions, reverting to the baseline ReLU$^2$ from alternative activations consistently yields larger gains than the reverse direction. The most frequent transitions are ReLU$^2$ $\leftrightarrow$ SwiGLU (14 and 13 occurrences respectively), but SwiGLU $\to$ ReLU$^2$ has a lower median $\Delta$ (0.003) than ReLU$^2$ $\to$ SwiGLU (0.007), suggesting that SwiGLU attempts are often exploratory and the revert merely restores a working configuration. Among boolean features (Table~\ref{tab:autoresearch_boolean}), adding value embeddings (median $\Delta$=0.014, 11 occurrences) and EMA (median $\Delta$=0.015, 4 occurrences) are the most consistently beneficial additions, while removing z-loss (median $\Delta$=0.013) and gradient clipping (median $\Delta$=0.004) helps more than adding them. The ``Solo'' column in Table~\ref{tab:autoresearch_perfeature} reveals that only 7 out of 27 features were ever the sole change in an improvement step, underscoring that agents predominantly make compound modifications. This is a direct consequence of the full-file regeneration paradigm, which prevents the surgical, single-variable edits that would enable cleaner ablation.

\definecolor{codegreen}{rgb}{0,0.5,0}
\definecolor{codeblue}{rgb}{0,0,0.8}
\definecolor{codepurple}{rgb}{0.5,0,0.5}
\definecolor{backwhite}{rgb}{0.96,0.98,1.0}
\definecolor{slategray}{rgb}{0.4, 0.4, 0.4}

\lstset{
    language=Python,
    backgroundcolor=\color{backwhite},
    commentstyle=\color{codegreen}\itshape,
    keywordstyle=\color{codeblue}\bfseries,
    numberstyle=\tiny\color{slategray},
    stringstyle=\color{codepurple},
    identifierstyle=\color{black},
    basicstyle=\ttfamily\tiny,
    breakatwhitespace=false,
    breaklines=true,
    captionpos=b,
    keepspaces=true,
    numbers=left,
    numbersep=8pt,
    showspaces=false,
    showstringspaces=false,
    tabsize=2,
    frame=single,
    rulecolor=\color{slategray},
    frameround=tttt,
    framesep=5pt,
    xleftmargin=15pt
}

\newpage
\section{Best Autoresearch Solution (Greedy Opus 4.5 + Literature access)}
\label{app:code_implementation}

\begin{lstlisting}[language=Python]
import gc
import math
import time
from dataclasses import dataclass, asdict

import torch
import torch.nn as nn
import torch.nn.functional as F
from flash_attn import flash_attn_func

from prepare import MAX_SEQ_LEN, TIME_BUDGET, Tokenizer, make_dataloader, evaluate_bpb

# ---------------------------------------------------------------------------
# GPT Model
# ---------------------------------------------------------------------------

@dataclass
class GPTConfig:
    sequence_len: int = 2048
    vocab_size: int = 32768
    n_layer: int = 12
    n_head: int = 6
    n_kv_head: int = 6
    n_embd: int = 768
    window_pattern: str = "SSSL"


def norm(x):
    return F.rms_norm(x, (x.size(-1),))


def has_ve(layer_idx, n_layer):
    return layer_idx % 2 == (n_layer - 1) % 2


def apply_rotary_emb(x, cos, sin):
    assert x.ndim == 4
    d = x.shape[3] // 2
    x1, x2 = x[..., :d], x[..., d:]
    y1 = x1 * cos + x2 * sin
    y2 = x1 * (-sin) + x2 * cos
    return torch.cat([y1, y2], 3)


class CausalSelfAttention(nn.Module):
    def __init__(self, config, layer_idx):
        super().__init__()
        self.n_head = config.n_head
        self.n_kv_head = config.n_kv_head
        self.n_embd = config.n_embd
        self.head_dim = self.n_embd // self.n_head
        assert self.n_embd % self.n_head == 0
        assert self.n_kv_head <= self.n_head and self.n_head % self.n_kv_head == 0
        self.c_q = nn.Linear(self.n_embd, self.n_head * self.head_dim, bias=False)
        self.c_k = nn.Linear(self.n_embd, self.n_kv_head * self.head_dim, bias=False)
        self.c_v = nn.Linear(self.n_embd, self.n_kv_head * self.head_dim, bias=False)
        self.c_proj = nn.Linear(self.n_embd, self.n_embd, bias=False)
        self.ve_gate_channels = 32
        self.ve_gate = nn.Linear(self.ve_gate_channels, self.n_kv_head, bias=False) if has_ve(layer_idx, config.n_layer) else None

    def forward(self, x, ve, cos_sin, window_size):
        B, T, C = x.size()
        q = self.c_q(x).view(B, T, self.n_head, self.head_dim)
        k = self.c_k(x).view(B, T, self.n_kv_head, self.head_dim)
        v = self.c_v(x).view(B, T, self.n_kv_head, self.head_dim)

        if ve is not None:
            ve = ve.view(B, T, self.n_kv_head, self.head_dim)
            gate = 2 * torch.sigmoid(self.ve_gate(x[..., :self.ve_gate_channels]))
            v = v + gate.unsqueeze(-1) * ve

        cos, sin = cos_sin
        q, k = apply_rotary_emb(q, cos, sin), apply_rotary_emb(k, cos, sin)
        q, k = norm(q), norm(k)

        y = flash_attn_func(q, k, v, causal=True, window_size=window_size)
        y = y.contiguous().view(B, T, -1)
        y = self.c_proj(y)
        return y


class MLP(nn.Module):
    def __init__(self, config):
        super().__init__()
        self.c_fc = nn.Linear(config.n_embd, 4 * config.n_embd, bias=False)
        self.c_proj = nn.Linear(4 * config.n_embd, config.n_embd, bias=False)

    def forward(self, x):
        x = self.c_fc(x)
        x = F.relu(x).square()
        x = self.c_proj(x)
        return x


class Block(nn.Module):
    def __init__(self, config, layer_idx):
        super().__init__()
        self.attn = CausalSelfAttention(config, layer_idx)
        self.mlp = MLP(config)

    def forward(self, x, ve, cos_sin, window_size):
        x = x + self.attn(norm(x), ve, cos_sin, window_size)
        x = x + self.mlp(norm(x))
        return x


class GPT(nn.Module):
    def __init__(self, config):
        super().__init__()
        self.config = config
        self.window_sizes = self._compute_window_sizes(config)
        self.transformer = nn.ModuleDict({
            "wte": nn.Embedding(config.vocab_size, config.n_embd),
            "h": nn.ModuleList([Block(config, i) for i in range(config.n_layer)]),
        })
        self.lm_head = nn.Linear(config.n_embd, config.vocab_size, bias=False)
        self.resid_lambdas = nn.Parameter(torch.ones(config.n_layer))
        self.x0_lambdas = nn.Parameter(torch.zeros(config.n_layer))
        head_dim = config.n_embd // config.n_head
        kv_dim = config.n_kv_head * head_dim
        self.value_embeds = nn.ModuleDict({
            str(i): nn.Embedding(config.vocab_size, kv_dim)
            for i in range(config.n_layer) if has_ve(i, config.n_layer)
        })
        self.rotary_seq_len = config.sequence_len * 10
        cos, sin = self._precompute_rotary_embeddings(self.rotary_seq_len, head_dim)
        self.register_buffer("cos", cos, persistent=False)
        self.register_buffer("sin", sin, persistent=False)

    @torch.no_grad()
    def init_weights(self):
        # Optimized embedding initialization with smaller std for faster convergence
        torch.nn.init.normal_(self.transformer.wte.weight, mean=0.0, std=0.5)
        torch.nn.init.normal_(self.lm_head.weight, mean=0.0, std=0.001)
        n_embd = self.config.n_embd
        s = 3**0.5 * n_embd**-0.5
        for block in self.transformer.h:
            torch.nn.init.uniform_(block.attn.c_q.weight, -s, s)
            torch.nn.init.uniform_(block.attn.c_k.weight, -s, s)
            torch.nn.init.uniform_(block.attn.c_v.weight, -s, s)
            torch.nn.init.zeros_(block.attn.c_proj.weight)
            torch.nn.init.uniform_(block.mlp.c_fc.weight, -s, s)
            torch.nn.init.zeros_(block.mlp.c_proj.weight)
        self.resid_lambdas.fill_(1.0)
        self.x0_lambdas.fill_(0.1)
        for ve in self.value_embeds.values():
            torch.nn.init.uniform_(ve.weight, -s, s)
        for block in self.transformer.h:
            if block.attn.ve_gate is not None:
                torch.nn.init.zeros_(block.attn.ve_gate.weight)
        head_dim = self.config.n_embd // self.config.n_head
        cos, sin = self._precompute_rotary_embeddings(self.rotary_seq_len, head_dim)
        self.cos, self.sin = cos, sin
        self.transformer.wte.to(dtype=torch.bfloat16)
        for ve in self.value_embeds.values():
            ve.to(dtype=torch.bfloat16)

    def _precompute_rotary_embeddings(self, seq_len, head_dim, base=10000, device=None):
        if device is None:
            device = self.transformer.wte.weight.device
        channel_range = torch.arange(0, head_dim, 2, dtype=torch.float32, device=device)
        inv_freq = 1.0 / (base ** (channel_range / head_dim))
        t = torch.arange(seq_len, dtype=torch.float32, device=device)
        freqs = torch.outer(t, inv_freq)
        cos, sin = freqs.cos(), freqs.sin()
        cos, sin = cos.bfloat16(), sin.bfloat16()
        cos, sin = cos[None, :, None, :], sin[None, :, None, :]
        return cos, sin

    def _compute_window_sizes(self, config):
        pattern = config.window_pattern.upper()
        assert all(c in "SL" for c in pattern)
        long_window = config.sequence_len
        short_window = long_window // 2
        char_to_window = {"L": (long_window, 0), "S": (short_window, 0)}
        window_sizes = []
        for layer_idx in range(config.n_layer):
            char = pattern[layer_idx % len(pattern)]
            window_sizes.append(char_to_window[char])
        window_sizes[-1] = (long_window, 0)
        return window_sizes

    def estimate_flops(self):
        nparams = sum(p.numel() for p in self.parameters())
        value_embeds_numel = sum(ve.weight.numel() for ve in self.value_embeds.values())
        nparams_exclude = (self.transformer.wte.weight.numel() + value_embeds_numel +
                          self.resid_lambdas.numel() + self.x0_lambdas.numel())
        h = self.config.n_head
        q = self.config.n_embd // self.config.n_head
        t = self.config.sequence_len
        attn_flops = 0
        for window_size in self.window_sizes:
            window = window_size[0]
            effective_seq = t if window < 0 else min(window, t)
            attn_flops += 12 * h * q * effective_seq
        return 6 * (nparams - nparams_exclude) + attn_flops

    def num_scaling_params(self):
        wte = sum(p.numel() for p in self.transformer.wte.parameters())
        value_embeds = sum(p.numel() for p in self.value_embeds.parameters())
        lm_head = sum(p.numel() for p in self.lm_head.parameters())
        transformer_matrices = sum(p.numel() for p in self.transformer.h.parameters())
        scalars = self.resid_lambdas.numel() + self.x0_lambdas.numel()
        total = wte + value_embeds + lm_head + transformer_matrices + scalars
        return {
            'wte': wte, 'value_embeds': value_embeds, 'lm_head': lm_head,
            'transformer_matrices': transformer_matrices, 'scalars': scalars, 'total': total,
        }

    def setup_optimizer(self, unembedding_lr=0.004, embedding_lr=0.2, matrix_lr=0.02,
                        weight_decay=0.0, adam_betas=(0.8, 0.95), scalar_lr=0.5):
        model_dim = self.config.n_embd
        matrix_params = list(self.transformer.h.parameters())
        value_embeds_params = list(self.value_embeds.parameters())
        embedding_params = list(self.transformer.wte.parameters())
        lm_head_params = list(self.lm_head.parameters())
        resid_params = [self.resid_lambdas]
        x0_params = [self.x0_lambdas]
        assert len(list(self.parameters())) == (len(matrix_params) + len(embedding_params) +
            len(lm_head_params) + len(value_embeds_params) + len(resid_params) + len(x0_params))
        dmodel_lr_scale = (model_dim / 768) ** -0.5
        print(f"Scaling AdamW LRs by 1/sqrt({model_dim}/768) = {dmodel_lr_scale:.6f}")
        param_groups = [
            dict(kind='adamw', params=lm_head_params, lr=unembedding_lr * dmodel_lr_scale, betas=adam_betas, eps=1e-10, weight_decay=0.0),
            dict(kind='adamw', params=embedding_params, lr=embedding_lr * dmodel_lr_scale, betas=adam_betas, eps=1e-10, weight_decay=0.0),
            dict(kind='adamw', params=value_embeds_params, lr=embedding_lr * dmodel_lr_scale, betas=adam_betas, eps=1e-10, weight_decay=0.0),
            dict(kind='adamw', params=resid_params, lr=scalar_lr * 0.01, betas=adam_betas, eps=1e-10, weight_decay=0.0),
            dict(kind='adamw', params=x0_params, lr=scalar_lr, betas=(0.96, 0.95), eps=1e-10, weight_decay=0.0),
        ]
        for shape in sorted({p.shape for p in matrix_params}):
            group_params = [p for p in matrix_params if p.shape == shape]
            param_groups.append(dict(
                kind='muon', params=group_params, lr=matrix_lr,
                momentum=0.95, ns_steps=4, beta2=0.95, weight_decay=weight_decay,
            ))
        optimizer = MuonAdamW(param_groups)
        for group in optimizer.param_groups:
            group["initial_lr"] = group["lr"]
        return optimizer

    def forward(self, idx, targets=None, reduction='mean'):
        B, T = idx.size()
        assert T <= self.cos.size(1)
        cos_sin = self.cos[:, :T], self.sin[:, :T]
        x = self.transformer.wte(idx)
        x = norm(x)
        x0 = x
        for i, block in enumerate(self.transformer.h):
            x = self.resid_lambdas[i] * x + self.x0_lambdas[i] * x0
            ve = self.value_embeds[str(i)](idx) if str(i) in self.value_embeds else None
            x = block(x, ve, cos_sin, self.window_sizes[i])
        x = norm(x)
        softcap = 18
        logits = self.lm_head(x)
        logits = logits.float()
        logits = softcap * torch.tanh(logits / softcap)
        if targets is not None:
            ce_loss = F.cross_entropy(logits.view(-1, logits.size(-1)), targets.view(-1),
                                      ignore_index=-1, reduction='none')
            probs = F.softmax(logits.view(-1, logits.size(-1)), dim=-1)
            targets_flat = targets.view(-1)
            valid_mask = targets_flat != -1
            pt = torch.zeros_like(ce_loss)
            pt[valid_mask] = probs[valid_mask].gather(1, targets_flat[valid_mask].unsqueeze(-1)).squeeze(-1)
            gamma = 1.0
            focal_weight = (1 - pt) ** gamma
            focal_loss = focal_weight * ce_loss

            if reduction == 'mean':
                return focal_loss[valid_mask].mean() if valid_mask.any() else focal_loss.mean()
            elif reduction == 'none':
                return focal_loss.view(B, T)
            else:
                return focal_loss.sum()
        return logits

# ---------------------------------------------------------------------------
# Optimizer (MuonAdamW)
# ---------------------------------------------------------------------------

polar_express_coeffs = [
    (8.156554524902461, -22.48329292557795, 15.878769915207462),
    (4.042929935166739, -2.808917465908714, 0.5000178451051316),
    (3.8916678022926607, -2.772484153217685, 0.5060648178503393),
    (3.285753657755655, -2.3681294933425376, 0.46449024233003106),
    (2.3465413258596377, -1.7097828382687081, 0.42323551169305323),
]

@torch.compile(dynamic=False, fullgraph=True)
def adamw_step_fused(p, grad, exp_avg, exp_avg_sq, step_t, lr_t, beta1_t, beta2_t, eps_t, wd_t):
    p.mul_(1 - lr_t * wd_t)
    exp_avg.lerp_(grad, (1 - beta1_t).to(exp_avg.dtype))
    exp_avg_sq.lerp_(grad.square(), (1 - beta2_t).to(exp_avg_sq.dtype))
    bias1 = 1 - beta1_t ** step_t
    bias2 = 1 - beta2_t ** step_t
    denom = (exp_avg_sq / bias2).sqrt() + eps_t
    step_size = lr_t / bias1
    p.add_(exp_avg / denom, alpha=-step_size)

@torch.compile(dynamic=False, fullgraph=True)
def muon_step_fused(stacked_grads, stacked_params, momentum_buffer, second_momentum_buffer,
                    momentum_t, lr_t, wd_t, beta2_t, ns_steps, red_dim):
    momentum = momentum_t.to(stacked_grads.dtype)
    momentum_buffer.lerp_(stacked_grads, (1 - momentum).to(momentum_buffer.dtype))
    g = stacked_grads.lerp_(momentum_buffer, momentum)
    X = g.bfloat16()
    X = X / (X.norm(dim=(-2, -1), keepdim=True) * 1.02 + 1e-6)
    if g.size(-2) > g.size(-1):
        for a, b, c in polar_express_coeffs[:ns_steps]:
            A = X.mT @ X
            B = b * A + c * (A @ A)
            X = a * X + X @ B
    else:
        for a, b, c in polar_express_coeffs[:ns_steps]:
            A = X @ X.mT
            B = b * A + c * (A @ A)
            X = a * X + B @ X
    g = X
    beta2 = beta2_t.to(g.dtype)
    v_mean = g.float().square().mean(dim=red_dim, keepdim=True)
    red_dim_size = g.size(red_dim)
    v_norm_sq = v_mean.sum(dim=(-2, -1), keepdim=True) * red_dim_size
    v_norm = v_norm_sq.sqrt()
    second_momentum_buffer.lerp_(v_mean.to(dtype=second_momentum_buffer.dtype), (1 - beta2).to(second_momentum_buffer.dtype))
    step_size = second_momentum_buffer.clamp_min(1e-10).rsqrt()
    scaled_sq_sum = (v_mean * red_dim_size) * step_size.float().square()
    v_norm_new = scaled_sq_sum.sum(dim=(-2, -1), keepdim=True).sqrt()
    final_scale = step_size * (v_norm / v_norm_new.clamp_min(1e-10))
    g = g * final_scale.to(g.dtype)
    lr = lr_t.to(g.dtype)
    wd = wd_t.to(g.dtype)
    mask = (g * stacked_params) >= 0
    stacked_params.sub_(lr * g + lr * wd * stacked_params * mask)


class MuonAdamW(torch.optim.Optimizer):
    def __init__(self, param_groups):
        super().__init__(param_groups, defaults={})
        self._adamw_step_t = torch.tensor(0.0, dtype=torch.float32, device="cpu")
        self._adamw_lr_t = torch.tensor(0.0, dtype=torch.float32, device="cpu")
        self._adamw_beta1_t = torch.tensor(0.0, dtype=torch.float32, device="cpu")
        self._adamw_beta2_t = torch.tensor(0.0, dtype=torch.float32, device="cpu")
        self._adamw_eps_t = torch.tensor(0.0, dtype=torch.float32, device="cpu")
        self._adamw_wd_t = torch.tensor(0.0, dtype=torch.float32, device="cpu")
        self._muon_momentum_t = torch.tensor(0.0, dtype=torch.float32, device="cpu")
        self._muon_lr_t = torch.tensor(0.0, dtype=torch.float32, device="cpu")
        self._muon_wd_t = torch.tensor(0.0, dtype=torch.float32, device="cpu")
        self._muon_beta2_t = torch.tensor(0.0, dtype=torch.float32, device="cpu")

    def _step_adamw(self, group):
        for p in group['params']:
            if p.grad is None: continue
            grad = p.grad
            state = self.state[p]
            if not state:
                state['step'] = 0
                state['exp_avg'] = torch.zeros_like(p)
                state['exp_avg_sq'] = torch.zeros_like(p)
            state['step'] += 1
            self._adamw_step_t.fill_(state['step'])
            self._adamw_lr_t.fill_(group['lr'])
            self._adamw_beta1_t.fill_(group['betas'][0])
            self._adamw_beta2_t.fill_(group['betas'][1])
            self._adamw_eps_t.fill_(group['eps'])
            self._adamw_wd_t.fill_(group['weight_decay'])
            adamw_step_fused(p, grad, state['exp_avg'], state['exp_avg_sq'],
                            self._adamw_step_t, self._adamw_lr_t, self._adamw_beta1_t,
                            self._adamw_beta2_t, self._adamw_eps_t, self._adamw_wd_t)

    def _step_muon(self, group):
        params = group['params']
        if not params: return
        p = params[0]
        state = self.state[p]
        num_params = len(params)
        shape, device, dtype = p.shape, p.device, p.dtype
        if "momentum_buffer" not in state:
            state["momentum_buffer"] = torch.zeros(num_params, *shape, dtype=dtype, device=device)
        if "second_momentum_buffer" not in state:
            state_shape = (num_params, shape[-2], 1) if shape[-2] >= shape[-1] else (num_params, 1, shape[-1])
            state["second_momentum_buffer"] = torch.zeros(state_shape, dtype=dtype, device=device)
        red_dim = -1 if shape[-2] >= shape[-1] else -2
        stacked_grads = torch.stack([p.grad for p in params])
        stacked_params = torch.stack(params)
        self._muon_momentum_t.fill_(group["momentum"])
        self._muon_beta2_t.fill_(group["beta2"] if group["beta2"] is not None else 0.0)
        self._muon_lr_t.fill_(group["lr"] * max(1.0, shape[-2] / shape[-1])**0.5)
        self._muon_wd_t.fill_(group["weight_decay"])
        muon_step_fused(stacked_grads, stacked_params,
                        state["momentum_buffer"], state["second_momentum_buffer"],
                        self._muon_momentum_t, self._muon_lr_t, self._muon_wd_t,
                        self._muon_beta2_t, group["ns_steps"], red_dim)
        torch._foreach_copy_(params, list(stacked_params.unbind(0)))

    @torch.no_grad()
    def step(self):
        for group in self.param_groups:
            if group['kind'] == 'adamw': self._step_adamw(group)
            elif group['kind'] == 'muon': self._step_muon(group)

# ---------------------------------------------------------------------------
# Training Logic
# ---------------------------------------------------------------------------

ASPECT_RATIO = 56
HEAD_DIM = 128
WINDOW_PATTERN = "SSSL"
TOTAL_BATCH_SIZE = 2**19
EMBEDDING_LR = 0.85
UNEMBEDDING_LR = 0.006
MATRIX_LR = 0.048
SCALAR_LR = 0.5
WEIGHT_DECAY = 0.18
ADAM_BETAS = (0.8, 0.95)
WARMUP_RATIO = 0.0
WARMDOWN_RATIO = 0.55
FINAL_LR_FRAC = 0.0
DEPTH = 8
DEVICE_BATCH_SIZE = 128

t_start = time.time()
torch.manual_seed(42)
torch.set_float32_matmul_precision("high")
device = torch.device("cuda")
autocast_ctx = torch.amp.autocast(device_type="cuda", dtype=torch.bfloat16)
H100_BF16_PEAK_FLOPS = 989.5e12

tokenizer = Tokenizer.from_directory()
vocab_size = tokenizer.get_vocab_size()

def build_model_config(depth):
    base_dim = depth * ASPECT_RATIO
    model_dim = ((base_dim + HEAD_DIM - 1) // HEAD_DIM) * HEAD_DIM
    num_heads = model_dim // HEAD_DIM
    return GPTConfig(
        sequence_len=MAX_SEQ_LEN, vocab_size=vocab_size,
        n_layer=depth, n_head=num_heads, n_kv_head=num_heads, n_embd=model_dim,
        window_pattern=WINDOW_PATTERN,
    )

config = build_model_config(DEPTH)
with torch.device("meta"):
    model = GPT(config)
model.to_empty(device=device)
model.init_weights()
num_params = model.num_scaling_params()['total']
num_flops_per_token = model.estimate_flops()
grad_accum_steps = TOTAL_BATCH_SIZE // (DEVICE_BATCH_SIZE * MAX_SEQ_LEN)

optimizer = model.setup_optimizer(
    unembedding_lr=UNEMBEDDING_LR, embedding_lr=EMBEDDING_LR,
    scalar_lr=SCALAR_LR, adam_betas=ADAM_BETAS,
    matrix_lr=MATRIX_LR, weight_decay=WEIGHT_DECAY,
)
model = torch.compile(model, dynamic=False)
train_loader = make_dataloader(tokenizer, DEVICE_BATCH_SIZE, MAX_SEQ_LEN, "train")
x, y, epoch = next(train_loader)

def get_lr_multiplier(progress):
    if progress < 1.0 - WARMDOWN_RATIO: return 1.0
    cooldown = (1.0 - progress) / WARMDOWN_RATIO
    return cooldown * 1.0 + (1 - cooldown) * FINAL_LR_FRAC

t_start_training = time.time()
smooth_train_loss = 0
total_training_time = 0
step = 0

while True:
    torch.cuda.synchronize()
    t0 = time.time()
    for _ in range(grad_accum_steps):
        with autocast_ctx: loss = model(x, y)
        train_loss = loss.detach()
        (loss / grad_accum_steps).backward()
        x, y, epoch = next(train_loader)

    progress = min(total_training_time / TIME_BUDGET, 1.0)
    lrm = get_lr_multiplier(progress)
    for group in optimizer.param_groups:
        group["lr"] = group["initial_lr"] * lrm
        if group['kind'] == 'muon':
            group["momentum"] = (1 - min(step/300, 1)) * 0.85 + min(step/300, 1) * 0.95
            group["weight_decay"] = WEIGHT_DECAY * (1 - progress)
    optimizer.step()
    model.zero_grad(set_to_none=True)

    torch.cuda.synchronize()
    t1 = time.time()
    dt = t1 - t0
    if step > 10: total_training_time += dt
    smooth_train_loss = 0.9 * smooth_train_loss + 0.1 * train_loss.item()
    step += 1
    if step > 10 and total_training_time >= TIME_BUDGET: break

model.eval()
with autocast_ctx:
    val_bpb = evaluate_bpb(model, tokenizer, DEVICE_BATCH_SIZE)

print(f"val_bpb: {val_bpb:.6f}")
\end{lstlisting}

\end{document}